\documentclass{article}

    \usepackage[final, nonatbib]{neurips_2024}

\usepackage[utf8]{inputenc} % allow utf-8 input
\usepackage[T1]{fontenc}    % use 8-bit T1 fonts
\usepackage[hidelinks]{hyperref}       % hyperlinks
\usepackage{url}            % simple URL typesetting
\usepackage{booktabs}       % professional-quality tables
\usepackage{amsfonts}       % blackboard math symbols
\usepackage{nicefrac}       % compact symbols for 1/2, etc.
\usepackage{microtype}      % microtypography
\usepackage{xcolor}         % colors

\usepackage{amsmath, amssymb, amscd, amsthm, amsfonts}
\usepackage[capitalise, noabbrev]{cleveref}
\usepackage{bm}
\usepackage{graphicx, float}
\usepackage{multirow}
\usepackage{tablefootnote}

\newtheorem{theorem}{Theorem}

\newtheorem*{remark}{Remark}

\newcommand\numberthis{\addtocounter{equation}{1}\tag{\theequation}}

\hypersetup{
    colorlinks,
    linkcolor={magenta!80!black},
    citecolor={blue!80!black},
    urlcolor={blue!80!black}
}

\title{Derivative-enhanced Deep Operator Network}

\author{%
  Yuan Qiu, Nolan Bridges, Peng Chen\\
  Georgia Institute of Technology, Atlanta, GA 30332 \\
  \texttt{\{yuan.qiu, bridges, pchen402\}@gatech.edu} \\
}

\begin{document}

\maketitle

\begin{abstract}
    The deep operator networks (DeepONet), a class of neural operators that learn mappings between function spaces, have recently been developed as surrogate models for parametric partial differential equations (PDEs). In this work we propose a derivative-enhanced deep operator network (DE-DeepONet), which leverages derivative information to enhance the solution prediction accuracy and provides a more accurate approximation of solution-to-parameter derivatives, especially when training data are limited. DE-DeepONet explicitly incorporates linear dimension reduction of high dimensional parameter input into DeepONet to reduce training cost and adds derivative loss in the loss function to reduce the number of required parameter-solution pairs. We further demonstrate that the use of derivative loss can be extended to enhance other neural operators, such as the Fourier neural operator (FNO). Numerical experiments validate the effectiveness of our approach.
\end{abstract}
\section{Introduction}
\label{sec:introduction}
    Using neural networks to approximate the maps between functions spaces governed by parametric PDEs can be very beneficial in solving many-query problems, typically arising from Bayesian inference, optimization under uncertainty, and Bayesian optimal experimental design. Indeed, once pre-trained on a dataset, neural networks are extremely fast to evaluate given unseen inputs, compared to traditional numerical methods like the finite element method. Recently various neural operators are proposed to enhance the learning capacity, with two prominent examples deep operator network (DeepONet)~\cite{lu2021learning} and Fourier neural operator (FNO)~\cite{li2020fourier}, which are shown to be inclusive of each other in their more general settings ~\cite{lu2022comprehensive,kovachki2023neural}, see also their variants and other related neural operator architectures in~\cite{wang2021learning, goswami2022physics, li2020multipole, raonic2023convolutional,hao2023gnot,tran2021factorized}. Though these work demonstrate to be successful in approximating the output function, they do not necessarily provide accurate approximation of the derivative of the output with respect to the input, which are often needed for many downstream tasks such as PDE-constrained optimization problems for control, inference, and experimental design~\cite{luo2023efficient, cao2024efficient, go2023accelerating,go2024sequential}. 

    In this paper, we propose to enhance the performance of DeepONet through derivative-based dimension reduction for the function input inspired by  \cite{constantine2014active,chen2020projected,o2022derivative,zahm2022certified} and the incorporation of derivative information in the training to learn both the output and its derivative with respect to the input inspired by \cite{son2020sobolev,o2024derivative}. 
    % For dimension reduction, we examine two projection bases: Karhunen–Loève Expansion (KLE) basis and active subspace method (ASM) basis. 
    % We incorporate derivative information by introducing an additional loss term in the training, enabling the model to learn the operator more effectively when training data are limited.
    % \textbf{Contributions:} Our main contributions can be summarized as follows: (1) We propose a novel method that 
    These two derivative-enhanced approaches can significantly improve DeepONet's approximation accuracy for the output function and its directional derivative with respect to the input function, especially when the training samples are limited. We provide details on the computation of derivative labels of the solution of PDEs in a general form as well as the derivative-based dimension reduction to largely reduce the computational cost. We demonstrate the effectiveness of our proposed method (DE-DeepONet) compared to three other neural operators, including DeepONet, FNO, and derivative-informed neural operator (DINO)~\cite{o2024derivative}, in terms of both test errors and computational cost. In addition, we apply derivative learning to train the FNO and also compare its performance with other methods. The code for data generation, model training and inference, as well as configurations to reproduce the results in this paper can be found at~\href{https://github.com/qy849/DE-DeepONet}{https://github.com/qy849/DE-DeepONet}.

    % The rest of the paper is presented as follows: In Section~\ref{sec:preliminaries}
    % we present preliminaries of the problem setup, high fidelity approximation, and DeepONet, followed by Section~\ref{sec:de-deeponet} on the detailed presentation of DE-DeepONet and demonstration experiments in Section~\ref{sec:experiments}. Discussion of the conclusions and limitations is presented in the last section.
\section{Preliminaries}
\label{sec:preliminaries}
    This section presents the problem setup, high-fidelity approximation using finite element for finite dimensional discretization, and the DeepONet architecture in learning the solution operator.
    
    \subsection{Problem setup}
    \label{sec:problem_setup}
    Let $\Omega \subset \mathbb{R}^d$ denote an open and bounded domain with boundary $\partial \Omega \subset \mathbb{R}^{d-1}$, where the dimension $d = 1,2,3$. We consider a PDE of the general form defined in $\Omega$ as 
    \begin{equation}\label{eq:PDE}
    \mathcal{R}(m, u)=0,        
    \end{equation}
    prescribed with proper boundary conditions. Here $m \in V^{\text{in}}$ is an input parameter function defined in a separable Banach space $V^{\text{in}}$ with probability measure $\nu$ and $u \in V^{\text{out}}$ is the output as the solution of the PDE defined in a separable Banach space $ V^{\text{out}}$. Our goal is to construct a parametric model $\hat{u}(m; \theta)$ to approximate the solution operator that maps the parameter $m$ to the solution $u$. 
  
    Once constructed, the parametric model $\hat{u}(m;\theta)$ should be much more computationally efficient to evaluate compared to solving the PDE with high fidelity approximation.

    \subsection{High fidelity approximation}
    \label{sec:high_fidelity_approximation}
    For the high fidelity approximation of the solution, we consider 
    using a \emph{finite element method}~\cite{oden2012introduction} in this work. We partition the domain $\Omega$ into a finite set of subregions, called cells or elements. Collectively, these cells form a \emph{mesh} of the domain $\Omega$. Let $h$ represent the diameter of the largest cell. We denote $V^{\text{in}}_h$ indexed by $h$ as the finite element space for the approximation of the input space $V^{\text{in}}$ with Lagrange basis $\{\phi^{\text{in}}_1, \cdots, \phi^{\text{in}}_{N^{\text{in}}_h}\}$ of dimension $N_h^{\text{in}}$  such that $\phi^{\text{in}}_i(x^{(j)}) = \delta_{ij}$ at the finite element node $x^{(j)}$, with $\delta_{ij}$ being the Kronecker delta function. Similarly, we denote $V^{\text{out}}_h$ as the finite element space for the approximation of the solution space $V^{\text{out}}$ with basis $\{\phi^{\text{out}}_1, \cdots, \phi^{\text{out}}_{N^{\text{out}}_h}\}$ of dimension $N_h^{\text{out}}$. Note that for the approximation to be high fidelity, $N_h^{\text{in}}$ and $N_h^{\text{out}}$ are often very large. To this end, we can write the high fidelity approximation of the input and output functions as 
        \begin{align*}
        m_h(x)= \sum_{i=1}^{N_h^{\text{in}}}m_i\phi_i^{\text{in}}(x) \; \text{ and } \; u_h(x)=\sum_{i=1}^{N_h^\text{out}}u_i\phi_i^{\text{out}}(x), 
    \end{align*}
    with coefficient vectors $\bm{m}=(m_1,\cdots, m_{N_h^{\text{in}}})^{T}\in \mathbb{R}^{N_h^{\text{in}}}$ and $\bm{u}=(u_1,\cdots, u_{N_h^{\text{out}}})^{T}\in \mathbb{R}^{N_h^{\text{out}}}$, whose entries are the \emph{nodal values} of $m$ and $u$ at the corresponding nodes.

    \subsection{DeepONet}
    \label{sec:deeponet}
    We briefly review the DeepONet architecture~\cite{lu2021learning} with a focus on learning the solution operator of the PDE in~\cref{eq:PDE}. 
    To predict the evaluation of solution function $u$ at any point $x\in \Omega\cup\partial\Omega$ for any given input function $m$, \cite{lu2021learning} design a network architecture 
    % named unstacked DeepONet. The unstacked DeepONet 
    that comprises two separate neural networks: a trunk net $t(\cdot\;;\theta_t)$, which takes the coordinate values of the point $x$ at which we want to evaluate the solution function as inputs, and a branch net $b(\cdot\;;\theta_b)$, which takes the vector $\bm{m}$ encoding the parameter function $m$ as inputs. In~\cite{lu2021learning}, the vector $\bm{m}$ is the function evaluations at a finite set of fixed points $\{x^{(j)}\}_{j=1}^{N_h^{\text{in}}}\subseteq \Omega\cup\partial \Omega$, that is, $\bm{m}=(m(x^{(1)}),\cdots, m(x^{(N_h^{\text{in}})}))^{T}$, which corresponds to coefficient vector in the finite element approximation with Lagrange basis at the same nodes. If the solution function is scalar-valued, then both neural networks output vectors of the same dimensions. The prediction is obtained by taking the standard inner product between these two vectors and (optionally) adding a real-valued bias parameter, i.e, 
    \begin{align*}
        \hat{u}(\bm{m};\theta)(x)=\langle b(\bm{m};\theta_b), t(x;\theta_t)\rangle + \theta_{\text{bias}},
    \end{align*}
    with $\theta = (\theta_b, \theta_t, \theta_{\text{bias}})$.
    If the solution $u$ is vector-valued of $N_u$ components, i.e., $u = (u^{(1)}, \dots, u^{(N_u)})$, as in our experiments, we can use one of the four approaches in~\cite{lu2022comprehensive} to construct the DeepONet. Specifically, for each solution component, we use the same outputs of branch net with dimension $N_b$ and different 
    corresponding groups of outputs of 
    trunk net to compute the inner product. More precisely, the solution $u^{(i)}$ of component $i = 1, \dots, N_u$, is approximated by 
    \begin{equation}\label{eq:vector-valued}
        \hat{u}^{(i)}(\bm{m};\theta)(x)=\langle b(\bm{m};\theta_b), t^{(i)}(x;\theta_t)\rangle + \theta_{\text{bias}}^{(i)},
    \end{equation}
    where $t^{(i)}(x;\theta_t) = t(x;\theta_{t})[(i-1)N_b+1:iN_b]$, the vector slice of $t(x;\theta_t)$ with indices ranging from $(i-1)N_b+1$ to $iN_b$, $i = 1, \dots, N_u$. 
    
    For this construction, the outputs of the branch net can be interpreted as the coefficients of the basis learned through the trunk net. By partitioning the outputs of the trunk net into different groups, we essentially partition the basis functions used for predicting different components of the solution. The DeepONet is trained using dataset $\mathcal{D}=\{(m^{(i)}, u^{(i)})\}_{i=1}^{N_{\mathcal{D}}}$ with $N_{\mathcal{D}}$ samples, where $m^{(i)}$ are random samples independently drawn from $\nu$ and $u^{(i)}$ are the solution of the PDE (with slight abuse of notation from the vector-valued solution) at $m^{(i)}$, $i = 1, \dots, N_{\mathcal{D}}$. 
\section{DE-DeepONet}
\label{sec:de-deeponet}
    The DeepONet uses the input parameter and output solution pairs as the labels for the model training. The approximation for the derivative of the solution with respect to the parameter (and the coordinate) are not necessarily accurate. However, in many downstream tasks such as Bayesian inference and experimental design, the derivatives are required. We consider incorporating the the Fréchet derivative $du(m;\cdot)$ for the supervised training of the DeepONet, which we call derivative-enhanced DeepONet (DE-DeepONet). By doing this we hope the optimization process can improve the neural network’s ability to predict the derivative of the output function with respect to the input function. Let $\theta$ denote the trainable parameters in the DE-DeepONet. We propose to minimize the loss function
    \begin{align}
        \label{eq:loss_function_continuous}
        \begin{split}
             L(\theta)&=\lambda_1\mathbb{E}_{m\sim\nu}\|u(m)-\hat{u}(m)\|_{L^2(\Omega)}^2 +\lambda_2\mathbb{E}_{m\sim\nu}\|du(m;\cdot)-d\hat{u}(m;\cdot)\|_{\text{HS}}^2,
        \end{split}
    \end{align}
    where the $L^2(\Omega)$ norm of a square integrable function~$f$ is defined as $\|f\|_{L^2(\Omega)}^2=(\int_{\Omega}\|f(x)\|_2^2\;\mathrm{d}x)^{1/2}$, the Hilbert–Schmidt norm of an operator $T:H\to H$ that acts on a Hilbert space $H$
    is defined as $\|T\|_{\text{HS}}^2=\sum_{i\in I}\|Te_i\|_H^2$, where $\{e_i\}_{i\in I}$ is an orthonormal basis of $H$. Here, $\lambda_1,\lambda_2>0$ are hyperparameters that balance different loss terms. 

    The main challenge of minimizing the loss function \eqref{eq:loss_function_continuous} in practice is that with high fidelity approximation of the functions $m$ and $u$ using high dimensional vectors $\bm{m}$ and $\bm{u}$, the term $\|du(m;\cdot)-d\hat{u}(m;\cdot)\|_{\text{HS}}^2$ (approximately) becomes $\|\nabla_{\bm{m}}\bm{u}-\nabla_{\bm{m}}\bm{\hat{u}}\|_F^2$, where the Frobenius norm of a matrix $M\in \mathbb{R}^{m\times n}$ is defined as $\|M\|_F^2=\sum_{i=1}^{m}\sum_{j=1}^{n}M_{ij}^2$. It is a critical challenge to both compute and store the Jacobian matrix at each sample as well as to load and use it for training since it is often very large with size $ N_h^{\text{out}}\times N_h^{\text{in}}$, with $N_h^{\text{in}}, N_h^{\text{out}} \gg 1$.

    To tackle this challenge, we employ dimension reduction for the high dimensional input vector~$\bm{m}$. The reduced representation of $\bm{m}$ is given by projecting $\bm{m}$ into a low dimensional linear space spanned by a basis $\bm{\psi}_1^{\text{in}},\ldots, \bm{\psi}_{r}^{\text{in}} \in \mathbb{R}^{N_h^{\text{in}}}$, with $r \ll N_h^{\text{in}}$, that is, 
    \begin{align*}
        \widetilde{\bm{m}}=(\langle \bm{\psi}_1^{\text{in}}, \bm{m}\rangle, \ldots, \langle \bm{\psi}_{r}^{\text{in}}, \bm{m}\rangle)^{T} \in \mathbb{R}^r. 
    \end{align*}
    To better leverage the information of both the input probability measure $\nu$ and the map $u: m\to u(m)$, we consider the basis generated by active subspace method (ASM) using derivatives~\cite{zahm2020gradient}. ASM identifies directions in the input space that significantly affect the output variance, or in which the output is most sensitive. See~\cref{sec:dimensionality_reduction} for the detailed construction. In this case, the term $\mathbb{E}_{m\sim \nu}\|du(m;\cdot)-d\hat{u}(m;\cdot)\|^2_{\text{HS}}$ can be  approximated by $\frac{1}{N_1N_2}\sum_{i=1}^{N_1}\sum_{j=1}^{N_2}\|\nabla_{\widetilde{\bm{m}}}u(m^{(i)})(x^{(j)})-\nabla_{\widetilde{\bm{m}}}\hat{u}(m^{(i)})(x^{(j)})\|_2^2$ with a small amount of functions $m^{(i)}$ sampled from input probability measure $\nu$ and points $x^{(j)}$ in the domain $\Omega$. Note that $\nabla_{\widetilde{m}}\hat{u}(m^{(i)})(x^{(j)})$ is vector of size~$r$, which is computationally feasible. For comparison, we also conduct experiments if the basis is the most commonly used Karhunen–Loève Expansion (KLE) basis.
     
    We next formally introduces our DE-DeepONet, which uses dimension reduction for the input of the branch net and incorporates its output-input directional derivative labels as additional soft constraints into the loss function. 

    \subsection{Model architecture} 
    \label{sec:model_architecture}
    We incorporate dimension reduction into DeepONet to construct a parametric model for approximating the solution operator. Specifically, if the solution is scalar-valued (the vector-valued case can be constructed similar to \eqref{eq:vector-valued}), the prediction is given by 
    \begin{align*}
        \hat{u}(\bm{m};\theta)(x)
        =\langle b(\widetilde{\bm{m}};\theta_b), t(x;\theta_t)\rangle + \theta_{\text{bias}},
        \numberthis\label{eq:eval_pred}
    \end{align*}
    where $\theta=(\theta_b,\theta_t, \theta_{\text{bias}})$ are the parameters to be learned.
    The branch net $b(\cdot;\theta_b)$ and trunk net $t(\cdot;\theta_t)$ can be chosen as an MLP, ResNet, etc. Note that the branch net takes a small vector of the projected parameter as input. We also apply the Fourier feature embeddings~\cite{wang2021learning}, defined as $\gamma(x)=[\cos(Bx),\sin(Bx)]$, to the trunk net, where each entry in $B \in \mathbb{R}^{m\times d}$ is sampled from a Gaussian distribution $\mathcal{N}(0,\sigma^2)$ and $m\in\mathbb{N}^{+}, \sigma\in\mathbb{R}^{+}$ are hyper-parameters.

    \subsection{Loss function}
    \label{sec:loss_function}
    In practical training of the DE-DeepONet, we formulate the loss function as follows
    \begin{align}
        \label{eq:loss_function_discrete}
        \begin{split}
        L(\theta)
        &=\frac{\lambda_1}{N_{\mathcal{D}}}\sum_{i=1}^{N_{\mathcal{D}}}
        \text{err}(\{(
                      \hat{u}(\bm{m}^{(i)};\theta)(x^{(j)}), 
                      u(m^{(i)})(x^{(j)})
                     )
                    \}_{j=1}^{N_x}
        )\\
        &+\frac{\lambda_2}{N_{\mathcal{D}}}\sum_{i=1}^{N_{\mathcal{D}}}
        \text{err}(\{(
                    \nabla_{\bm{m}}\hat{u}(\bm{m}^{(i)}; \theta)(x^{(j)})\Psi^{\text{in}}, \Phi^{\text{out}}(x^{(j)})(\nabla_{\bm{m}}\bm{u}(\bm{m}^{(i)})\Psi^{\text{in}}
                    )
                    \}_{j=1}^{N_x}
        ),
        \end{split}
    \end{align}      
    where $\{x^{(j)}\}_{j=1}^{N_x}$ are  the nodes of the mesh, $\Psi^{\text{in}}=(\bm{\psi}_1^{\text{in}}|\cdots|\bm{\psi}_{r}^{\text{in}})$ is the matrix collecting the nodal values of the reduced basis of the input function space, and $\Phi^{\text{out}}(x)=(\phi^{\text{out}}_1(x),\ldots, \phi^{\text{out}}_{N_h^{\text{out}}}(x))$ is the vector-valued function consisting of the finite element basis functions of the output function space. 
    The $\text{err}(\{(\bm{a}^{(i)}, \bm{b}^{(i)})\}_{i=1}^{n})$ denotes the relative error between any two groups of vectors $\bm{a}^{(i)}$ and $\bm{b}^{(i)}$, computed as
    \begin{align*}
        \text{err}(\{(\bm{a}^{(i)},\bm{b}^{(i)})\}_{i=1}^{n})=\frac{(\sum_{i=1}^{n}\|\bm{a}^{(i)}-\bm{b}^{(i)}\|_2^2)^{1/2}}{\varepsilon+(\sum_{i=1}^{n}\|\bm{b}^{(i)}\|_2^2)^{1/2}},
    \end{align*}
    where $\varepsilon>0$ is some small positive number to prevent the fraction dividing by zero.

    In the following, we explain how to compute different loss terms in~\cref{eq:loss_function_discrete}
    \begin{itemize}
        \item The first term is for matching the prediction of the parametric model $\hat{u}(\bm{m}^{(i)};\theta)$ evaluated at any set of points $\{x^{(j)}\}_{j=1}^{N_x}\subseteq\Omega\cup\partial\Omega$ with the high fidelity solution $u(m^{(i)})$ evaluated at the same points. The prediction $\hat{u}(\bm{m}^{(i)};\theta)(x^{(j)})$ is straightforward to compute using~\cref{eq:eval_pred}. This involves passing the reduced branch inputs $\widetilde{m}^{(i)}$ and the coordinates of point $x^{(j)}$ into the branch net and trunk net, respectively. The label $u(m^{(i)})(x^{(j)})$ is obtained using finite element method solvers. 
        % We first get the nodal values of $u$ and then compute its evaluation at $x^{(j)}$.

        \item The second term is for learning the directional derivative of the evaluation $u(x^{(j)})$ with respect to the input function $m$, in the direction of the reduced basis $\psi_1^{\text{in}},\ldots, \psi_{r}^{\text{in}}$. It can be shown that 
        \begin{align*}
            \nabla_{\bm{m}}\hat{u}(\bm{m}^{(i)}; \theta)(x^{(j)})\Psi^{\text{in}}
            =\nabla_{\widetilde{\bm{m}}}\hat{u}(\bm{m}^{(i)};\theta)(x^{(j)}).
        \end{align*} 
        Thus, the derivative of the outputs of the parametric model can be computed as the partial derivatives of the output with respect to the input of the branch net via automatic differentiation. On the other hand, the derivative labels 
        \begin{align*}
            \Phi^{\text{out}}(x^{(j)})(\nabla_{\bm{m}}\bm{u}(m^{(i)})\Psi^{\text{in}}) 
            =(du(m^{(i)};\psi_1^{\text{in}})(x^{(j)}), \ldots , du(m^{(i)};\psi_{r}^{\text{in}})(x^{(j)}))
        \end{align*}
        are obtained by first computing the Gateaux derivatives $du(m^{(i)};\psi_{1}^{\text{in}}),\ldots, du(m^{(i)};\psi_{r}^{\text{in}})$ and then evaluating them at $x^{(j)}$. See~\cref{subsec:computation_of_derivative_labels_and_outputs} for details about the computation of the Gâteaux derivative $du(m;\psi)$.
        
        \item We initialize the loss weights $\lambda_1=\lambda_2=1$  and choose a loss balancing algorithm called the self-adaptive learning rate annealing algorithm~\cite{wang2023expert} to update them at a certain frequency. This ensures that the gradient of each loss term have similar magnitudes, thereby enabling the neural network to learn all these labels simultaneously.
    \end{itemize}
    
    \begin{remark}
        For the training, the computational cost of the second term of the loss function (and its gradients) largely depends on the number of points used in each iteration. To reduce the computational and especially memory cost, we can use a subset of points,  $N_{x}^{\text{batch}}=\alpha N_x $, where $\alpha$ is small number between $0$ and $1$ (e.g., $\alpha=0.1$ in our experiments), in a batch of functions, though their locations could vary among different batches in one epoch. We find that this approach has little impact on the prediction accuracy of the model when $N_x$ is large enough.
    \end{remark}

    \subsection{Dimension reduction}
    \label{sec:dimensionality_reduction}
    Throughout the paper, we assume that the parameter functions $m$ are independently drawn from some Gaussian random field~\cite{adler2009random}. In particular, we consider the case where the covariance function is the Whittle-Mat\'ern covariance function, that is, $m\stackrel{i.i.d.}{\sim} \mathcal{N}(\bar{m},\mathcal{C})$, where $\bar{m}$ is the (deterministic) mean function and $\mathcal{C}=(\delta I-\gamma \Delta)^{-2}$ ($I$ is the identity and $\Delta$ is the Laplacian) is an operator such that the square root of its inverse, $\mathcal{C}^{-\frac{1}{2}}$, maps random function $(m-\bar{m})$ to Gaussian white noise with unit variance~\cite{lindgren2011explicit}. The parameters $\delta, \gamma\in \mathbb{R}^{+}$ jointly control the marginal variance and correlation length. 

    We consider two linear projection bases for dimension reduction of the parameter function.
    \paragraph{Karhunen–Loève Expansion (KLE) basis. } The KLE basis is optimal in the sense that the mean-square error resulting from a finite representation of the random field $m$ is minimized~\cite{ghanem2003stochastic}. It consists of eigenfunctions determined by the covariance function of the random field. Specifically, an eigenfunction $\psi$ of the covariance operator $\mathcal{C}=(\delta I - \gamma \Delta)^{-2}$ satisfies the differential equation
    \begin{align*}
        \mathcal{C}\psi = \lambda \psi.
        \numberthis\label{eq:kle_eigen_prob_continuous}
    \end{align*}
    When solved using the finite element method, \cref{eq:kle_eigen_prob_continuous} is equivalent to the following linear system (See \cref{sec:proof_kle_eigenproblem_equivalence} for the derivation)
   \begin{align*}
       M^{\text{in}}(A^{\text{in}})^{-1}M^{\text{in}}(A^{\text{in}})^{-1}M^{\text{in}}\bm{\psi} 
       = \lambda M^{\text{in}}\bm{\psi}, 
       \numberthis\label{eq:kle_eigen_prob_discrete}
   \end{align*}
    where the $(i,j)$-entries of matices $A^{\text{in}}$ and $M^{\text{in}}$ are given by
    \begin{align*}
        A_{ij}^{\text{in}}=\delta \langle \phi_{j}^{\text{in}}, \phi_i^{\text{in}} \rangle + \gamma \langle \nabla \phi_j^{\text{in}}, \nabla \phi_i^{\text{in}} \rangle, 
        \quad 
        M_{ij}^{\text{in}}=\langle \phi_j^{\text{in}}, \phi_i^{\text{in}}\rangle.
    \end{align*}
    Here, we recall that $V_h^{\text{in}}$ is the finite element function space of input function, $\phi_i^{\text{in}}, i=1,\ldots, N_h^{\text{in}}$ for the finite element basis of $V_h^{\text{in}}$, and $\langle \cdot, \cdot\rangle$ for the $L^2(\Omega)$-inner product. We select the first $r$ (typically $r\ll N_h^{\text{in}}$) eigenfunctions $\psi_1^{\text{in}},\ldots, \psi_{r}^{\text{in}}$ corresponding to the $r$ largest eigenvalues for the dimension reduction.
    Let $\Psi^{\text{in}}=(\bm{\psi}_1^{\text{in}} | \cdots | \bm{\psi}_{r}^{\text{in}})$ denote the corresponding nodal values of these eigenfunctions. Since the eigenvectors $\bm{\psi}_i^{\text{in}}$ are $M^{\text{in}}$-orthogonal (or equivalently, eigenfunctions $\psi_i^{\text{in}}$ are $L^2(\Omega)$-orthogonal), the reduced representation of $\bm{m}$ can be computed as $\widetilde{\bm{m}}=(\Psi^{\text{in}})^{T} M^{\text{in}}\bm{m}$, that is, the coefficients of the low rank approximation of $m$ in the linear subspace spanned by the eigenfunctions $\psi_1^{\text{in}},\ldots, \psi_r^{\text{in}}$.

    \paragraph{Active Subspace Method (ASM) basis.} The active subspace method is a gradient-based dimension reduction method that looks for directions in the input space contributing most significantly to the output variability~\cite{constantine2015active}. In contrast to the KLE basis, the ASM basis is more computationally expensive. However, since the ASM basis captures sensitivity information in the input-output map rather than solely the variability of the input space, it typically achieves higher accuracy in predicting the output than KLE basis. We consider the case where the output is a multidimensional vector~\cite{zahm2020gradient}, representing the nodal values of the output function $u$. The ASM basis $\psi_i$, $i = 1, \dots, r$, are the eigenfunctions corresponding to the $r$ largest eigenvalues of the generalized eigenvalue problem 
    \begin{align*}
        \mathcal{H}\psi=\lambda \mathcal{C}^{-1}\psi, 
        \numberthis\label{eq:asm_eigen_prob_continuous}
    \end{align*}
    where the action of operator $\mathcal{H}$ on function $\psi$ is given by  
    \begin{align*}
        \mathcal{H}\psi=\mathbb{E}_{m\sim \nu(m)}[d^{*}u(m;du(m; \psi))]. 
        \numberthis\label{eq:H_action_continuous}
    \end{align*}
    Here, $du(m;\psi)$ is the Gâteaux derivative of $u$ at $m\in V_h^{\text{in}}$ in the direction of $\psi\in V_h^{\text{in}}$, defined as $\lim_{\varepsilon \to 0}(u(m+\varepsilon\psi)-u(m))/\varepsilon$, and $d^{*}u(m;\cdot)$ is the adjoint of the operator $du(m;\cdot)$. When solved using finite element method, ~\cref{eq:asm_eigen_prob_continuous} is equivalent to the following linear system (See~\cref{sec:proof_asm_eigenproblem_equivalence} for the derivation)
    \begin{align*}
        H\bm{\psi}=\lambda C^{-1}\bm{\psi}, 
        \numberthis\label{eq:asm_eigen_prob_discrete}
    \end{align*}
    where the action of matrix $H$ on vector $\bm{\psi}$ is given by
    \begin{align*}
       H\bm{\psi}=
       \mathbb{E}_{\bm{m}\sim \nu(\bm{m})}[
       (\nabla_{\bm{m}}\bm{u})^{T}M^{\text{out}}(\nabla_{\bm{m}}\bm{u})\bm{\psi}
       ], 
       \numberthis\label{eq:H_action_discrete}
    \end{align*}
    and the action of matrix $C^{-1}$ on vector $\bm{\psi}$ is given by
    \begin{align*}
        C^{-1}\bm{\psi}=A^{\text{in}}(M^{\text{in}})^{-1}A^{\text{in}}\bm{\psi}.
        \numberthis\label{eq:C_inverse_action_discrete}
    \end{align*}
    Here, $M^{\text{out}}$ denotes the mass matrix of the output function space, i.e., $M^{\text{out}}_{ij}=\langle \phi_j^{\text{out}}, \phi_{i}^{\text{out}}\rangle$. In practice, when solving~\cref{eq:asm_eigen_prob_discrete}, we obtain its left hand side through computing 
    \begin{align*}
        (\langle \mathcal{H}\psi, \phi_1^{\text{in}}\rangle, \ldots, \langle \mathcal{H}\psi, \phi_{N_{h}^{\text{in}}}^{\text{in}}\rangle)^{T}
    \end{align*}
    and its right hand side through the matrix-vector multiplication in~\cref{eq:C_inverse_action_discrete} (See ~\cref{sec:proof_asm_eigenproblem_equivalence} for details).     Similar to the KLE case, let $\Psi^{\text{in}}$ denote the nodal values of $r$ dominant eigenfunctions. Since the eigenvectors $\bm{\psi}_i^{\text{in}}$ are $C^{-1}$-orthogonal, the reduced representation of $\bm{m}$ can be computed as $\widetilde{\bm{m}}=(\Psi^{\text{in}})^{T}C^{-1}\bm{m}$. We use a scalable \emph{double pass randomized algorithm}~\cite{villa2021hippylib} implemented in \href{https://hippylib.github.io/}{hIPPYlib} to solve the generalized eigenproblems~\cref{eq:kle_eigen_prob_discrete} and~\cref{eq:asm_eigen_prob_discrete}.

    To this end, we present the computation of the derivative label and the action of $\mathcal{H}$ as follows.

    \begin{theorem}
    \label{thm:h_action}
        Suppose the PDE in the general form of \eqref{eq:PDE} is well-posed with a unique solution map from the input function $m\in V^{\text{in}}$ to the output function $u\in V^{\text{out}}$ with dual $(V^{\text{out}})^{\prime}$. Suppose the PDE operator $\mathcal{R}:V^{\text{in}}\times V^{\text{out}}\to (V^{\text{out}})^{\prime}$ is differentiable with derivatives $\partial_{m}\mathcal{R}:V^{\text{in}}\to (V^{\text{out}})^{\prime}$ and $\partial_{u}\mathcal{R}: V^{\text{out}}\to (V^{\text{out}})^{\prime}$, and in addition $\partial_{u}\mathcal{R}$ is invertible with invertible adjoint $(\partial_{u}\mathcal{R})^{*}$. 
        Then the directional derivative $p = du(m;\psi)$ for any function $\psi\in V^{\text{in}}$, and an auxillary variable $q$ such that $d^{*}u(m;p) = -(\partial_m \mathcal{R})^{*}q$ can be obtained as the solution of the linearized PDEs
        \begin{align*}
            (\partial_u \mathcal{R})p+(\partial_m \mathcal{R})\psi&=0, \numberthis\label{eq:linear_PDE_1} \\
            (\partial_u\mathcal{R})^{*}q&=p.
            \numberthis\label{eq:linear_PDE_2}
        \end{align*}
    \end{theorem}

    \emph{Proof sketch.} We perturb $m$ with $\varepsilon \psi$ for any small $\varepsilon>0$ and obtain $\mathcal{R}(m+\varepsilon\psi, u(m+\varepsilon\psi))=0$. Using Taylor expansion to expand it to the first order and letting $\varepsilon$ approach $0$, we obtain~\cref{eq:linear_PDE_1}, where $p=du(m;\psi)$. Next we compute $d^{*}u(m;p)$. By~\cref{eq:linear_PDE_1}, we have $du(m;\psi)=-(\partial_u \mathcal{R})^{-1}(\partial_m \mathcal{R})\psi$. Thus, $d^{*}u(m;\psi)=-(\partial_m \mathcal{R})^{*}(\partial_u \mathcal{R})^{-*}p$. Then we can first solve for $q=(\partial_u \mathcal{R})^{-*}p$ and then compute $-(\partial_m \mathcal{R})^{*}q$. 
    See~\cref{sec:proof_thm_h_action} for a full proof.

\section{Experiments}
\label{sec:experiments}
In this section, we present experiment results to demonstrate the derivative-enhanced accuracy and cost of our method on two test problems of nonlinear vector-valued PDEs in comparison with DeepONet, FNO, and DINO. Details about the data generation and training can be found in~\cref{sec:experimental_details}.

    \subsection{Input probability measure}
    In all test cases, we assume that the input functions $m^{(i)}$ are i.i.d. samples drawn from a Gaussian random field with mean function $\bar{m}$ and covariance operator $(\delta I-\gamma\Delta)^{-\alpha}$. We take $\alpha = 2$ in two space dimensions so the covariance operator is of trace class \cite{stuart2010inverse}. It is worth noting that the parameters $\gamma$ and $\delta$ jointly control the marginal variance and correlation length, for which we take values to have large variation of the input samples that lead to large variation of the output PDE solutions. In such cases, especially when the mapping from the input to output is highly nonlinear as in our test examples, a vanilla neural network approximations tend to result in relatively large errors, particularly with a limited number of training data. To generate samples from this Gaussian random field, we use a scalable (in mesh resolution) sampling algorithm implemented in \href{https://hippylib.github.io/}{hIPPYlib}~\cite{villa2021hippylib}. 

    \subsection{Governing equations}
    We consider two nonlinear PDE examples, including a nonlinear vector-valued hyperelasticity equation with one state (displacement), and a nonlinear vector-valued Navier--Stokes equations with multiple states (velocity and pressure). For the hyperelasticity equation, we consider an experimental scenario where a square of hyperelasticity material is secured along its left edge while a fixed upward-right force is applied to its right edge~\cite{cao2023residual}. Our goal is to learn the map between (the logarithm of) the material's Young's modulus and its displacement. We also consider the Navier--Stokes equations that describes viscous, incompressible creeping flow. In particular, we consider the lid-driven cavity case where a square cavity consisting of three rigid walls with no-slip conditions and a lid moving with a tangential unit velocity. We consider that the uncertainty comes from the viscosity term and aim to predict the velocity field. See~\cref{sec:governing_equations} for details.

    \subsection{Evaluation metrics}
    \label{sec:evaluation_metrics}
    We use the following three metrics to evaluate the performance of different methods. For the approximations of the solution $u(m)$, we compute the relative error in the $L^2(\Omega)$ norm and $H^1(\Omega)$ norm on the test dataset $\mathcal{D}_{\text{test}}=\{(m^{(i)}, u^{(i)})\}_{i=1}^{N_{\text{test}}}$, that is, 
    \begin{align*}
        \frac{1}{N_{\text{test}}}\sum_{m^{(i)}\in \mathcal{D}_{\text{test}}}\frac{\|\hat{u}(m^{(i)};\theta)-u(m^{(i)})\|_{L^2(\Omega)}}{\|u(m^{(i)})\|_{L^2(\Omega)}}, 
        \quad 
        \frac{1}{N_{\text{test}}}\sum_{m^{(i)}\in \mathcal{D}_{\text{test}}}\frac{\|\hat{u}(m^{(i)};\theta)-u(m^{(i)})\|_{H^1(\Omega)}}{\|u(m^{(i)})\|_{H^1(\Omega)}},
    \end{align*}
    where 
    \begin{align*}
        \|u\|_{L^2(\Omega)}=\int_{\Omega}\|u(x)\|^2_2\;\mathrm{d}x=\bm{u}^{T}M^{\text{out}}\bm{u}, 
        \quad \|u\|_{H^1(\Omega)}=(\|u\|_{L^2(\Omega)}^2+\|\nabla u\|_{L^2(\Omega)}^2)^{1/2}.
    \end{align*}
   
    For the approximation of the Jacobian $du(m;\cdot)$, we compute the relative error (for the discrete Jacobian) in the Frobenius norm on $\mathcal{D}_{\text{test}}$ along random directions $\omega=\{\omega_i\}_{i=1}^{N_{\text{dir}}}$, that is, 
    \begin{align*}
        \frac{1}{|\mathcal{D}_{\text{test}}|}\sum_{m^{(i)}\in \mathcal{D}_{\text{test}}}
        \frac{\|d\hat{u}(m^{(i)};\omega)-du(m^{(i)};\omega)\|_F}{\|du(m^{(i)};\omega)\|_F}, 
    \end{align*}
    where $\omega_i$ are samples drawn from the same distribution as $m$. 

    \subsection{Main results}
    % We observe consistent performance in terms of enhanced approximation accuracy for the three different equations. 
    We compare the prediction errors measured in the above three evaluation metrics and computational cost in data generation and neural network training for four neural operator architectures, including DeepONet \cite{lu2021learning}, FNO \cite{li2020fourier}, DINO \cite{o2024derivative}, and our DE-DeepONet. We also add experiments to demonstrate the performance of the FNO trained with the derivative loss (DE-FNO) and the DeepONet trained with input dimension reduction but without the derivative loss (DeepONet (ASM) or DeepONet (KLE)). For FNO, we use additional position embedding \cite{kovachki2023neural} that improves its approximation accuracy in our test cases. For DINO, we use ASM basis (v.s.\ KLE basis) as the input reduced basis and POD basis (v.s.\ output ASM basis \cite{o2024derivative}) as the output reduced basis, which gives the best approximation accuracy. For the input reduced basis of DE-DeepONet, we also test and present the results for both KLE basis and ASM basis. 

    \begin{figure}[H]
        \begin{center}
        \centerline{\includegraphics[width=0.33\columnwidth]{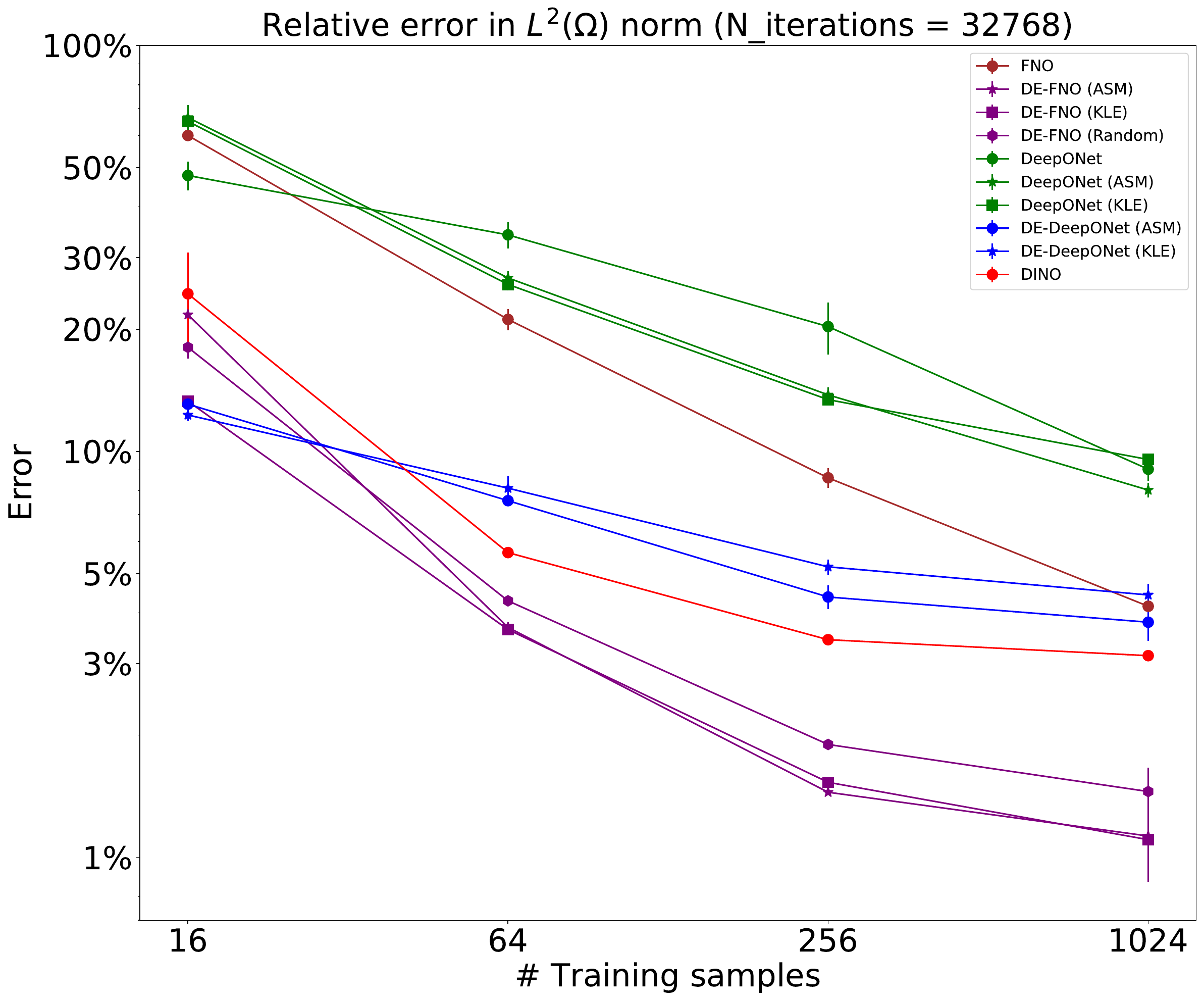}
                    \includegraphics[width=0.33\columnwidth]{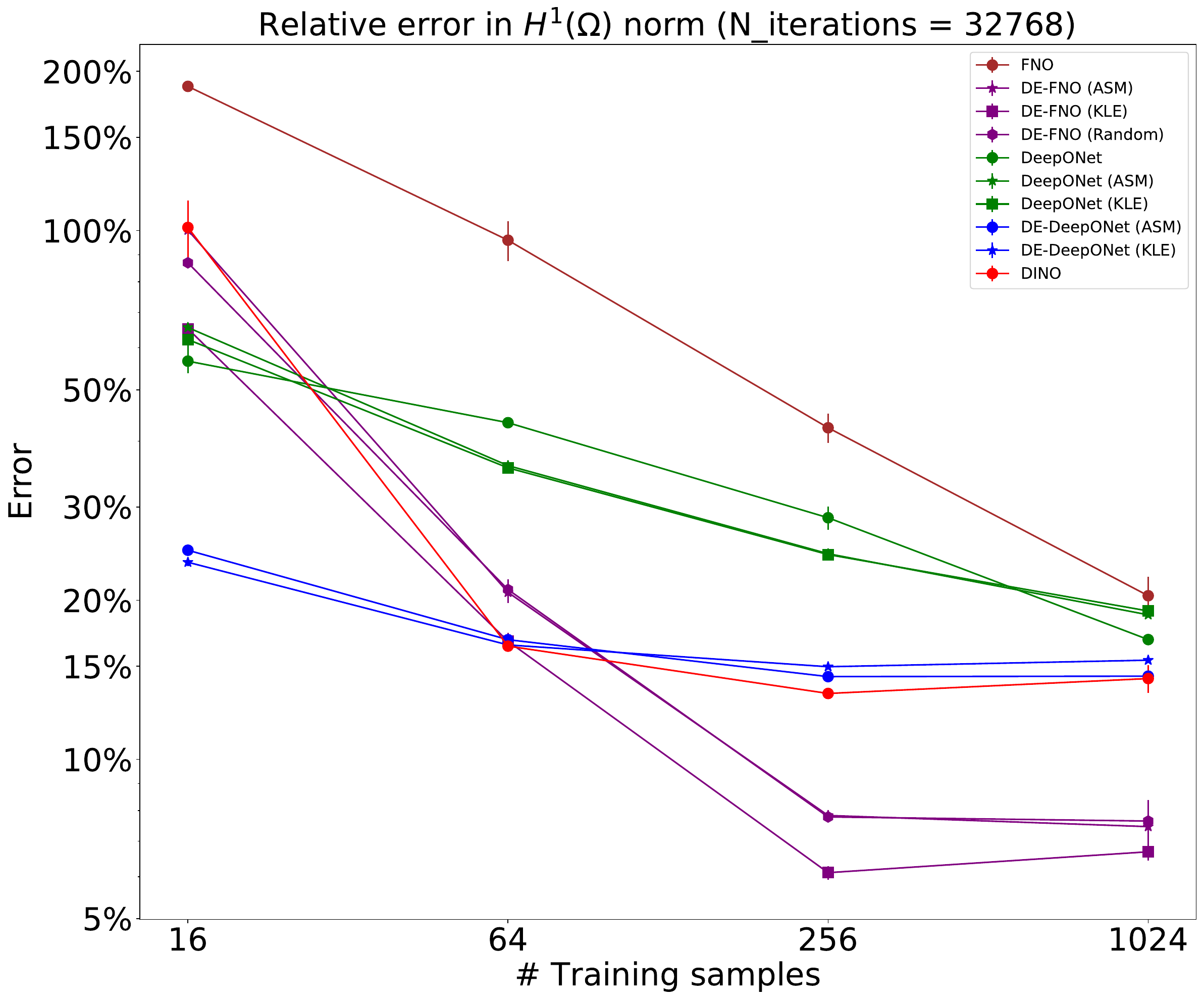}
                    \includegraphics[width=0.33\columnwidth]{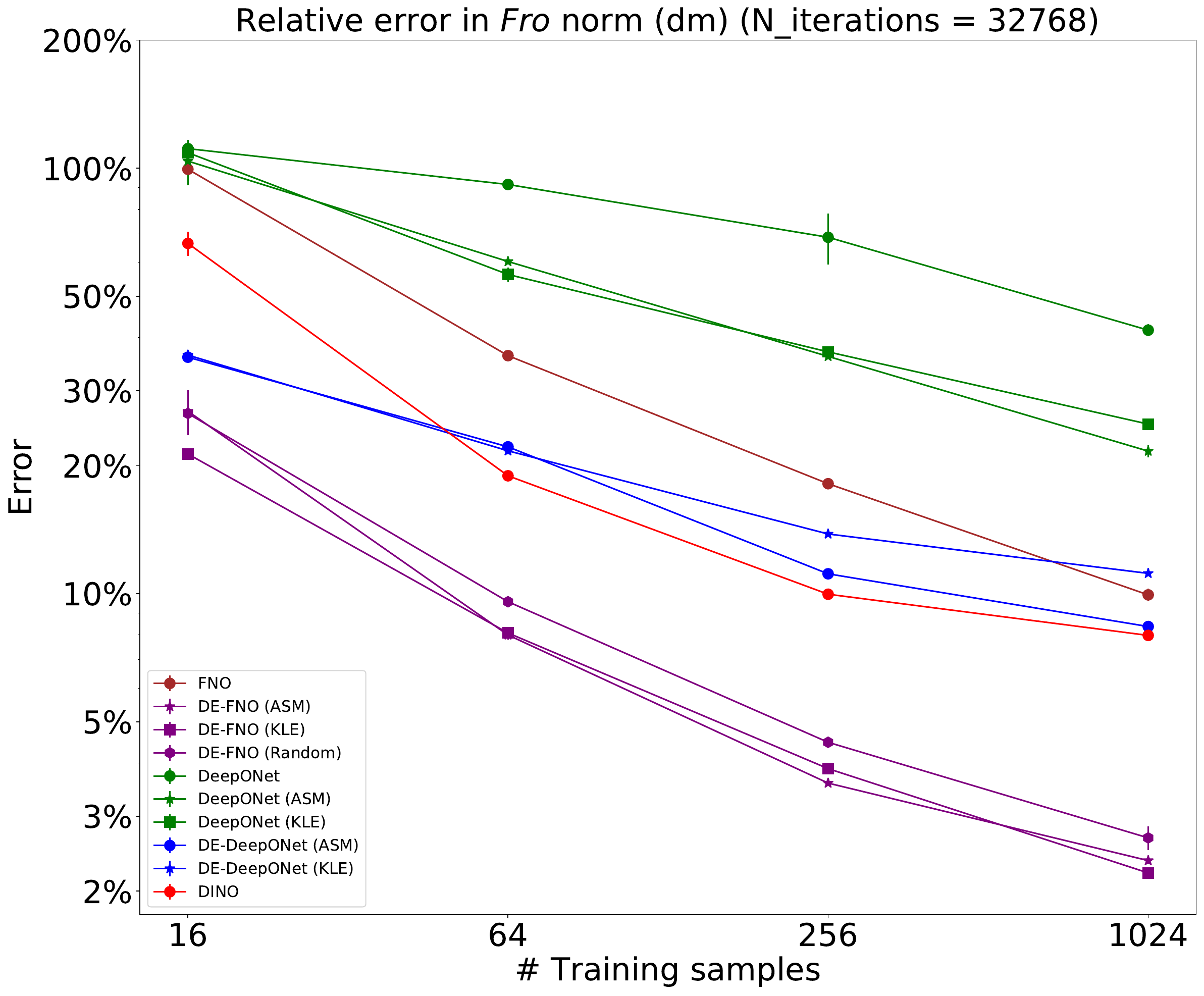}
        }
       \centerline{\includegraphics[width=0.33\columnwidth]{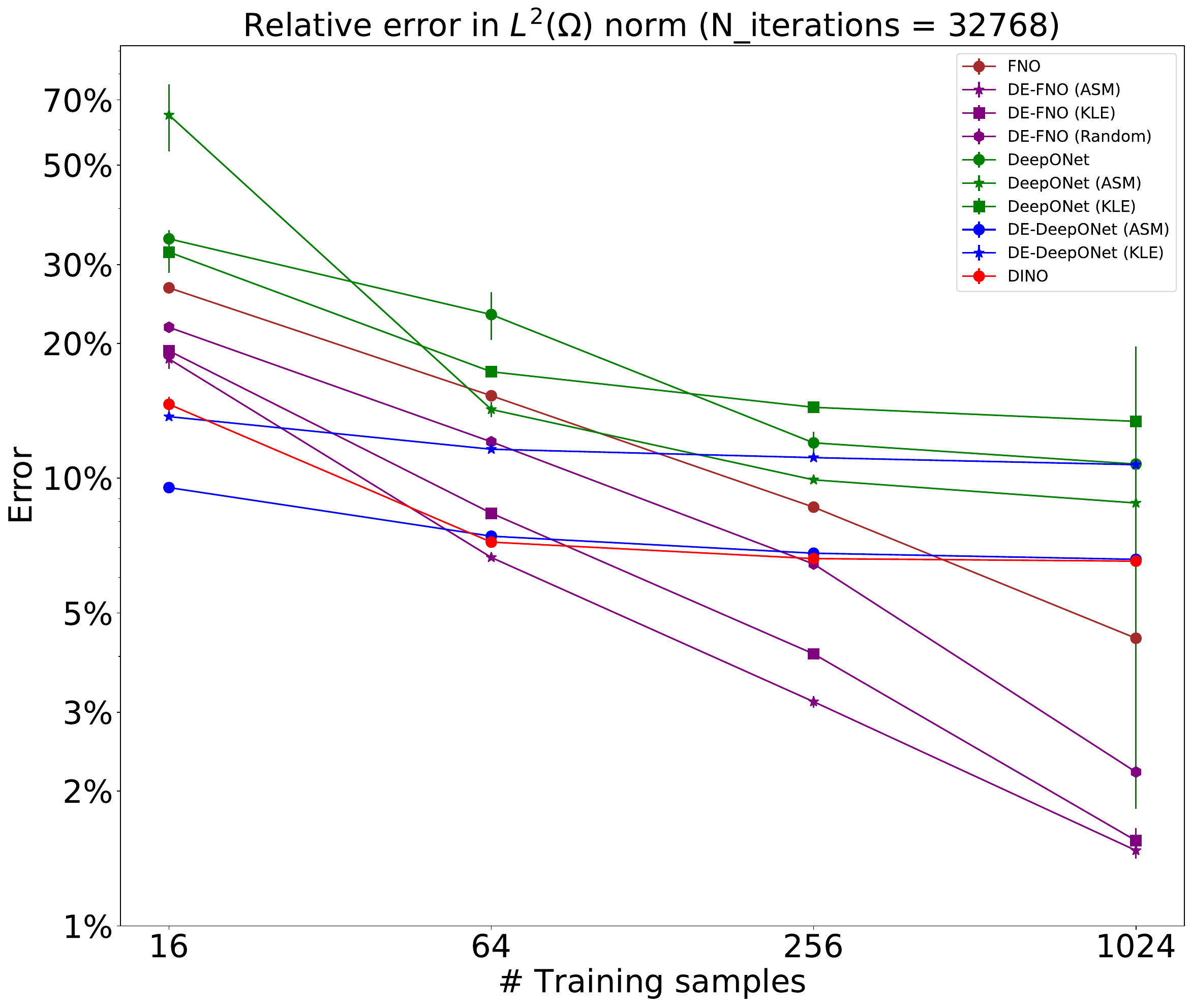}
                    \includegraphics[width=0.33\columnwidth]{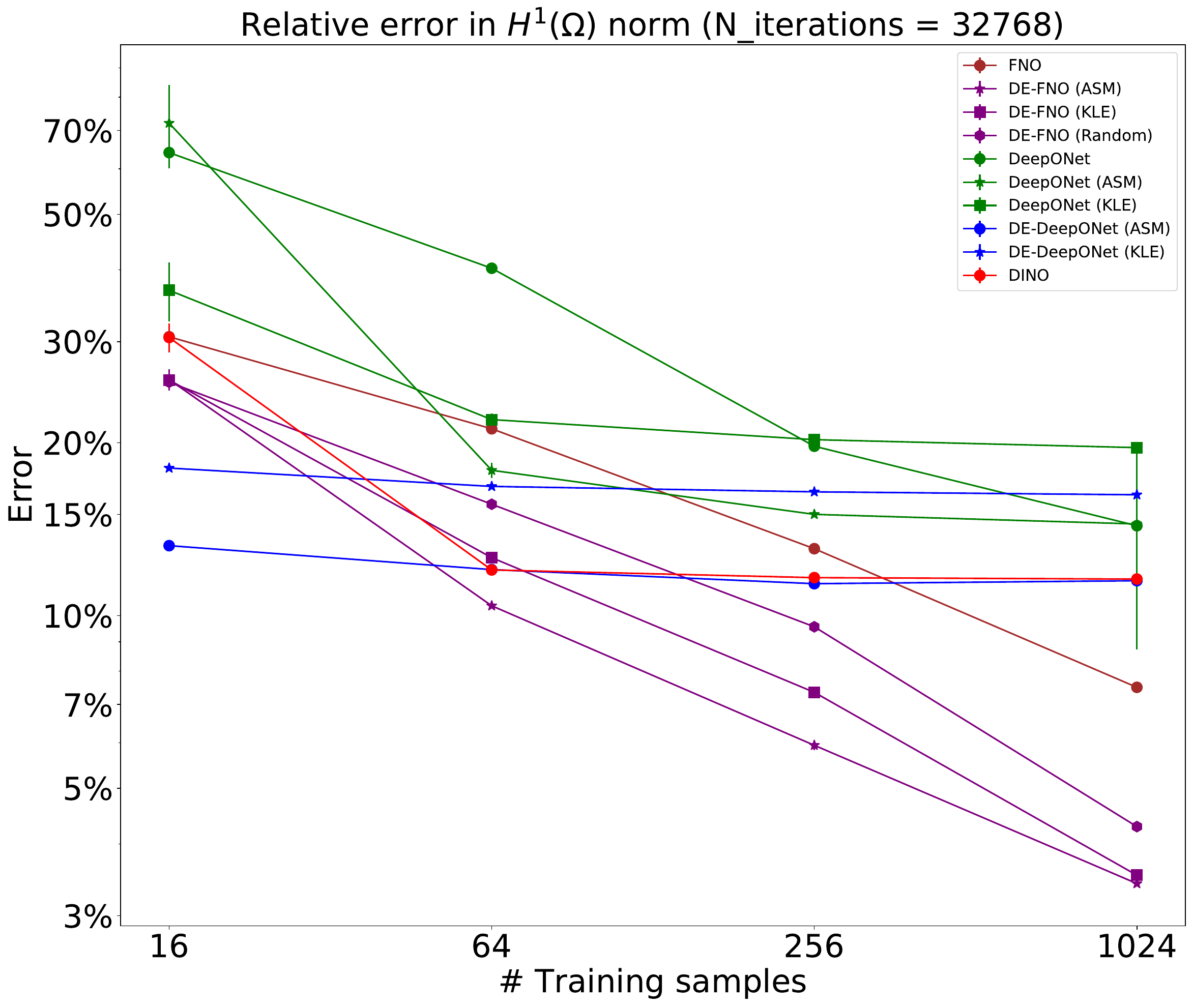}
                    \includegraphics[width=0.33\columnwidth]{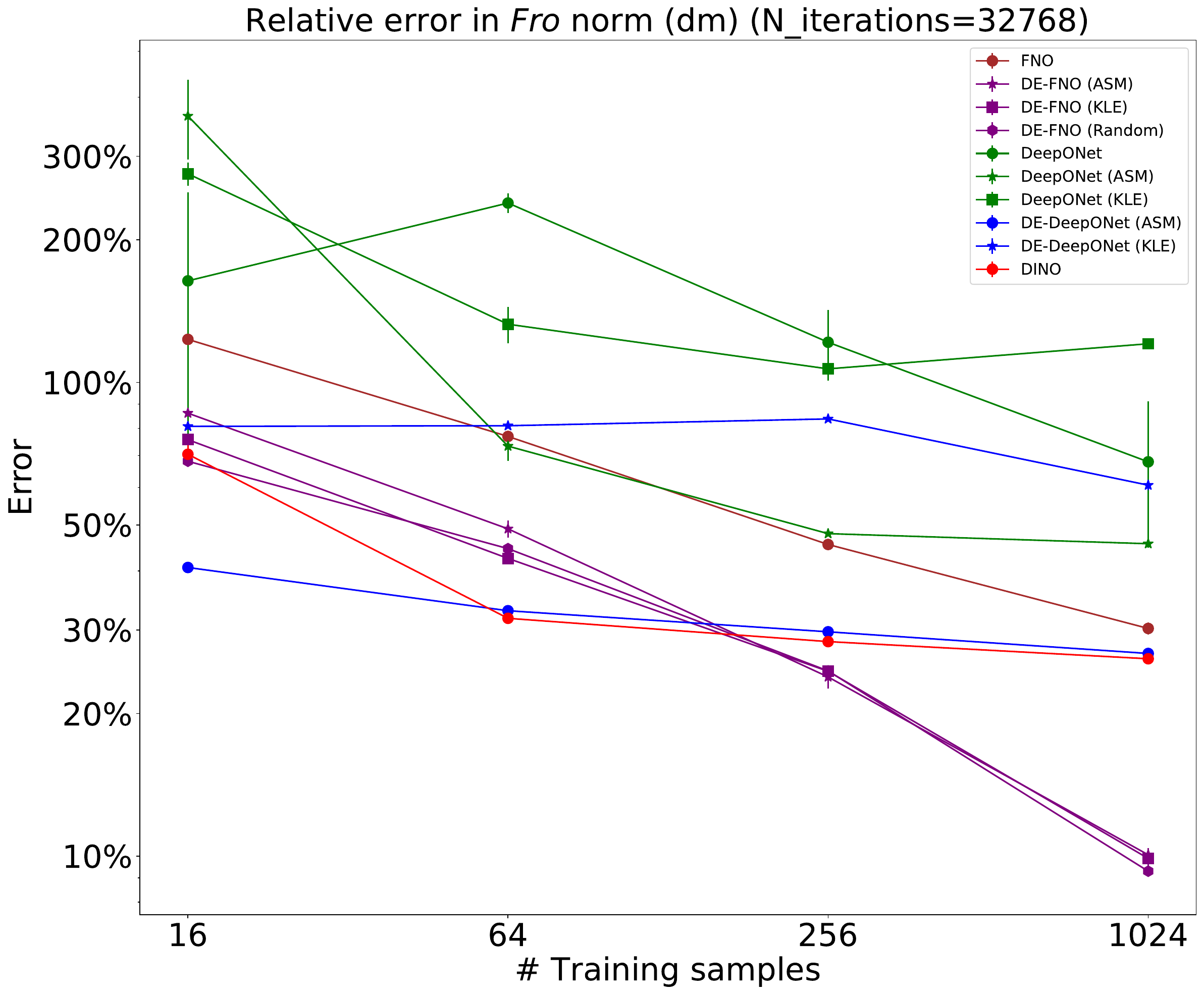}
        }        
        \caption{Mean relative errors ($\pm$ standard deviation) over 5 random seeds of neural network training for a varying number of training samples for the [top: hyperelasticity; bottom: Navier--Stokes] equation using different methods. Relative errors in the $L^2(\Omega)$ norm (left) and $H^1({\Omega})$ norm (middle) for the prediction of $u=(u_1,u_2)$. Right: Relative error in the Frobenius (Fro) norm for the prediction of $du(m;\omega)=(du_1(m;\omega), du_2(m;\omega))$.
        }
    \label{fig:error_plot_combined}
    \end{center}
    \vskip -2em
    \end{figure}

    \begin{figure}[H]
        \begin{center}
        \centerline{\includegraphics[width=0.33\columnwidth]{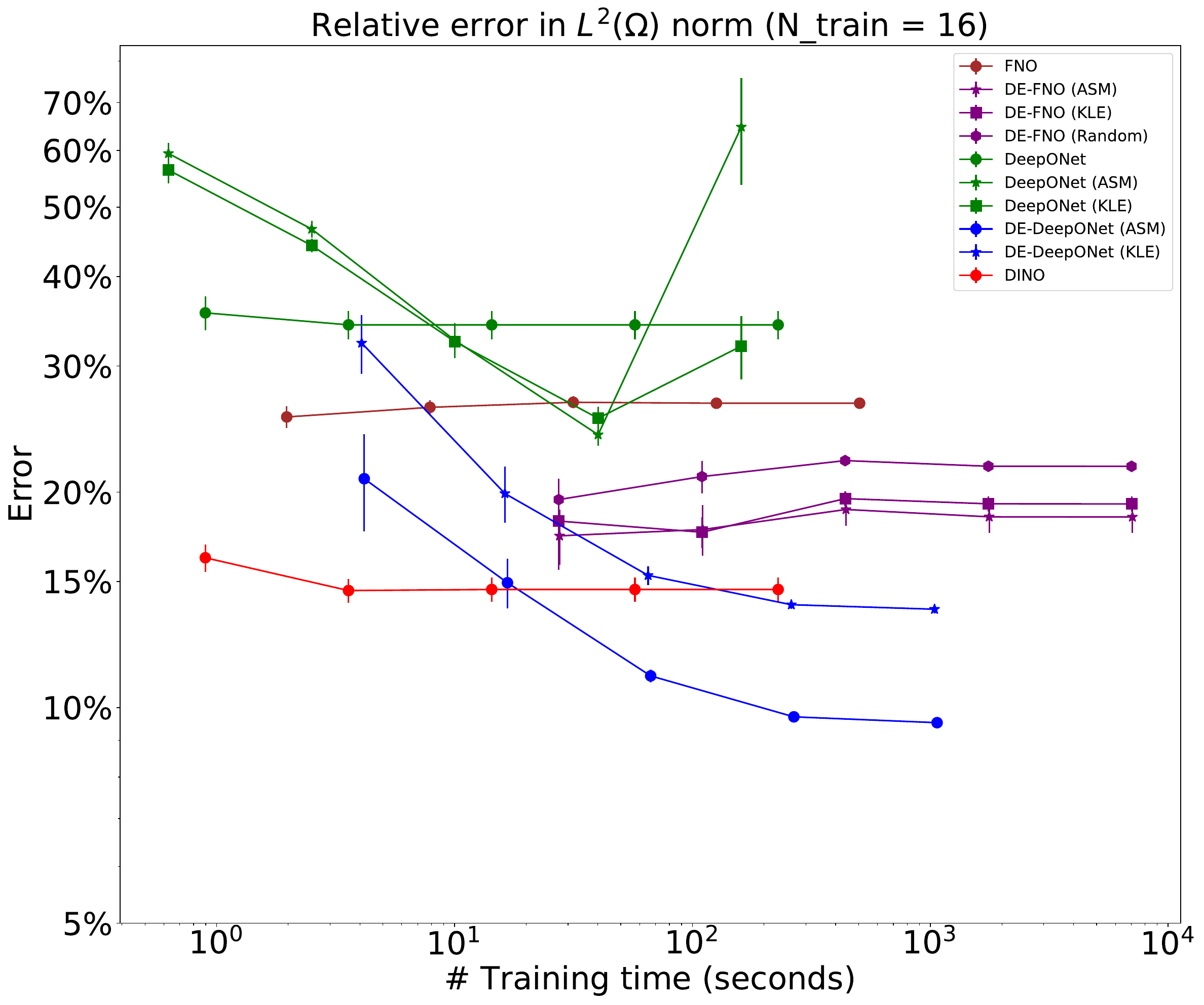}
                    \includegraphics[width=0.33\columnwidth]{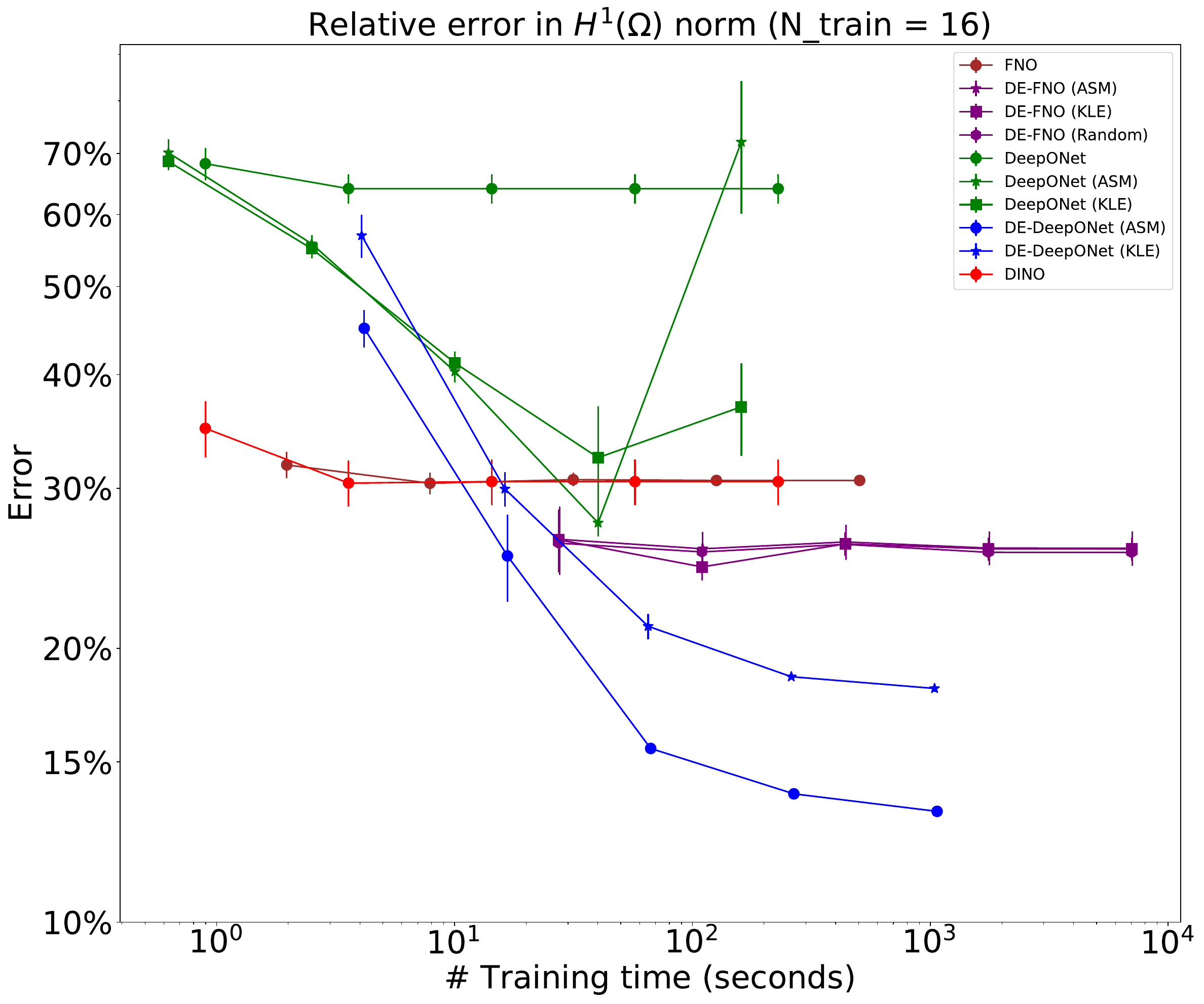}
                    \includegraphics[width=0.33\columnwidth]{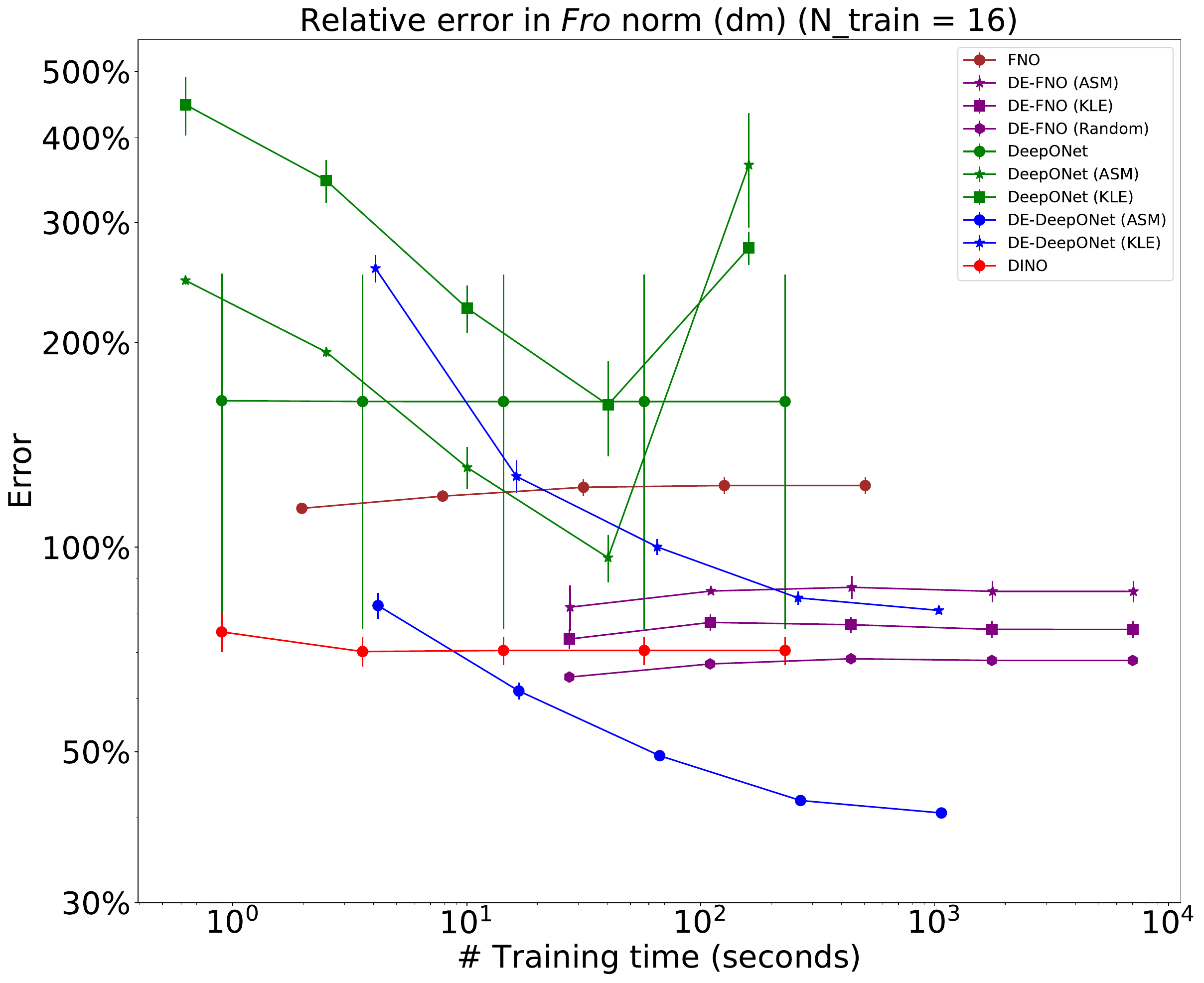}
        }
        \centerline{\includegraphics[width=0.33\columnwidth]{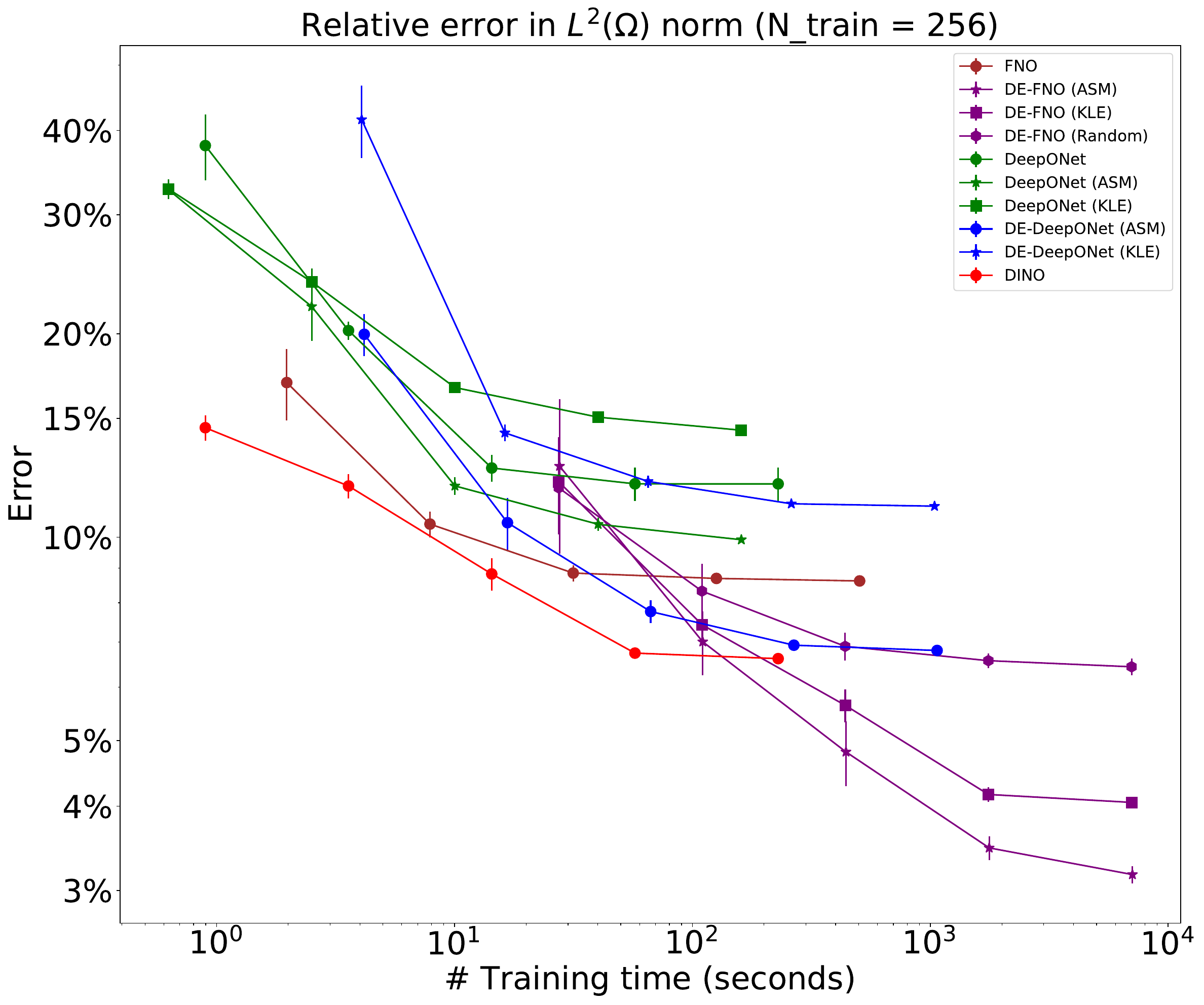}
                    \includegraphics[width=0.33\columnwidth]{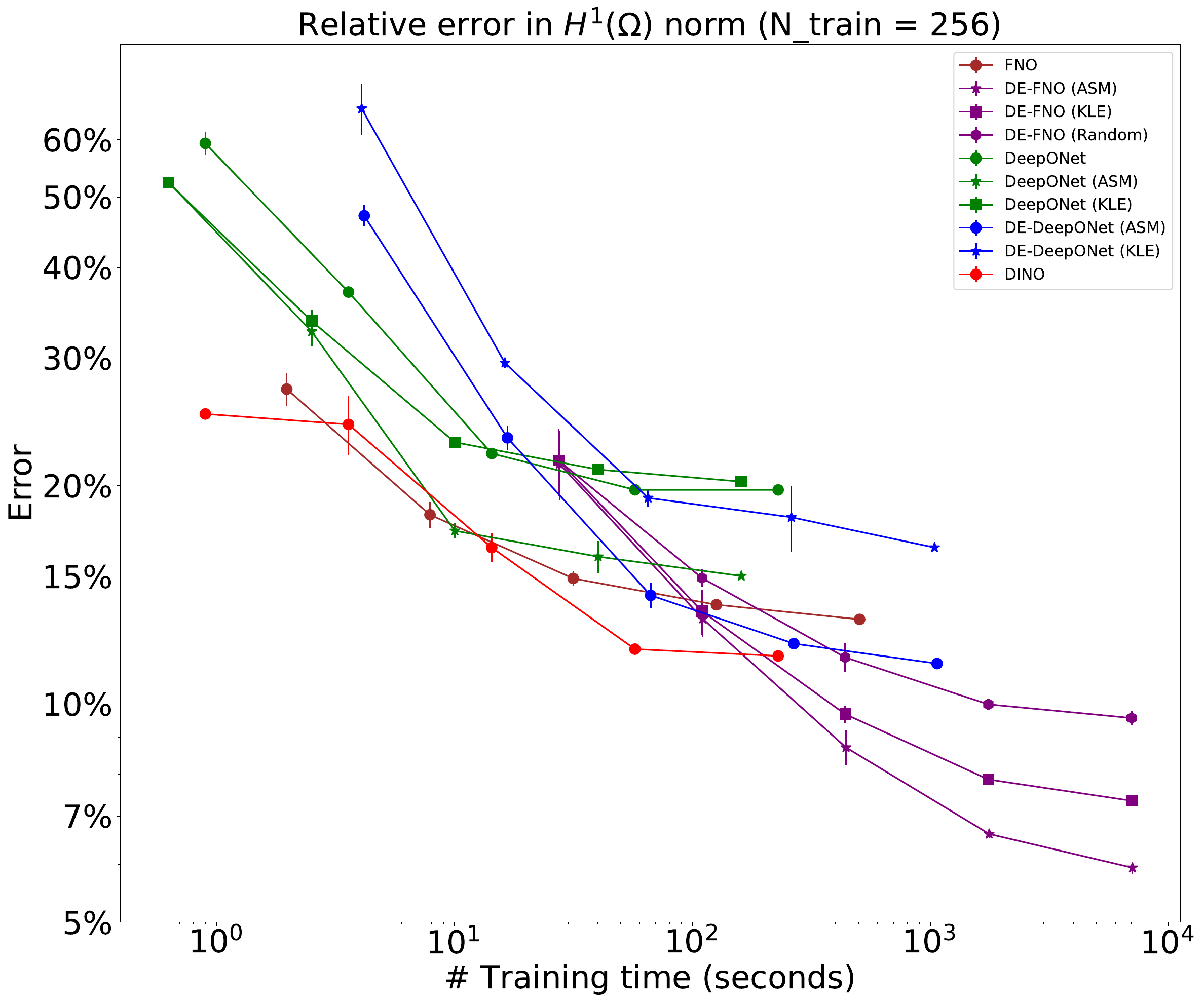}
                    \includegraphics[width=0.33\columnwidth]{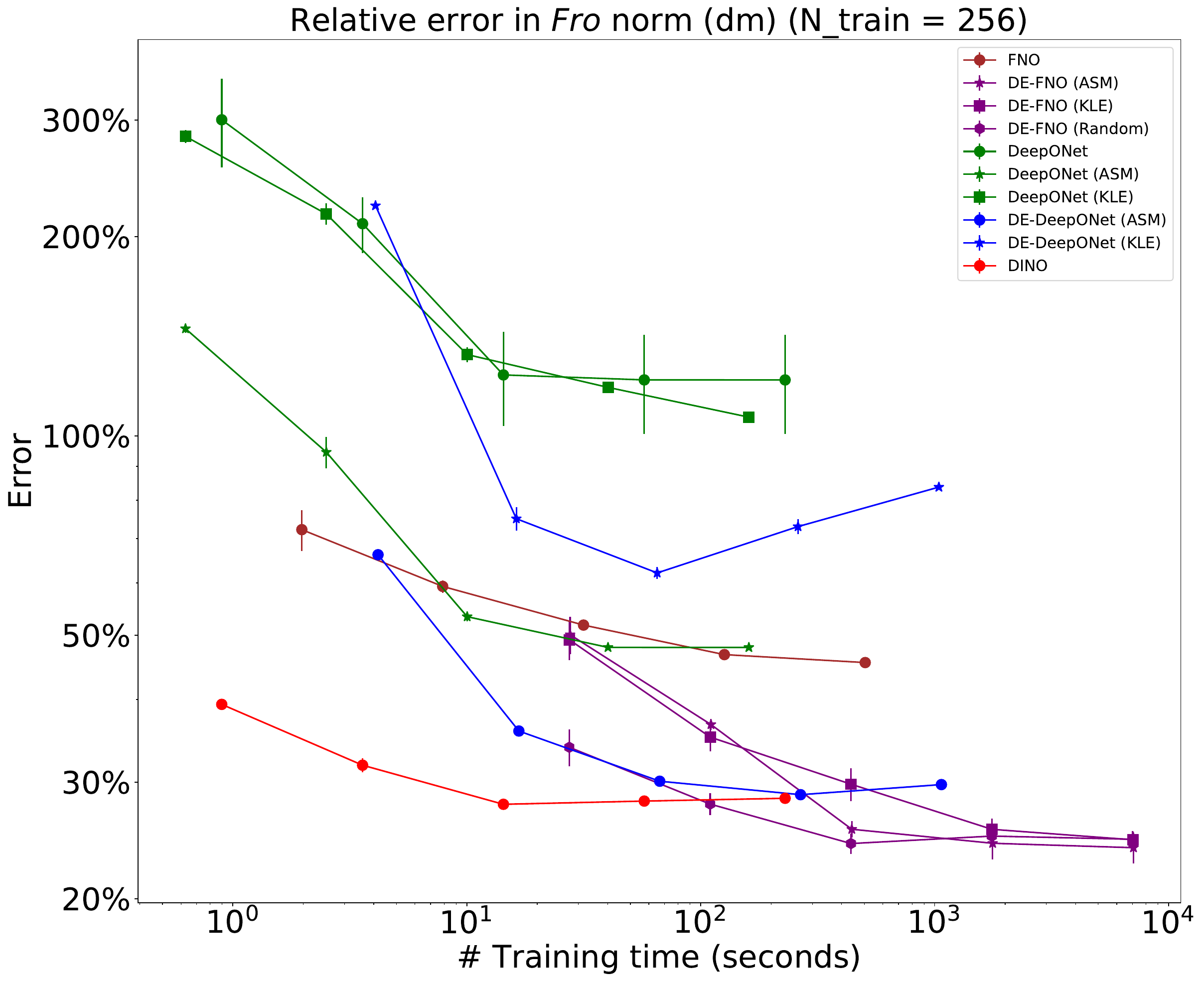}
        }       
        \caption{Mean relative errors ($\pm$ standard deviation) over 5 random seeds versus model training time for the Navier--Stokes equations when the number of training samples is [top: 16; bottom: 256].
        }
    \label{fig:error_plot_navier_stokes_convergence_16_and_256}
    \end{center}
    \vskip -2em
    \end{figure}

    \paragraph{Test errors. }In~\cref{fig:error_plot_combined}, we show a comparison of different methods for the hyperelasticity and Navier--Stokes problem in predicting the solution (in $L^2$ norm, left, and $H^1$ norm, middle) and its derivative (in Frobenius norm, right) with respect to parameter in the direction of $N_{\text{dir}}=128$ test samples $\{\omega_i\}_{i=1}^{N_{\text{dir}}}$. First, we can observe significant improvement of the approximation accuracy of DE-DeepONet compared to the vanilla DeepONet for all three metrics in all cases of training samples. Second, we can see that FNO leads to larger approximation errors than DINO and DE-DeepONet for all metrics except when number of training samples is 1024. We believe that the reason why FNO performs better than DE-DeepONet and DINO when training samples are large enough is mainly due to the use of input dimensionality reduction in DE-DeepONet and DINO (where the linear reduction error cannot be eliminated by increasing training samples) whereas in FNO we use full inputs. We also see that DE-FNO performs the best among all models when the training samples are sufficient (256 or 1024), although in the compensation of much longer training time shown in~\cref{fig:error_plot_navier_stokes_convergence_16_and_256}. Third, we can see that the approximation accuracy of DINO is similar to DE-DeepONet (ASM) but requires much longer inference time as shown in~\cref{tab:computational_time_model_inference} (see~\cref{sec:related work} and~\cref{subsubsec:derivative_outputs_of_neural_networks} for reasons). In DINO, the output reduced basis dimension is set to be smaller than or equal to the number of training samples as the output POD basis are computed from these samples, i.e., 16 for 16 samples and 64 for $\geq 64$ samples. Increasing the output dimension beyond 64 does not lead to smaller errors in our test. Finally, we can observe that DE-DeepONet using ASM basis leads to smaller errors than using KLE basis, especially for the Navier-Stokes problem.

    \begin{table}[H]
        \begin{minipage}{0.3\textwidth} 
            \caption{\small Output reconstruction error with 16 input reduced bases}\label{tab:output_reconstruction_error_16}
            \vskip 0.1in
            \begin{center}
            \begin{scriptsize}
                \begin{tabular}{@{}ccc@{}}
                \toprule
                \multirow{4}{*}{\( \text{Dataset}  \)} & \multicolumn{2}{c}{\(\text{Relative $L^2$ error} \)} \\
                \cmidrule(l){2-3}
                                  & KLE    & ASM   \\ 
                \midrule            
                Hyperelasticity    & 3.8\%  & 2.7\%     \\
                Navier--Stokes     & 17.4\% & 5.8\%   \\
                \bottomrule
                \end{tabular}
            \end{scriptsize}        
            \end{center}
            \vskip -0.1in
        \end{minipage}
        \hfill
        \begin{minipage}{0.65\textwidth} % Adjust the width to control space for the text
            Moreover, we present the output reconstruction error due to the input dimension reduction using KLE or ASM basis in~
            \cref{tab:output_reconstruction_error_16}.  The errors provide the lower bound of the relative errors in $L^2(\Omega)$ norm of DINO and DE-DeepONet. We can see that using the ASM basis results in a lower output reconstruction error than the KLE basis (more significant difference observed in the more nonlinear Navier--Stokes equations). See~\cref{sec:output_reconstruction_error} for the decay of the reconstruction error with increasing number of reduced basis. 
        \end{minipage}
    \end{table}
    
    In addition, we provide full visualizations of the ground truth and prediction of both solution and derivatives in~\cref{sec:visualization}. Visually, DE-DeepONet (ASM) consistently provides the best estimate (in terms of pattern and smoothness similarity with the ground truth) for both the solution and its derivative when the training data is limited (16 or 64 samples).
    
    \paragraph{Data generation computational cost.}  We use MPI and the finite element library FEniCS \cite{logg2012automated} to distribute the computational load of offline data generation to 64 processes for the PDE models considered in this work. See~\cref{tab:computational_time_data_generation_whole} for the wall clock time in generating the samples of Gaussian random fields (GRFs), solving the PDEs, computing the $r = 16$ reduced basis functions (KLE or ASM) corresponding to the $16$ largest eigenvalues, generating the derivative labels, respectively. In~\cref{tab:computational_time_data_generation_different_number_samples}, We also provide the total wall clock time of data generation of DE-DeepONet (ASM) (we only includes the major parts -- computing high fidelity solution, ASM basis and dm labels [16 directions]) when $N_\text{train} = 16, 64, 256, 1024$ using 16 CPU processors.

   \begin{table}[H]
    \caption{\small Wall clock time (in seconds) for data generation on 2 $\times$ AMD EPYC 7543 32-Core Processors}
    \label{tab:computational_time_data_generation_whole}
    \vskip 0.1in
    \begin{center}
    \begin{scriptsize}
        \begin{tabular}{@{}cccccc@{}}
        \toprule
        \multirow{4}{*}{\( \text{Dataset} \)} & \multicolumn{5}{c}{\( \text{Process} \)} \\
        \cmidrule(l){2-6}
                          &GRFs                          & PDEs                     & KLE                    & ASM                       & dm labels                        \\ 
                          &$(N_{\text{all}}=2000)$       & $(N_{\text{all}}=2000)$  & $(r=16)$               & $(r=16)$                  & $(N_{\text{all}}=2000, r=16)$    \\ 
                          & ($64$ procs)                 & ($64$ procs)             & ($1$ procs)            & ($16$ procs)              & ($64$ procs)                     \\ 
        \midrule
        Hyperelasticity   & $1.1$                        & $9.7$                    & $0.4$                  & $1.4$                     & $19.5$                           \\
        Navier--Stokes    & $1.9$                        & $99.1$                   & $1.3$                  & $9.7$                     & $125.5$                          \\
        \bottomrule
        \end{tabular}
    \end{scriptsize}        
    \end{center}
    \vskip -0.1in
    \end{table}

   \begin{table}[H]
    \begin{minipage}{0.4\textwidth}
    \caption{\small Wall clock time (in seconds) for data generation with different number of training samples using 16 CPU processors}
    \label{tab:computational_time_data_generation_different_number_samples}
    \vskip 0.1in
    \begin{center}
    \begin{scriptsize}
        \begin{tabular}{@{}ccccc@{}}
        \toprule
        \multirow{4}{*}{\( \text{Dataset} \textbackslash N_{\text{train}} \)} & \multicolumn{4}{c}{\(\text{PDEs + ASM basis + dm labels} \)} \\
        \cmidrule(l){2-5}
                          & 16                           & 64                       & 256                    & 1024                       \\ 
        \midrule
        Hyperelasticity   & 2                            & 5                        & 16                      & 61                    \\
        Navier--Stokes    & 17                           & 38                       & 124                     & 470                     \\
        \bottomrule
        \end{tabular}
    \end{scriptsize}        
    \end{center}
    \end{minipage}
    \hfill
    \begin{minipage}{0.58\textwidth}
       \caption{\small Wall clock time (seconds/iteration with batch size 8) for training on a single NVIDIA RTX A6000~GPU}
        \label{tab:computational_time_neural_network_training}
        \vskip 0.1in
        \begin{center}
        \begin{scriptsize}
            \begin{tabular}{@{}cccccc@{}}
            \toprule
            \multirow{2}{*}{\( \text{Dataset} \)} & \multicolumn{5}{c}{\( \text{Model} \)} \\
            \cmidrule(l){2-6}
                               & DeepONet       & FNO           & DINO         & DE-FNO   & DE-DeepONet        \\ 
            \midrule 
            Hyperelasticity    & 0.007          & 0.015         & 0.007        &  0.215   & 0.022           \\
            Navier--Stokes     & 0.007          & 0.015         & 0.007        &  0.216   & 0.033             \\
            \bottomrule
            \end{tabular}
        \end{scriptsize}
    \end{center}
    \end{minipage}
    \vskip -0.1in
    \end{table}

    \begin{table}[H]
    \caption{\small Total wall clock time (in seconds) for each model inferring on 500 test samples of both the solution and dm in 128 random directions, using a single GPU and a single CPU (except where specified)}
    \label{tab:computational_time_model_inference}
    \vskip -0.1in
    \begin{center}
    \begin{scriptsize}
        \begin{tabular}{@{}cccccc@{}}
        \toprule
        \multirow{4}{*}{\( \text{Dataset} \)} & \multicolumn{5}{c}{\(\text{Model} \)} \\
        \cmidrule(l){2-6}
                          & DeepONet                     & FNO/DE-FNO                &  DINO$^1$        & DE-DeepONet         & Numerical solver        \\ 
                          &                              &                           &  1 GPU + 1 CPU/16 CPUs       &                   &      0 GPU + 16 CPUs            \\ 
        \midrule
        Hyperelasticity   & 3                            & 33                        & 69/7                       & 10            &  166     \\
        Navier--Stokes    & 3                            & 33                        & 2152/151                    & 18              & 1103    \\
        \bottomrule
        \end{tabular}
    \end{scriptsize}        
    \end{center}
    \footnotesize{$^1$ The inference time of DINO is dominated by the time required to compute evaluations of all finite element basis functions at the grid points using FEniCS (which may not be the most efficient, see~\cref{subsubsec:derivative_outputs_of_neural_networks}). Even though these grid points overlap with parts of the finite element nodes—allowing us to skip evaluations by extracting the relevant nodes—for a fairer comparison with DE-DeepONet (in terms of its ability to evaluate at any arbitrary point), we assume they are arbitrary points requiring explicit evaluation.}
    \vskip -0.1in
    \end{table}

    \paragraph{Model training computational cost.} We present comparisons of the wall clock time of each optimization iteration (with batch size 8) of different methods in~\cref{tab:computational_time_neural_network_training} and convergence plot (error versus training time) in~\cref{fig:error_plot_navier_stokes_convergence_16_and_256} and the figures in~\cref{sec:convergence_plot}. We find that incorporating derivative loss leads to longer training time as expected. However, when the training data are limited, the increased computation cost is compensated for a significant reduction of errors. We note that there are potential ways to further reduce the training cost, e.g., by training the model with additional derivative loss only during the later stage of training, or by using fewer points for computing the derivative losses in each iteration. Additionally, thanks to the dimension reduction of the input, we can define a relatively small neural network and thus are able to efficiently compute the derivatives using automatic differentiation.

\section{Related work}
\label{sec:related work}
    Our work is related to Sobolev training for neural networks \cite{czarnecki2017sobolev}, which was found to be effective in their application to model distillation/compression and meta-optimization. In the domain of surrogate models for parametric partial differential equations, our work is more closely related to derivative-informed neural operator (DINO)~\cite{o2024derivative} which is based on a derivative-informed projected neural network (DIPNet)~\cite{o2022derivative}, and presents an extension to enhance the performance of the DeepONet. Compared to DINO, although the DeepONet architecture (and its formulation of dm loss) requires longer training time, it offers the following advantages: (1) Potentially shorter inference time. The additional trunk net (which receives spatial coordinates) allows us to quickly query the sensitivity of output function at any point when input function is perturbed in any direction. While DINO can only provide the derivative of the output coefficients respect to the input coefficients (we call reduced dm), in order to compute the sensitivity at a batch of points, we need to post process the reduced dm by querying the finite element basis on these points and computing large matrix multiplications; (2) Greater flexibility and potential for improvements. Although both DeepONet and DINO approximate solution by a linear combination of a small set of functions, these functions together in DeepONet is essentially the trunk net, which is "optimized" via model training, whereas in DINO, they are POD or derivative-informed basis precomputed on training samples. When using DINO, if we encounter a case where the training samples not enough to accurately compute the output basis, the large approximation error between the linear subspace and solution manifold will greatly restrict the model prediction accuracy (see~\cref{fig:error_plot_navier_stokes_convergence_16_and_256} when $N_{\text{train}}=16$). And the reduced dm labels only supports linear reduction of output. However, it is possible that we can further improve DeepONet by, e.g., adding physical losses (to enhance generalization performance) and Fourier feature embeddings (to learn high-frequency components more effectively) on the trunk net~\cite{wang2021learning} and replacing the inner product of the outputs of two networks by more flexible operations~\cite{pan2023neural,hao2023gnot} (to enhance expressive power). The dm loss formulation of our work is broadly suitable any network architecture that has multiple subnetworks, where at least one of them receives high-dimensional inputs.

\section{Discussion}
\label{sec:discussion}
    In this work, we proposed a new neural operator--Derivative-enhanced Deep Operator Network (DE-DeepONet) to address the limited accuracy of DeepONet in both function and derivative approximations. Specifically, DE-DeepONet employs a derivative-informed reduced representation of input function and incorporates additional loss into the loss function for the supervised learning of the derivative of the output with respect to the inputs of the branch net. Our experiments for nonlinear PDE problems with high variations in both input and output functions demonstrate that adding this loss term to the loss function greatly enhances the accuracy of both function and derivative approximations, especially when the training data are limited. We also demonstrate that the use of derivative loss can be extended to enhance other neural operators, such as the Fourier neural operator.

    We presented matrix-free computation of the derivative label and the derivative-informed dimension reduction for a general form of PDE problems by using randomized algorithms and linearized PDE solves. Thanks to this scalable approach, the computational cost in generating the derivative label data is shown to be only marginally higher than generating the input-output function pairs for the test problems, especially for the more complex Navier--Stokes equations which require more iterations in the nonlinear solves than the hyperelasticity equation. 
   
    \paragraph{Limitations:} We require the derivative information in the training and dimension reduction using ASM, which may not be available if the explicit form of the PDE is unknown or if the simulation only provides input-output pairs from some legacy code. Another limitation is that dimension reduction of the input function plays a key role in scalable data generation and training, which may not be feasible or accurate for intrinsically very high-dimensional problems such as high frequency wave equations. Such problems are also very challenging and remain unsolved by other methods to our knowledge.

\clearpage
\section*{Acknowledgements}
We would like to thank Jinwoo Go, Dingcheng Luo and Lianghao Cao for insightful discussions and helpful feedback.

\bibliographystyle{ieeetr}
\bibliography{reference}
\clearpage

%%%%%%%%%%%%%%%%%%%%%%%%%%%%%%%%%%%%%%%%%%%%%%%%%%%%%%%%%%%%

\appendix
\section{Proofs}
\label{sec:proofs}

    \subsection{Proof of ~\cref{thm:h_action}}
    \label{sec:proof_thm_h_action}
    We first show how to compute $p$ given the PDE residual $\mathcal{R}(m,u)=0$. Since $u$ is uniquely determined by $m$, we can write $u=u(m)$ so that $\mathcal{R}(m, u(m))\equiv 0$ holds for any $m\in V^{\text{in}}$. Thus, for any $\varepsilon>0$ and $\psi\in V^{\text{in}}$, we have  
    \begin{align*}
        \mathcal{R}(m+\varepsilon\psi, u(m+\varepsilon\psi))=0.
    \end{align*}
    Using the Taylor expansion we obtain
    \begin{align*}
        (\partial_m \mathcal{R}(m,u(m)))\varepsilon\psi +(\partial_u \mathcal{R}(m,u(m)))\delta u \approx 0,
        \numberthis\label{eq:Taylor_expansion}
    \end{align*}
    where $\delta u=u(m+\varepsilon\psi)-u(m)$. Dividing both sides of \cref{eq:Taylor_expansion} by $\varepsilon$ and letting $\varepsilon$ approach~$0$ yields
    \begin{align*}
        (\partial_m \mathcal{R}(m,u(m)))\psi+(\partial_u \mathcal{R}(m,u(m)))du(m;\psi)=0.
    \end{align*}
    For ease of notation, we write
    \begin{align*}
       \partial_m \mathcal{R}= \partial_m \mathcal{R}(m,u(m)), 
       \quad \partial_u \mathcal{R}=\partial_u \mathcal{R}(m, u(m)). 
    \end{align*}
    Then $p$ is the solution to the linear PDE
    \begin{align*}
        (\partial_m \mathcal{R})\psi+ (\partial_u \mathcal{R})p=0.
        \numberthis\label{eq:p_pde_strong_form}
    \end{align*}
    We solve~\cref{eq:p_pde_strong_form} via its weak form
    \begin{align*}
        \langle (\partial_m \mathcal{R})\psi, v \rangle + \langle (\partial_u \mathcal{R})p, v \rangle = 0, 
        \numberthis\label{eq:p_pde_weak_form}
    \end{align*}
    where $v$ is a test function in $V^{\text{out}}$.

    Next we show how to compute $d^{*}u(m;p)$. By~\cref{eq:p_pde_strong_form}, we have $du(m;\psi)=-(\partial_u \mathcal{R})^{-1}(\partial_m \mathcal{R})\psi$. Thus, 
    \begin{align}
    \label{eq:r_explicit}
        \begin{split}
        w:=d^{*}u(m;p)
        &=(-(\partial_u \mathcal{R})^{-1}(\partial_m \mathcal{R}))^{*}p\\
        &=-(\partial_m \mathcal{R})^{*}(\partial_u \mathcal{R})^{-*}p.
        \end{split}
    \end{align}
    Let $q:=(\partial_u \mathcal{R})^{-*}p$. We can solve for $q$ via  the weak form
    \begin{align*}
        \langle (\partial_u \mathcal{R})^{*}q, v\rangle = \langle p, v\rangle,
    \end{align*}
    or equivalently, 
    \begin{align*}
        \langle q, (\partial_u \mathcal{R})v\rangle = \langle p, v\rangle,
        \numberthis\label{eq:q_pde_weak_form}
    \end{align*}
    where $v$ is a test function in $V^{\text{out}}$.
    
    By~\cref{eq:r_explicit}, we have $w=-(\partial_m \mathcal{R})^{*}q$. For any test function $v\in V^{\text{in}}$, it holds that 
    \begin{align}
    \label{eq:r_weak_form}
        \begin{split}
            \langle w, v \rangle 
            &=\langle-(\partial_m \mathcal{R})^{*}q, v\rangle\\
            &=\langle q, -(\partial_m \mathcal{R})v\rangle.
        \end{split}
    \end{align}
     Note that we do not need to solve for $w$ explicitly; we only compute $\langle w, v\rangle$ with $v$ as the finite element basis functions $\phi_1^{\text{in}}, \ldots, \phi_{N_h^{\text{in}}}^{\text{in}}$. The cost of computing the right hand side of \cref{eq:r_weak_form} arises from evaluating the directional derivative and its inner product with the finite element basis functions.

    \begin{remark}
        By~\cref{eq:H_action_continuous}, we use $N_{\text{grad}}$ samples~$\{(m^{(i)}, u^{(i)})\}_{i=1}^{N_{\text{grad}}}$ to compute the Monte Carlo estimate of the action of operator $\mathcal{H}$ on any function $\psi\in V^{\text{in}}$, that is, 
        \begin{align*}
            \mathcal{H}\psi\approx 
            \frac{1}{N_{\text{grad}}}\sum_{i=1}^{N_{\text{grad}}}d^{*}u(m^{(i)};du(m^{(i)};\psi)).
            \numberthis\label{eq:H_action_continuous_approx}
        \end{align*}
    \end{remark}

    \begin{remark}
        When using the double pass randomized algorithm to obtain the first $r$ eigenpairs in ~\cref{eq:asm_eigen_prob_discrete}, we need to compute the action of $\mathcal{H}$ on $2(r+s)$ random functions in $V^{\text{in}}$ (e.g., their nodal values are sampled from the standard Gaussian distribution), where $s\in\mathbb{N}^{+}$ is typically a small oversampling parameter, often chosen between $5$ and $20$. To speed up the computation, we first compute the LU factorization of the  matrices resulting from the discretization of the linear PDEs in~\cref{eq:p_pde_weak_form} and~\cref{eq:q_pde_weak_form}. Then the action of $d^{*}u(m^{(i)};du(m^{(i)};\cdot))$ on these random functions can be efficiently computed via the forward and backward substitution. Furthermore, the computational time can be significantly reduced by parallelizing the computation of the average value in~\cref{eq:H_action_continuous_approx} across multiple processors.
    \end{remark}

    \subsection{Proof of the equivalence of~\cref{eq:kle_eigen_prob_continuous} 
    and~\cref{eq:kle_eigen_prob_discrete}} 
    \label{sec:proof_kle_eigenproblem_equivalence}
    By the definition of $\mathcal{C}=(\delta I-\gamma \Delta)^{-2}$, \cref{eq:kle_eigen_prob_continuous} is equivalent to 
    \begin{align*}
        \psi = \lambda (\delta I -\gamma \Delta)^2\psi. 
        \numberthis\label{eq:kle_eigen_prob_continuous_equivalent_form}
    \end{align*}
    We first compute $(\delta I -\gamma \Delta)\psi$. To do this, let $p=(\delta I-\gamma \Delta)\psi$ and multiply both sides of this equation by a test function $v$ and integrate
    \begin{align*}
        \int_{\Omega} p(x) v(x)\;\mathrm{d}x 
        &=\int_{\Omega} (\delta I -\gamma \Delta)\psi(x)v(x)\;\mathrm{d}x \\
        &=\delta \int_{\Omega} \psi(x)v(x)\;\mathrm{d}x+\gamma\int_{\Omega}\langle\nabla\psi(x), \nabla v(x)\rangle\;\mathrm{d}x,
        \numberthis\label{eq:kle_p_weak_form}
    \end{align*}
    where, in the second equality,  we use integration by parts and the assumption that the test function vanishes on the boundary. Then we substitute $v$ with all of the finite element basis functions $\phi_1^{\text{in}}, \ldots, \phi_{N_h^{\text{in}}}^{\text{in}}$ and collect the corresponding linear equations for the nodal values of $p$
    \begin{align*}
        A^{\text{in}}\bm{\psi}=M^{\text{in}}\bm{p}, 
        \numberthis\label{eq:kle_p_linear_system}
    \end{align*}
    where the $(i,j)$-entries of $A^{\text{in}}$ and $M^{\text{in}}$ are given by
    \begin{align*}
        A_{ij}^{\text{in}}
        =\delta \langle \phi_{j}^{\text{in}}, \phi_i^{\text{in}} \rangle 
        + \gamma \langle \nabla \phi_j^{\text{in}}, \nabla \phi_i^{\text{in}} \rangle, 
        \quad M_{ij}^{\text{in}}=\langle \phi_j^{\text{in}}, \phi_i^{\text{in}}\rangle.
    \end{align*}
    By~\cref{eq:kle_eigen_prob_continuous_equivalent_form}, function $p$ satisfies $\psi=\lambda (\delta I-\gamma \Delta)p$. In the same manner we can see that
    \begin{align*}
        \lambda A^{\text{in}}\bm{p}=M^{\text{in}}\bm{\psi}. 
        \numberthis\label{eq:kle_p_linear_system_2}
    \end{align*}
    Note that both matrices $M^{\text{in}}$ and $ A^{\text{in}}$ are symmetric positive definite and thus nonsingular. Combining~\cref{eq:kle_p_linear_system} with~\cref{eq:kle_p_linear_system_2} yields
    \begin{align*}
       \lambda A^{\text{in}}(M^{\text{in}})^{-1}A^{\text{in}}\bm{\psi}=M^{\text{in}}\bm{\psi}, 
       \numberthis\label{eq:kle_psi_linear_system}
    \end{align*}
    or equivalently,
    \begin{align*}
        M^{\text{in}}(A^{\text{in}})^{-1}M^{\text{in}}(A^{\text{in}})^{-1}M^{\text{in}}\bm{\psi}=\lambda M^{\text{in}}\bm{\psi}.  
        \numberthis\label{eq:kle_psi_linear_system_2}
    \end{align*}

    \subsection{Proof of the equivalence of~\cref{eq:asm_eigen_prob_continuous} and~\cref{eq:asm_eigen_prob_discrete}}
    \label{sec:proof_asm_eigenproblem_equivalence}
    By~\cref{eq:asm_eigen_prob_continuous}, for any test function $v\in V_h^{\text{in}}$, it holds that
    \begin{align*}
        \langle \mathcal{H}\psi, v\rangle = \langle \lambda\mathcal{C}^{-1}\psi, v\rangle.
        \numberthis\label{eq:asm_eigen_prob_weak_form}
    \end{align*}
    In particular, for any $m\sim \nu(m)$, as we let $v$ go through all of the finite element basis functions $\phi_i^{\text{in}}\in V_h^{\text{in}}$, we can show that 
    \begin{align*}
        \begin{pmatrix}
         \langle d^{*}u(m;du(m;\psi)), \phi_1^{\text{in}}\rangle \\
         \vdots\\ 
         \langle d^{*}u(m;du(m;\psi)), \phi_{N_h^{\text{in}}}^{\text{in}}\rangle\\
       \end{pmatrix}
       =(\nabla_{\bm{m}}\bm{u})^{T}M^{\text{out}}(\nabla_{\bm{m}}\bm{u})\bm{\psi}.
       \numberthis\label{eq:H_action_weak_form}
    \end{align*}
    Indeed, by the definition of Gateaux derivative, we have
    \begin{align}
    \label{eq:gateaux_derivative_explicit_form}
        \begin{split}
            du(m;\psi)
            &=\lim_{\varepsilon\to 0}\frac{u(m+\varepsilon\psi)-u(m)}{\varepsilon}\\
            &=\lim_{\varepsilon\to 0}\sum_{i=1}^{N_h^{\text{out}}}\frac{\bm{u}_i(m+\varepsilon\psi)-\bm{u}_i(m)}{\varepsilon}\phi_i^{\text{out}}(x)\\
            &=\lim_{\varepsilon\to 0}\sum_{i=1}^{N_h^{\text{out}}}\frac{\bm{u}_i(\bm{m}_1+\varepsilon\bm{\psi}_1,\cdots, \bm{m}_{N_h^{\text{in}}}+\varepsilon\bm{\psi}_{N_{h}^{\text{in}}})-\bm{u}_i(\bm{m}_1,\cdots,\bm{m}_{N_h^{\text{in}}})}{\varepsilon}\phi_i^{\text{out}}(x)\\
            &=\sum_{i=1}^{N_h^{\text{out}}}(\frac{\partial \bm{u}_i}{\partial \bm{m}_1}\bm{\psi}_1+\cdots+\frac{\partial \bm{u}_i}{\partial \bm{m}_{N_h^{\text{in}}}}\bm{\psi}_{N_h^{\text{in}}})\phi_i^{\text{out}}(x)\\
            &=\phi^{\text{out}}(x)(\nabla_{\bm{m}}\bm{u})\bm{\psi},
        \end{split}
    \end{align}
    where $\phi^{\text{out}}(x)=(\phi_1^{\text{out}}(x), \ldots, \phi_{N_h^{\text{out}}}(x))$ are the finite element basis functions of output function space $V_h^{\text{out}}$. 
    
    Then for any test function $v\in V_h^{\text{in}}$, it holds that
    \begin{align*}
        \langle d^{*}u(m;p), v\rangle
        & = \langle p, du(m; v)\rangle \quad \text{(by the definition of adjoint operator)}\\
        & = \langle 
                \phi^{\text{out}}(x)(\nabla_{\bm{m}}\bm{u})\bm{\psi}, 
                \phi^{\text{out}}(x)(\nabla_{\bm{m}}\bm{u})\bm{v} 
            \rangle \quad \text{(by~\cref{eq:gateaux_derivative_explicit_form})}\\
        &=\bm{v}^{T}(\nabla_{\bm{m}}\bm{u})^{T}M^{\text{out}}(\nabla_{\bm{m}}\bm{u})\bm{\psi}, 
    \end{align*}
    where $M^{\text{out}}$ is the mass matrix of output function space $V_h^{\text{out}}$, i.e., $M^{\text{out}}_{ij}=\langle \phi_j^{\text{out}}, \phi_i^{\text{out}}\rangle$ for $1\leq i,j\leq N_h^{\text{out}}$. Note that if we replace $v$ by the $i$-th finite element basis functions $\phi_i^{\text{in}}$, then $\bm{v}$ becomes the standard unit vector $e_i\in\mathbb{R}^{N_h^{\text{in}}}$ (with the $k$-th entry one and all others zero). Thus, 
    \begin{align*}
        \langle d^{*}u(m;p), \phi_i^{\text{in}}\rangle = e_i^{T}(\nabla_{\bm{m}}\bm{u})^{T}M^{\text{out}}(\nabla_{\bm{m}}\bm{u})\bm{\psi}, \quad 1\leq i\leq N_h^{\text{in}}.
    \end{align*}
    Concatenating all the above equations yields 
    \begin{align*}
       \begin{pmatrix}
         \langle d^{*}u(m;p), \phi_1^{\text{in}}\rangle \\
         \vdots\\ 
         \langle d^{*}u(m;p), \phi_{N_h^{\text{in}}}^{\text{in}}\rangle\\
       \end{pmatrix}
       =(\nabla_{\bm{m}}\bm{u})^{T}M^{\text{out}}(\nabla_{\bm{m}}\bm{u})\bm{\psi}.
    \end{align*}
    Next, we prove that 
    \begin{align*}
         \begin{pmatrix}
         \langle \lambda\mathcal{C}^{-1}\psi, \phi_1^{\text{in}}\rangle \\
         \vdots\\ 
         \langle \lambda\mathcal{C}^{-1}\psi, \phi_{N_h^{\text{in}}}^{\text{in}}\rangle\\
       \end{pmatrix}
       =\lambda A^{\text{in}}(M^{\text{in}})^{-1}A^{\text{in}}\bm{\psi}.
       \numberthis\label{eq:C_inverse_action_weak_form}
    \end{align*}
    Indeed, if we let $w=\lambda\mathcal{C}^{-1}\psi$, then similar to the argument in~\cref{sec:proof_kle_eigenproblem_equivalence}, we have 
    \begin{align*}
        \lambda A^{\text{in}}(M^{\text{in}})^{-1}A^{\text{in}}\bm{\psi}=M^{\text{in}}\bm{w}, 
    \end{align*}
    Note that 
    \begin{align*}
        \langle w, \phi_i^{\text{in}}\rangle = e_i^{T}M^{\text{in}}\bm{w}, \quad 1\leq i\leq N_h^{\text{in}}.
    \end{align*}
    Thus, 
    \begin{align*}
        \begin{pmatrix}
        \langle w, \phi_1^{\text{in}}\rangle \\
         \vdots\\ 
         \langle w, \phi_{N_h^{\text{in}}}^{\text{in}}\rangle\\
        \end{pmatrix}
        =IM^{\text{in}}\bm{w}=\lambda A^{\text{in}}(M^{\text{in}})^{-1}A^{\text{in}}\bm{\psi}.
    \end{align*}
    Combining~\eqref{eq:asm_eigen_prob_weak_form}, ~\eqref{eq:H_action_weak_form} and~\eqref{eq:C_inverse_action_weak_form} yields
    \begin{align*}
        \mathbb{E}_{\bm{m}\sim \nu(\bm{m})}[
       (\nabla_{\bm{m}}\bm{u})^{T}M^{\text{out}}(\nabla_{\bm{m}}\bm{u})\bm{\psi}
       ] = 
       \lambda A^{\text{in}}(M^{\text{in}})^{-1}A^{\text{in}}\bm{\psi}.
    \end{align*}
\section{Experimental details}
\label{sec:experimental_details}

    \subsection{Governing equations}
    \label{sec:governing_equations}
    
\begin{figure}[H]
        \begin{center}
        \centerline{\includegraphics[width=\columnwidth]{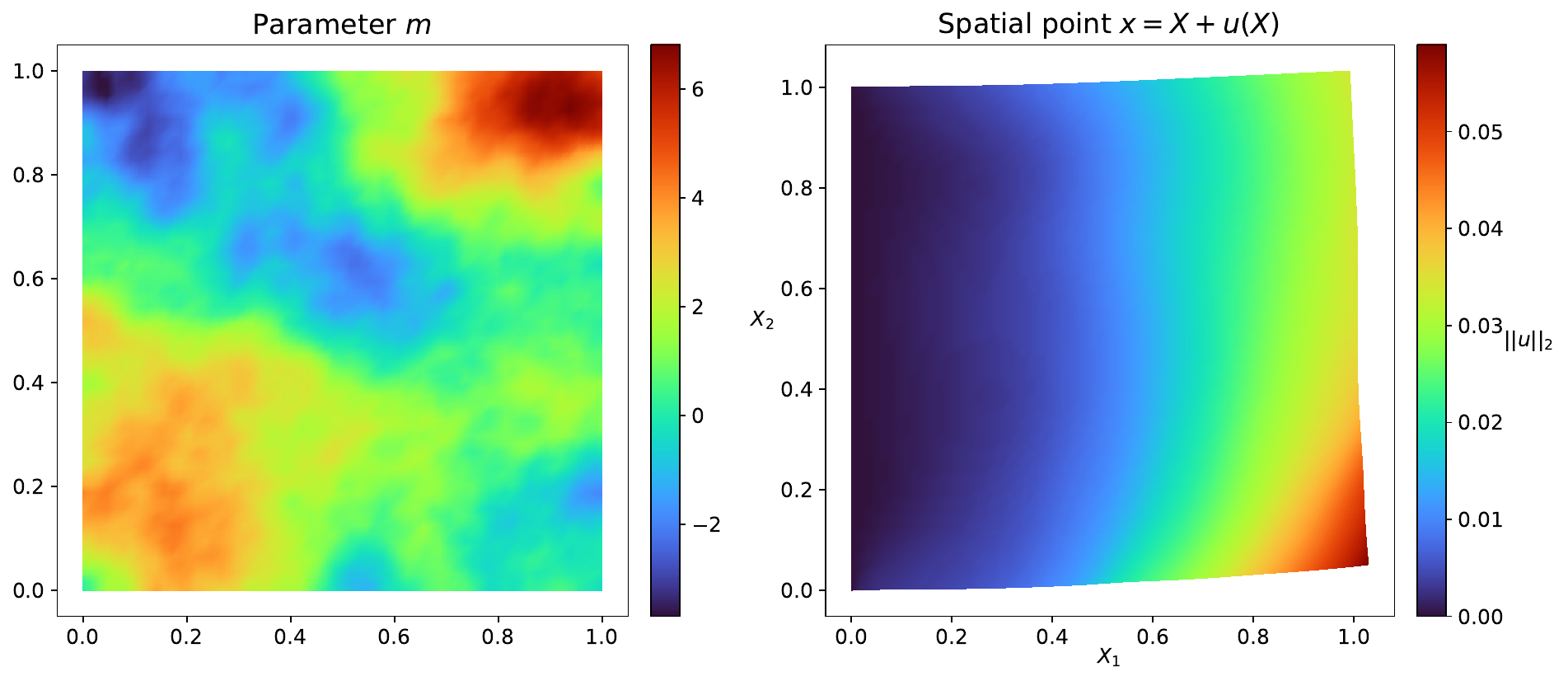}}
        \caption{Visualization of one parameter-solution pair of hyperelasticity equation. The color of the output indicates the magnitude of the displacement $u$ (which maps from domain $\Omega$ to $\mathbb{R}^2$) instead of its componentwise function $u_1$ or $u_2$. The skewed square shows locations of any domain point after deformation $X\to x$. See~\cref{fig:truth_prediction_hyperelasticity_u1,fig:truth_prediction_hyperelasticity_u2} for $u_1$ and $u_2$.}
    \label{fig:input_output_hyperelasticity}
    \end{center}
    \end{figure}

    \paragraph{Hyperelasticity equation.}
    We follow the problem setup in~\cite{cao2023residual}. Write $X$ (instead of $x$) for a material point in the domain $\Omega$ and $u=u(X):\mathbb{R}^2\to \mathbb{R}^2$ for the displacement of the material point. Under the influence of internal and/or external forces, the material point is mapped to a spatial point $x=x(X)=X+u(X):\mathbb{R}^2\to \mathbb{R}^2$. Let $F=\nabla_X x=I+\nabla_X u: \mathbb{R}^2\to\mathbb{R}^{2\times2}$ denote the deformation gradient. For a hyperelasticity material, the internal forces can be derived from a strain energy density 
    \begin{align*}
        W(X,C)&=\frac{\mu(X)}{2}(\mathrm{tr}(C)-3)+\frac{\lambda(X)}{2}(\ln (J))^2-\mu(X)\ln(J). 
    \end{align*}
    Here, $C=F^{T}F$ is the right Cauchy-Green strain tensor, $\mathrm{tr}(C)$ is the trace of matrix~$C$, 
    $J$ is the determinant of matrix~$F$, and $\mu(X), \lambda(X):\mathbb{R}^2\to \mathbb{R}$ are the Lamé parameters which we assume to be related to Young's modulus of elasticity $E(X):\mathbb{R}^2\to \mathbb{R}$ and Poisson ratio $\nu\in\mathbb{R}$ 
    \begin{align*}
        \mu(X)=\frac{E(X)}{2(1+\nu)}, \quad \lambda(X)=\frac{\nu E(X)}{(1+\nu)(1-2\nu)}.
    \end{align*}
    We assume the randomness comes from the Young's modulus $E(X)=e^{m(X)}+1$. Let $S=2\frac{\partial W}{\partial C}$ denote the second Piola-Kirchhoff stress tensor. We consider the case where the left boundary of the material is fixed, and the right boundary is subjected to stretching by an external force $t=t(X):\mathbb{R}^2\to\mathbb{R}^2$. The strong form of the steady state PDE can be written as   
    \begin{align*}
        \begin{cases}
            \nabla_X\cdot (FS)=0, &\quad X\in \Omega, \\
            u=0,  &\quad X\in \Gamma_{\text{left}}, \\
            FS\cdot n=0, &\quad X\in \Gamma_{\text{top}}\cup \Gamma_{\text{bottom}}, \\
            FS\cdot n=t, &\quad X\in \Gamma_{\text{right}},
        \end{cases}
    \end{align*}
    where $\Gamma_{\text{left}}, \Gamma_{\text{right}}, \Gamma_{\text{top}}$ and $\Gamma_{\text{bottom}}$ denote the left, right, top, and bottom boundary of the material domain $\Omega$, respectively, and $n$ is the unit outward normal vector on the boundary. Our goal is to learn the operator that maps the parameter $m$ to the displacement $u$. For demonstration, we choose $\bar{m}=0$, $\delta=0.4$, $\gamma=0.04$, Poisson ratio $\nu=0.4$, and the external force
    \begin{align*}
        t(X)=\left(0.06\exp(-0.25|X_2-0.5|^2), 0.03(1+0.1 X_2)\right)^{T}.
    \end{align*}
    In practice, we solve the PDE by first formulating the energy $\widetilde{W}$ in the weak form
    \begin{align*}
        \widetilde{W}=\int_{\Omega} W \mathrm{d}X - \int_{\Gamma_{\text{right}}} \langle t, u\rangle \;\mathrm{d}s
    \end{align*}
    and then solving for $u$ that satisfies the stationary condition, that is, the equation 
    \begin{align*}
        d\widetilde{W}(u;v)=0,
    \end{align*}
    holds for any test function $v$ in the state space. See~\cref{fig:input_output_hyperelasticity} for the visualization for one parameter-solution pair of the hyperelasticity equation.

\begin{figure}[H]
        \begin{center}
        \centerline{\includegraphics[width=\columnwidth]{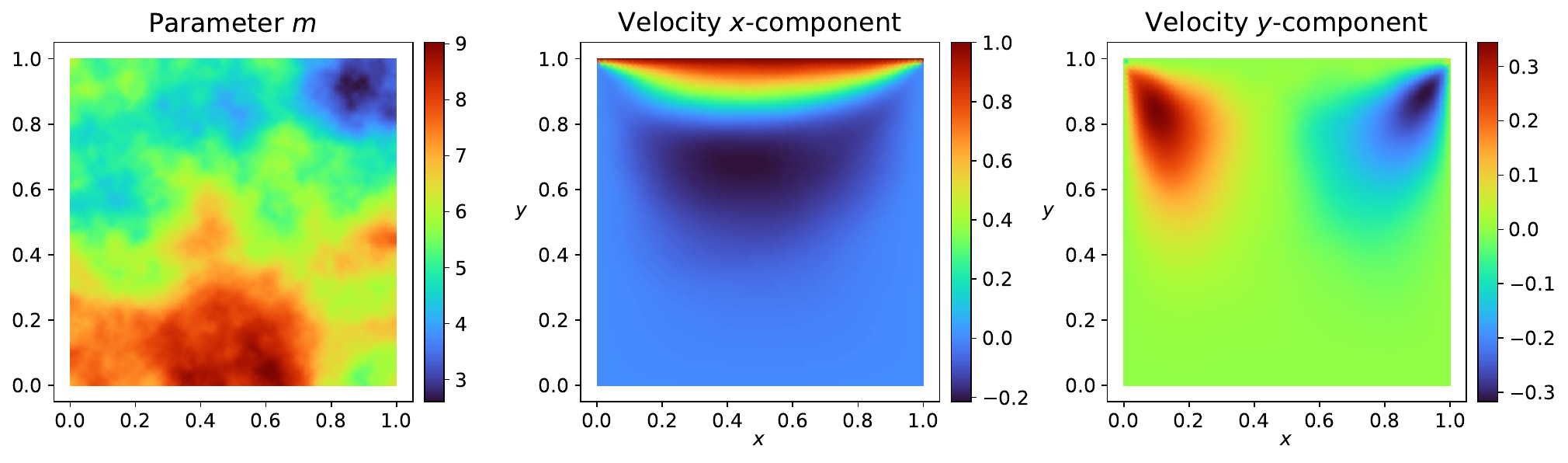}}
        \caption{Visualization of one parameter-solution pair of Navier--Stokes equations.}
    \label{fig:input_output_navier_stokes}
    \end{center}
    \end{figure}

    \paragraph{Navier--Stokes equations. } Let $u=u(x)\in\mathbb{R}^2$ and $p=p(x)\in\mathbb{R}$ denote the velocity and pressure at point $x\in \Omega=(0,1)^2$. The strong form of the PDE can be written as 
    \begin{align*}
        \begin{cases}
            -\nabla\cdot e^{m} \nabla u + (u \cdot \nabla)u+\nabla p = 0,  &\quad x\in\Omega, \\ 
            \nabla \cdot u = 0, &\quad x\in\Omega, \\
            u=(1,0)^{T}, &\quad x\in\Gamma_{\text{top}}, \\
            u=(0,0)^{T}, &\quad x \in\Gamma\setminus\Gamma_{\text{top}},
        \end{cases}
    \end{align*}
    where $\Gamma_{\text{top}}$ and $\Gamma$ denote the left and whole boundary of the cavity domain $\Omega$, respectively. Here, we assume that the randomness arises from the viscosity term $e^{m}$. Our goal is learn the operator that maps the parameter $m$ to the velocity~$u$. For demonstration, we choose $\bar{m}=6.908$ ($e^{\bar{m}}\approx 10^{3}$, thus the viscosity term dominates), $\delta=0.6$, and $\gamma=0.06$. See~\cref{fig:input_output_navier_stokes} for the visualization for one parameter-solution pair of the Navier--Stokes equations.

    \subsection{Data generation}
    \label{sec:data_generation}

\begin{figure}[H]
    \begin{center}
        \centerline{\includegraphics[width=\columnwidth]{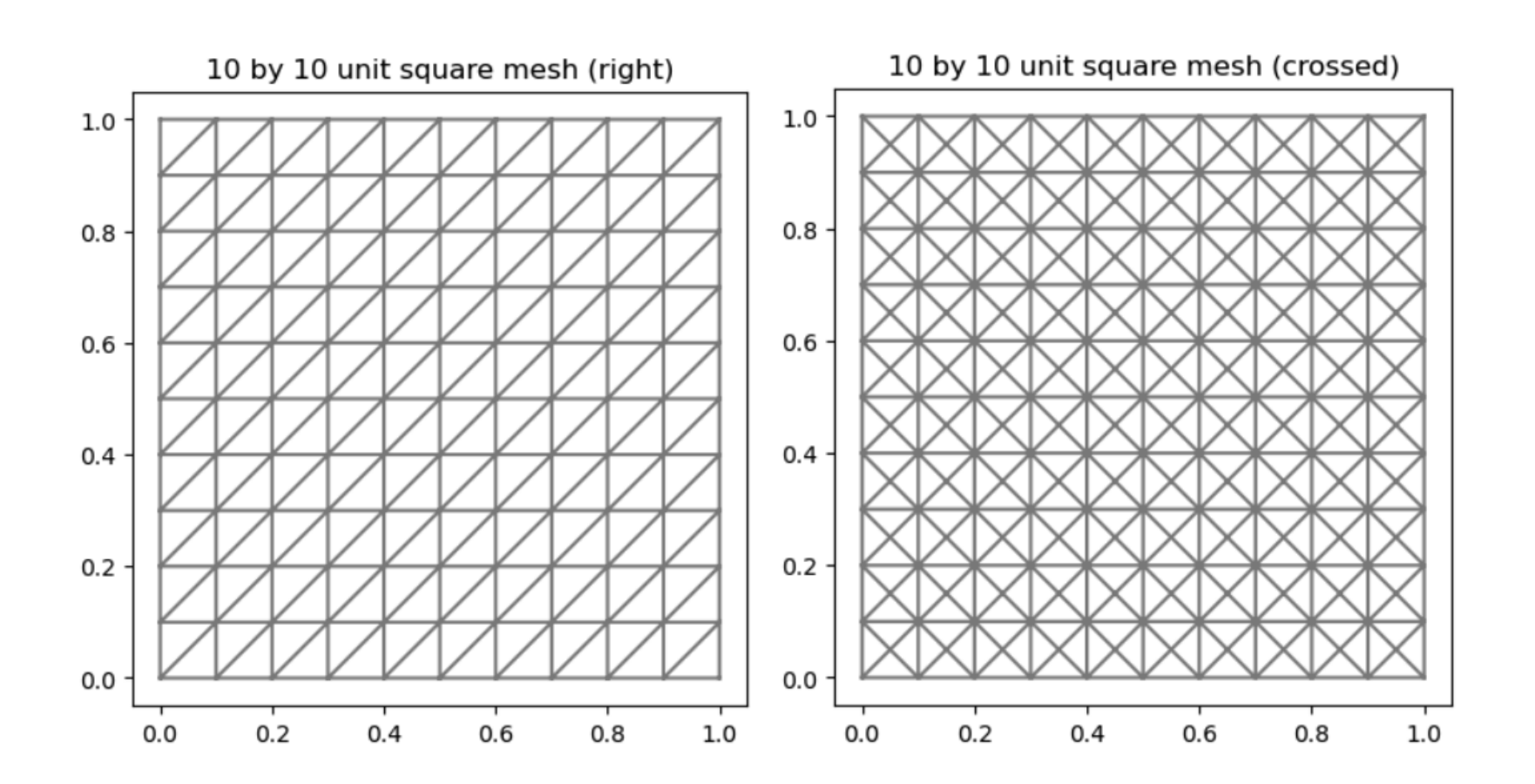}}
        \caption{Visualization of the $10$ by $10$ unit square mesh. Left: diagonal=`right'; Right: diagonal=`crossed'}
    \label{fig:mesh}
    \vskip -2em
    \end{center}
    \end{figure}

    For all PDEs in this work, we use the class \texttt{dolfin.UnitSquareMesh} to create a triangular mesh of the 2D unit square with $64$ cells in horizontal direction and $64$ cells in vertical direction. For the Darcy flow equation and hyperelasticity equation, we set the direction of the diagonals as 'right', while for the Navier--Stokes equation, we set the direction of the diagonals as 'crossed'. See~\cref{fig:mesh} for a visualization of the unit square mesh with a $10$ by $10$ resolution. 

    We use the class \texttt{dolfin.FunctionSpace} or \texttt{dolfin.VectorFunctionSpace} to create the finite element function space of input function and output function. For the finite element basis functions, we consider the Continuous Galerkin (CG) family (or the standard Lagrange family) with degree 1 or 2. See~\cref{tab:data_generation_configuration} for details.   

    \begin{table}[H]
    \caption{Configurations of data generation for different datasets}
    \label{tab:data_generation_configuration}
    \begin{center}
    \begin{tiny}
        \begin{tabular}{@{}cccccccc@{}}
        \toprule
        \multirow{2}{*}{\( \text{Dataset} \)} & \multicolumn{7}{c}{\( \text{Configurations}\)} \\
        \cmidrule(l){2-8}
                           & Mesh                    &   $\phi_i^{\text{in}}$  & $\phi_i^{\text{out}}$                  & $(N_{\text{train}}, N_{\text{test}})$ & $r$  & $N_{\text{grad}}$ & $s$ \\ 
        \midrule
        Hyperelasticity    & $64\times 64$, right    & CG (1D, deg=1)  &   CG (2D, deg=1)                               &   $(1500,500)$                        &   $16$           & $16$              & $10$ \\
        Navier--Stokes      & $64\times 64$, crossed  & CG (1D, deg=1)  &   $u$: CG (2D, deg=2), $p$: CG (1D, deg=1)     &   $(1500,500)$                        &   $16$           & $16$              & $10$ \\
        \bottomrule
        \end{tabular}
    \end{tiny}
    \end{center}
    \vskip -1em
    \end{table}
    
    We generate $N_{\text{train}}=1500$ and $N_{\text{test}}=500$ input-output pairs $(m^{(i)}, u^{(i)})$ for training and testing, respectively. We compute first $r=16$ KLE basis and ASM basis using double pass randomized algorithm with an oversampling parameter $s$ of $10$. In the computation of ASM basis, we use $N_{\text{grad}}=16$ samples for the Monte Carlo estimate of the action of operator $\mathcal{H}$ in~\cref{eq:H_action_continuous_approx}. In our method, we formulate the different labels into arrays with the shape as follows
    \begin{itemize}
        \itemsep-0.3em 
        \item Evaluation labels: ($N_{\text{train}}$, $N_x$, $N_{u}$)
        \item Derivative $m$ labels: ($N_{\text{train}}$, $N_x$, $r$, $N_{u}$)
    \end{itemize}
    Recall that $N_{\text{train}}$ is the number of functions used for training, $N_x$ is the number of nodes of the mesh, $N_{u}$ is the dimension of output function, $r$ is the number of reduced basis, and $d$ is the dimension of the domain.

    \subsection{Computation of derivative labels and outputs}
    \label{subsec:computation_of_derivative_labels_and_outputs}
    \subsubsection{Derivative labels $p:=du(m;\psi)$ as ground truth} Since the PDE residual $\mathcal{R}(m, u(m))\equiv 0$ holds for any $m\in V_h^{\text{in}}$, we have 
    \begin{align*}
        \mathcal{R}(m+\varepsilon\psi, u(m+\varepsilon\psi))=0, \quad \forall \varepsilon>0, \psi\in V_h^{\text{in}}. 
    \end{align*}
    By the Taylor expansion and $\mathcal{R}(m,u(m))=0$, we obtain
    \begin{align}
         (\partial_m \mathcal{R}(m,u(m)))\varepsilon\psi +(\partial_u \mathcal{R}(m,u(m)))\delta u \approx 0,
         \label{eq:dm_computation}
    \end{align}
    where $\delta u=u(m+\varepsilon \psi)-u(m)$. Dividing both sides of Eq.~\eqref{eq:dm_computation} by $\varepsilon$ and letting $\varepsilon$ approach $0$ yields 
    \begin{align}
        (\partial_m \mathcal{R})\psi + (\partial_u \mathcal{R})p=0, 
        \label{eq:dm_computation_2}
    \end{align}
    where $p:=du(m;\psi)$. We solve Eq.~\eqref{eq:dm_computation_2} for $p$ via its weak form 
    \begin{align*}
        \langle (\partial_m \mathcal{R})\psi, v\rangle + \langle (\partial_u \mathcal{R})p, v\rangle=0,
    \end{align*}
    where $v$ is a test function in  $V_h^{\text{out}}$.

    \paragraph{Example.} Consider the (nonlinear) diffusion-reaction equation 
    \begin{align*}
        \begin{cases}
            &-\nabla \cdot (e^{m}\nabla u) + u^3=1, \quad x\in\Omega,\\ 
            &u(x)=0, \quad x\in\partial\Omega.
        \end{cases}
    \end{align*}
    Then 
    \begin{itemize}
        \item $\mathcal{R}=-\nabla\cdot(e^{m}\nabla u)+u^3-1$
        \item $(\partial_{m}\mathcal{R})\psi=-\nabla\cdot(e^{m}\psi\nabla u)$ 
        \item $(\partial_{u}\mathcal{R})p=-\nabla\cdot (e^{m}\nabla p)+3u^2p$
    \end{itemize}
    Thus, $p$ satisfies the linear PDE
    \begin{align*}
        \langle e^{m}\psi\nabla u, \nabla v\rangle + 
        \langle e^{m}\nabla p,\nabla v\rangle + 
        \langle 3u^2p, v\rangle=0. 
    \end{align*}
    Using FEniCS, we can easily derive Gâteaux derivative via automatic symbolic differentiation instead of by hand. In this case, the Python code for representing the weak form of the residual $\langle \mathcal{R}, v\rangle$ and Gâteaux derivatives $\langle (\partial_{m}\mathcal{R})\psi, v\rangle$ and $\langle (\partial_{u}\mathcal{R})p, v\rangle$ can be written as 
    
    \qquad\texttt{import dolfin as dl}
    \begin{itemize}
        \item \texttt{R=(dl.inner(dl.exp(m)*dl.grad(u), dl.grad(v))*dl.dx}\\
        \texttt{+(u**3-dl.Constant(1.0))*v*dl.dx)}
        \item \texttt{dl.derivative(R,m,psi)}
        \item \texttt{dl.derivative(R,u,p)}
    \end{itemize}

    \subsubsection{Derivative outputs of neural networks}
    \label{subsubsec:derivative_outputs_of_neural_networks}
    
    \paragraph{DE-DeepONet.} 
    For notation simplicity, we illustrate the case where the input reduced basis is ASM basis and the output function is real-valued. The output of the model is given by 
    \begin{align*}
        \hat{u}(\bm{m};\theta)(x)=\langle b((\Psi^{\text{in}})^{T}C^{-1}\bm{m}; \theta_b), t(x;\theta_t)\rangle+\theta_{\text{bias}},
    \end{align*}
    where $\Psi^{\text{in}}=(\bm{\psi}_1^{\text{in}}|\cdots|\bm{\psi}_{r_{\text{in}}}^{\text{in}})\in \mathbb{R}^{N_{h}^{\text{in}}\times r_{\text{in}}}$ are the nodal values of input (ASM) reduced basis functions,~$C^{-1}$ is the inverse of the covariance matrix of the Gaussian random field $\nu$ where parameter $m$ is sampled from (Recall that the ASM basis are orthonormal in the inner product with weight matrix ${C}^{-1}$.) Thus, by the chain rule of derivative, for any test direction $\bm{\psi}_{\text{test}}$, one has 
    \begin{align*}
        \nabla_{\bm{m}}\hat{u}(\bm{m};\theta)(x)\bm{\psi}_{\text{test}}=\nabla_{\widetilde{m}}\langle b(\widetilde{m}), t(x;\theta_t)\rangle(\Psi^{\text{in}})^{T}C^{-1}\bm{\psi}_{\text{test}}, 
    \end{align*}
     where $\widetilde{m}=(\Psi^{\text{in}})^{T}C^{-1}\bm{{m}}$. The Jacobian matrix $\nabla_{\widetilde{m}}\langle b(\widetilde{m}), t(x;\theta_t)\rangle$ can be efficiently computed using, e.g., \texttt{torch.func.jacrev}, and further parallelized with \texttt{torch.vmap}. If $\bm{\psi}_{\text{test}}$ is in $\Psi^{\text{in}}$ during model training, we can see that $(\Psi^{\text{in}})^{T}C^{-1}\bm{\psi}_{\text{test}}$ becomes a unit vector which frees us the need to compute it, otherwise in the model inference stage when $\bm{\psi}_{\text{test}}$ is (the nodal values of) a random function sampled from $\nu$, we compute $T=(\Psi^{\text{in}})^{T}C^{-1}$ and then $T\bm{\psi}_{\text{test}}$. 
     
    \paragraph{DINO.} The output of the model is given by 
    \begin{align*}
        \hat{u}(\bm{m};\theta)(x)=\Phi^{\text{out}}(x)\Psi^{\text{out}}f_{\theta}((\Psi^{\text{in}})^{T}C^{-1}\bm{{m}}),
    \end{align*}
    where $\Phi^{\text{out}}(x)=(\phi_1^{\text{out}}(x), \ldots, \phi_{N_h^{\text{out}}}^{\text{out}}(x))\in \mathbb{R}^{1\times N_{h}^{\text{out}}}$ denotes the output finite element basis functions evaluated on point $x\in \Omega$, $\Psi^{\text{out}}=(\bm{\psi}_1^{\text{out}}|\cdots|\bm{\psi}_{r_{\text{out}}}^{\text{out}})\in \mathbb{R}^{N_{h}^{\text{out}}\times r_{\text{out}}}$ are the nodal values of output (POD) reduced basis functions, similarly for $\Psi^{\text{in}}$ denoting the input (ASM) reduced basis, and $C^{-1}$ is the inverse of the covariance matrix of the Gaussian random field where parameter $m$ is sampled from. Thus, by the chain rule of derivative, for any test direction $\bm{\psi}_{\text{test}}$, one has 
    \begin{align*}
        \nabla_{\bm{m}}\hat{u}(\bm{m};\theta)(x)\bm{\psi}_{\text{test}}=\Phi^{\text{out}}(x)\Psi^{\text{out}}\nabla_{\widetilde{m}} f_{\theta}(\widetilde{m})(\Psi^{\text{in}})^{T}C^{-1}\bm{\psi}_{\text{test}},
    \end{align*}
    where $\widetilde{m}=(\Psi^{\text{in}})^{T}C^{-1}\bm{{m}}$. For fast evaluations given different $x$, $\bm{m}$, and $\bm{\psi}_{\text{test}}$, we first compute $T_{\text{left}} = \Phi^{\text{out}}(x) \Psi^{\text{out}}$ for all points $x$ that need to be evaluated and $T_{\text{right}} = (\Psi^{\text{in}})^{T}C^{-1}$. Next, we compute $J = \nabla_{\widetilde{m}} f_{\theta}(\widetilde{m})$ using, e.g., \texttt{torch.func.jacrev}, and finally compute $T_{\text{left}} J T_{\text{right}} \bm{\psi}_{\text{test}}$.

    \paragraph{DeepONet \& FNO.} Both models receive the full high dimensional parameter $\bm{m}$, so we compute the directional derivative $\nabla_{\bm{m}}\hat{u}(\bm{m};\theta)\bm{\psi}_{\text{test}}$ using the Jacobian-vector product \texttt{torch.jvp} instead of the full Jacobian in DE-DeepONet and DINO. For the coordinates $x$ as additional inputs, we pad zeros to the tensor $\bm{\psi}_{\text{test}}$ to match the dimension of the input tensor $(\bm{m}, \{x^{(j)}\}_{j=1}^{N_x})$.

    \subsection{Training details}
    \label{sec:training_details}
    For the DeepONet, we parameterize the branch net using a CNN and the trunk net using a ResNet. For the FNO, we use the package \href
    {https://github.com/neuraloperator/neuraloperator}{neuraloperator}. For the DINO, we use $16$ ASM basis functions for the input dimension reduction and $64$ POD basis functions for the output dimension reduction, and parameterize the neural network using a ResNet. For the DE-DeepONet, both the branch net and trunk net are parameterized using ResNets. We train each model for 32768 iterations (with the same batch size 8) using an \texttt{AdamW} optimizer~\cite{loshchilov2017decoupled} and a \texttt{StepLR} learning rate scheduler (We disable learning rate scheduler for DE-DeepONet). See~\cref{tab:training_details_deeponet,tab:training_details_fno,tab:training_details_dino,tab:training_details_de_deeponet} for details. When the loss function comprises two terms, we apply the self-adaptive learning rate annealing algorithm from~\cite{wang2023expert}, with an update frequency of $100$ and a moving average parameter of $0.9$, to automatically adjust the loss weights $\{\lambda_i\}_{i=1}^{2}$ in~\cref{eq:loss_function_discrete}. Additionally, we standardize the inputs and labels before training.

    \begin{table}[H]
        \caption{Training details for DeepONet}
        \label{tab:training_details_deeponet}
        \begin{center}
        \begin{small}
            \begin{tabular}{@{}c p{5cm} p{5cm} @{}}
            \toprule
            \multirow{2}{*}{\( \text{} \)} & \multicolumn{2}{c}{\( \text{Dataset} \)} \\
            \cmidrule(l){2-3}
                            & Hyperelasticity & Navier--Stokes\\
            \midrule    
            branch net      &  CNN \newline
                             \null\qquad 6 hidden layers \newline 
                             \null\qquad 256 output dim \newline 
                             \null\qquad ReLU 
                            &  CNN \newline 
                             \null\qquad 6 hidden layers \newline 
                             \null\qquad 256 output dim \newline 
                             \null\qquad ReLU \\
            trunk net       &  ResNet \newline 
                            \null\qquad 3 hidden layers \newline 
                            \null\qquad 512 output dim \newline 
                            \null\qquad 512 hidden dim \newline
                            \null\qquad ReLU
                            & ResNet \newline 
                            \null\qquad 3 hidden layers \newline 
                            \null\qquad 512 output dim \newline 
                            \null\qquad 512 hidden dim \newline
                            \null\qquad ReLU \\
            initialization  & Kaiming Uniform                         & Kaiming Uniform  \\
            AdamW (lr, weight decay)  & $(10^{-3},10^{-6})$                       & $(10^{-3},10^{-5})$  \\
            StepLR (gamma, step size)  & $(0.6,10)$                              & $(0.7,10)$  \\
            number of Fourier features  & $64$                            & $64$ \\
            Fourier feature scale $\sigma$  &  $0.5$                       & $1.0$ \\
            \bottomrule
            \end{tabular}
        \end{small}        
        \end{center}
    \vskip -2em
    \end{table}

    \begin{table}[H]
    \caption{Training details for FNO \& DE-FNO}
    \label{tab:training_details_fno}
    \begin{center}
    \begin{small}
        \begin{tabular}{@{}ccc@{}}
        \toprule
        \multirow{2}{*}{\( \text{} \)} & \multicolumn{2}{c}{\( \text{Dataset} \)} \\
        \cmidrule(l){2-3}
                        & Hyperelasticity & Navier--Stokes\\
        \midrule    
        number of modes           & $$(32,32)$$       & $(32,32)$ \\
        in channels      & $3$              & $3$      \\
        out channels     & $2$              & $2$      \\
        hidden channels  & $64$             & $64$     \\
        number of layers        & $4$              & $4$      \\
        lifting channel ratio   & $2$              & $2$    \\
        projection channel ratio & $2$             & $2$    \\
        activation function      & GELU            & GELU    \\
        AdamW (lr, weight decay)  & $(5\times 10^{-3},10^{-4})$                       & $(5\times 10^{-3},10^{-4})$  \\
        StepLR (gamma, step size)  & $(0.9,25)$                              & $(0.9,50)$  \\
        \bottomrule
        \end{tabular}
    \end{small}        
    \end{center}
    \vskip -2em
    \end{table}

    \begin{table}[H]
        \caption{Training details for DINO}
        \label{tab:training_details_dino}
        \begin{center}
        \begin{small}
            \begin{tabular}{@{}c p{5cm} p{5cm} @{}}
            \toprule
            \multirow{2}{*}{\( \text{} \)} & \multicolumn{2}{c}{\( \text{Dataset} \)} \\
            \cmidrule(l){2-3}
                            & Hyperelasticity & Navier--Stokes\\
            \midrule    
            neural network     &  ResNet \newline
                             \null\qquad 3 hidden layers \newline 
                             \null\qquad 64 output dim \newline 
                             \null\qquad 128 hidden dim \newline
                             \null\qquad ELU
                               &  ResNet \newline 
                             \null\qquad 3 hidden layers \newline 
                             \null\qquad 64 output dim \newline
                             \null\qquad 256 hidden dim \newline 
                             \null\qquad ELU \\
            initialization  & Kaiming Normal                               & Kaiming Normal  \\
            AdamW (lr, weight decay)  & $(5\times 10^{-3},0.0)$                  & $(5\times 10^{-3},10^{-12})$  \\
            StepLR (gamma, step size)  & $(0.5,50)$                              & $(0.5,50)$  \\
            \bottomrule
            \end{tabular}
        \end{small}        
        \end{center}
    \vskip -2em
    \end{table}

    \begin{table}[H]
        \caption{Training details for DE-DeepONet}
        \label{tab:training_details_de_deeponet}
        \begin{center}
        \begin{small}
            \begin{tabular}{@{}c p{5cm} p{5cm} @{}}
            \toprule
            \multirow{2}{*}{\( \text{} \)} & \multicolumn{2}{c}{\( \text{Dataset} \)} \\
            \cmidrule(l){2-3}
                            & Hyperelasticity & Navier--Stokes\\
            \midrule    
            branch net      &  ResNet \newline
                             \null\qquad 3 hidden layers \newline 
                             \null\qquad 128 output dim \newline 
                             \null\qquad 128 hidden dim \newline
                             \null\qquad ELU 
                            &  ResNet \newline 
                             \null\qquad 3 hidden layers \newline 
                             \null\qquad 256 output dim \newline
                             \null\qquad 256 hidden dim \newline 
                             \null\qquad ELU \\
            trunk net       &  ResNet \newline 
                            \null\qquad 3 hidden layers \newline 
                            \null\qquad 256 output dim \newline 
                            \null\qquad 256 hidden dim \newline
                            \null\qquad ReLU
                            & ResNet \newline 
                            \null\qquad 3 hidden layers \newline 
                            \null\qquad 512 output dim \newline 
                            \null\qquad 512 hidden dim \newline
                            \null\qquad ReLU \\
            initialization  & Kaiming Uniform                         & Kaiming Uniform  \\
            AdamW (lr, weight decay)  & $(10^{-3},10^{-11})$                       & $(10^{-3},10^{-11})$  \\
            disable lr scheduler &   True                                   & True  \\ 
            number of Fourier features  & $64$                            & $64$ \\
            Fourier feature scale $\sigma$  &  $0.5$                       & $1.0$ \\
        $N_x^{\text{batch}}(=\alpha N_x)$ & $422\approx(0.1\times 65^2)$  & $422\approx (0.1\times 65^2)$  \\
            \bottomrule
            \end{tabular}
        \end{small}        
        \end{center}
    \vskip -2em
    \end{table}

   \begin{table}[H]
    \caption{Number of trainable parameters in each model}
    \label{tab:number_of_neural_network_parameters}
    \begin{center}
    \begin{small}
        \begin{tabular}{@{}ccccc@{}}
        \toprule
        \multirow{2}{*}{\( \text{Dataset} \)} & \multicolumn{4}{c}{\( \text{\# parameters}\)} \\
        \cmidrule(l){2-5}
                           & DeepONet       &  FNO  \& DE-FNO         &  DINO     & DE-DeepONet             \\ 
        \midrule
        Hyperelasticity    & 4.21 M    &  17.88 M     &  0.04 M & 0.27 M            \\
        Navier--Stokes     & 4.21 M    &  17.88 M     &  0.15 M & 1.02 M            \\
        \bottomrule
        \end{tabular}
    \end{small}        
    \end{center}
    \vskip -2em
    \end{table}

    \subsection{Convergence plot}
    \label{sec:convergence_plot}
    
\begin{figure}[H]
        \begin{center}
        \centerline{\includegraphics[width=0.33\columnwidth]{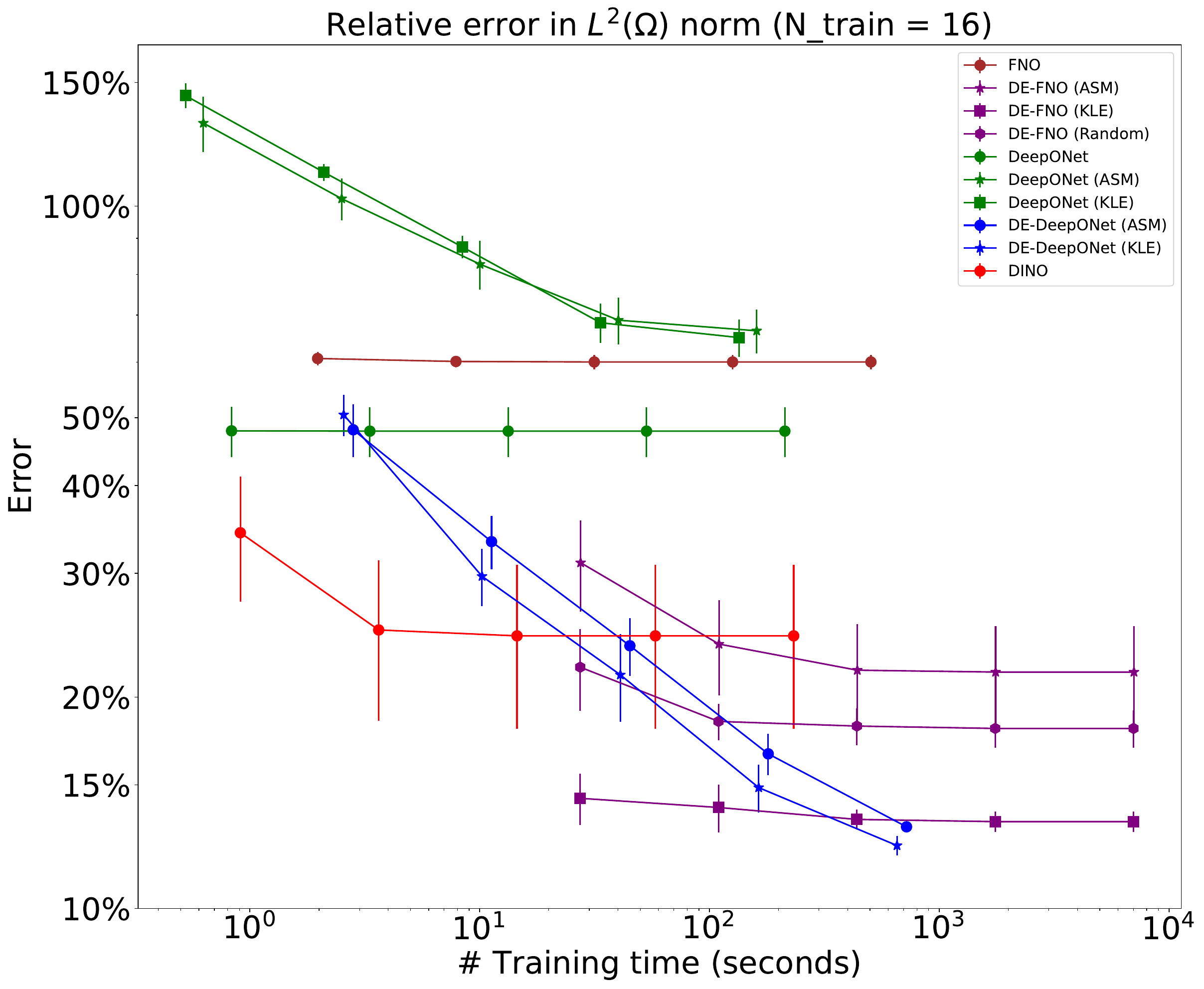}
                    \includegraphics[width=0.33\columnwidth]{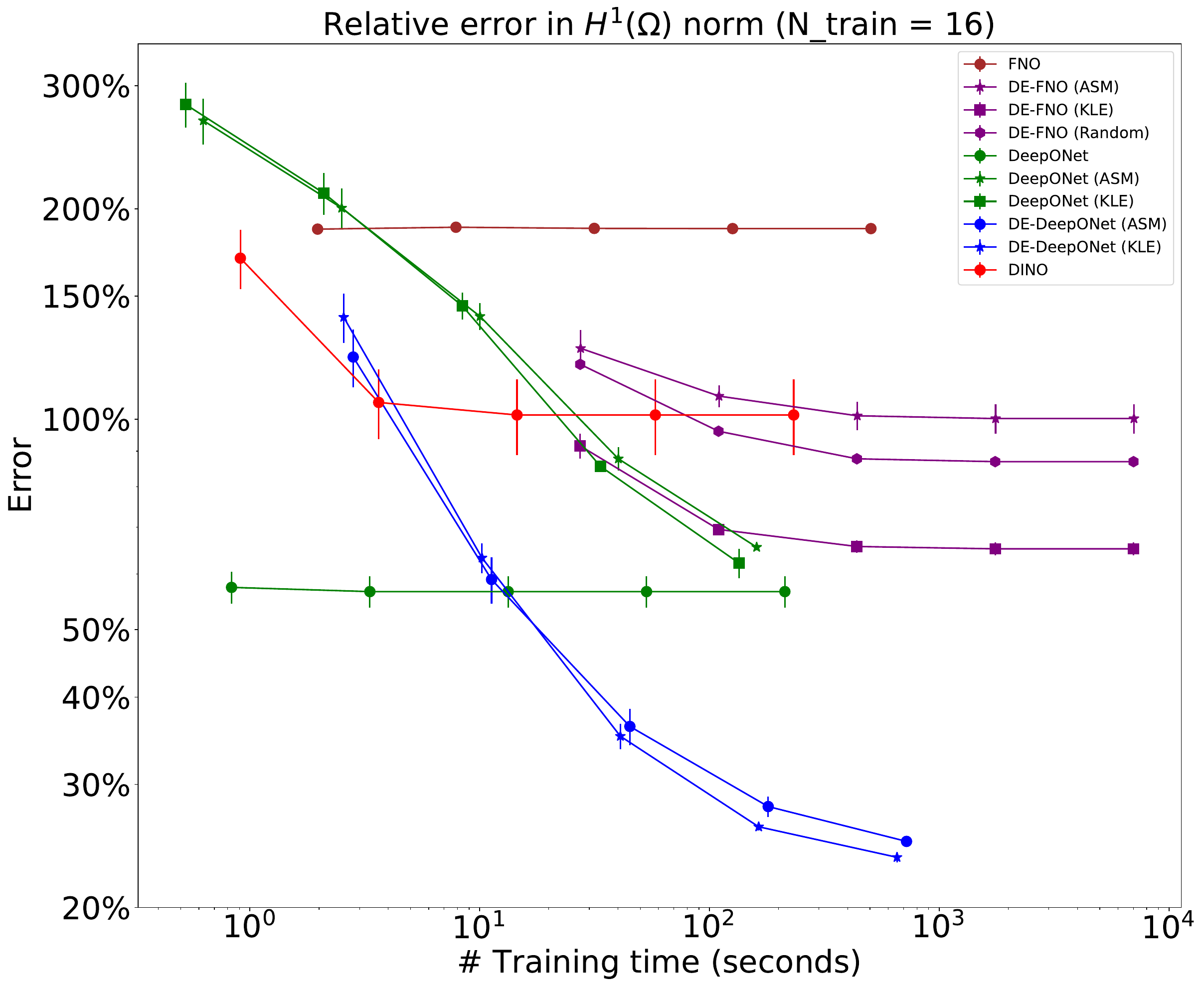}
                    \includegraphics[width=0.33\columnwidth]{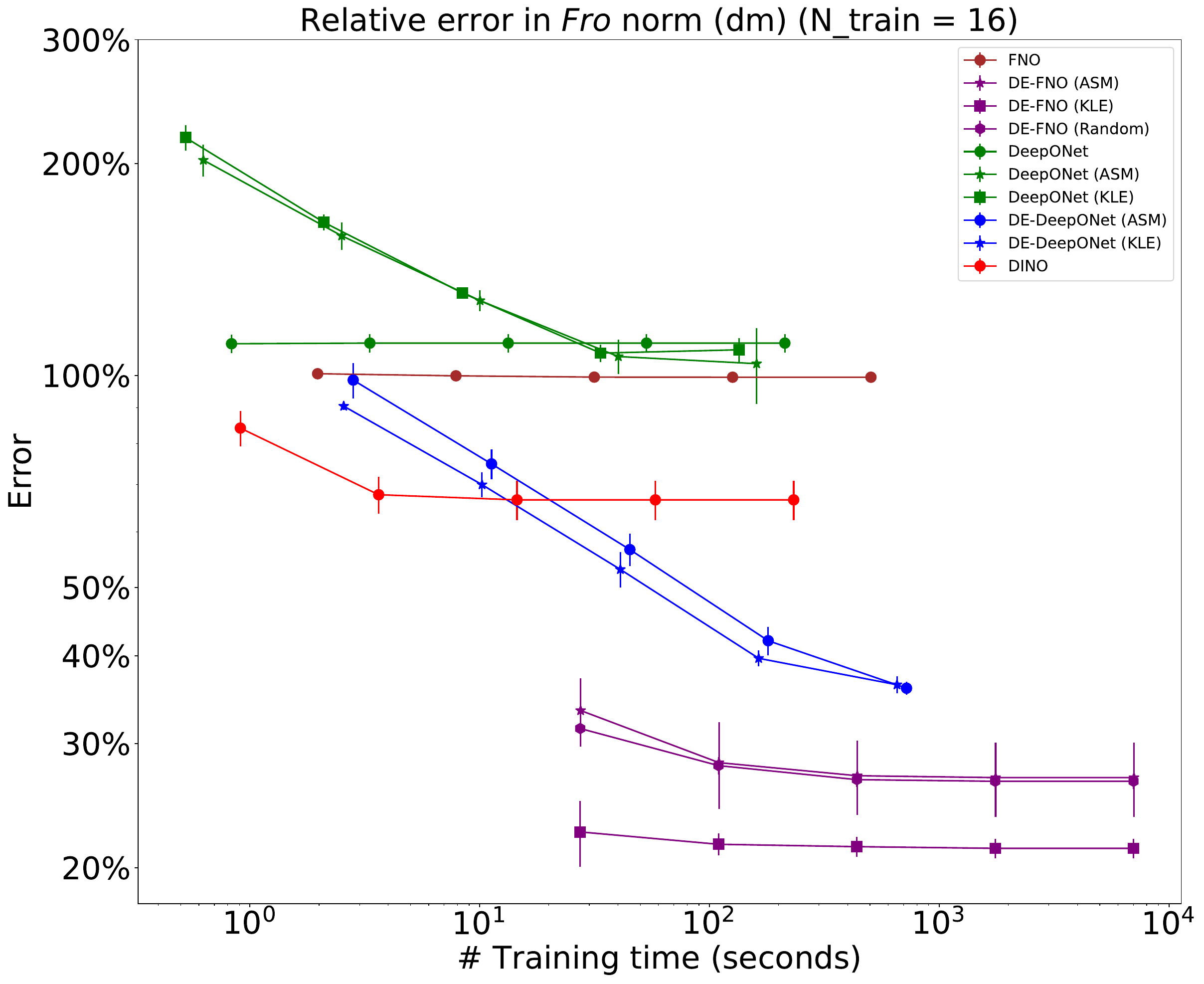}
        }
        \caption{Mean relative errors ($\pm$ standard deviation) over 5 trials versus model training time for the hyperelasticity equations when the number of training samples is 16.
        }
    \label{fig:error_plot_hyperelasticity_convergence_16}
    \end{center}
    \vskip -2em
    \end{figure}

\begin{figure}[H]
        \begin{center}
        \centerline{\includegraphics[width=0.33\columnwidth]{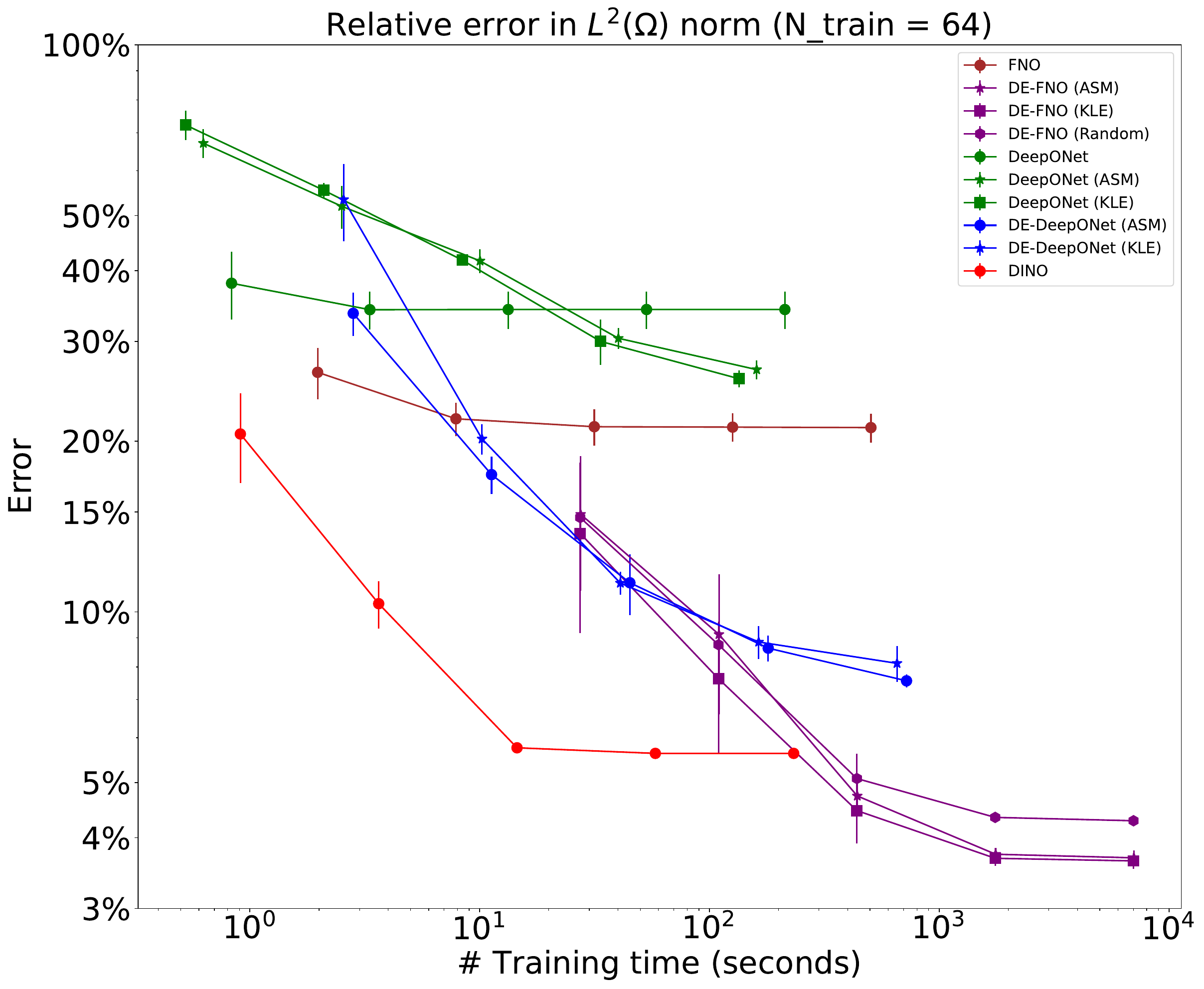}
                    \includegraphics[width=0.33\columnwidth]{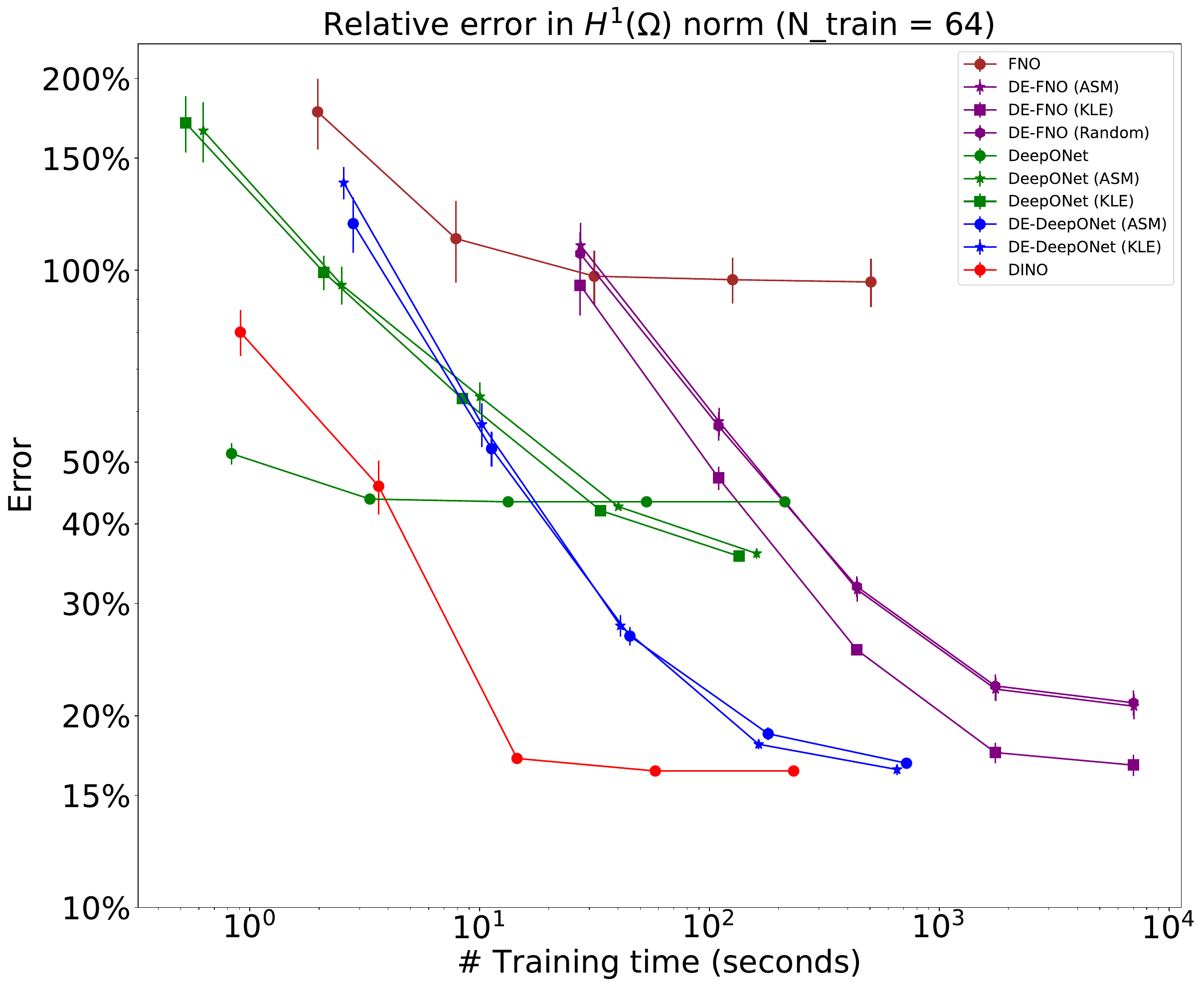}
                    \includegraphics[width=0.33\columnwidth]{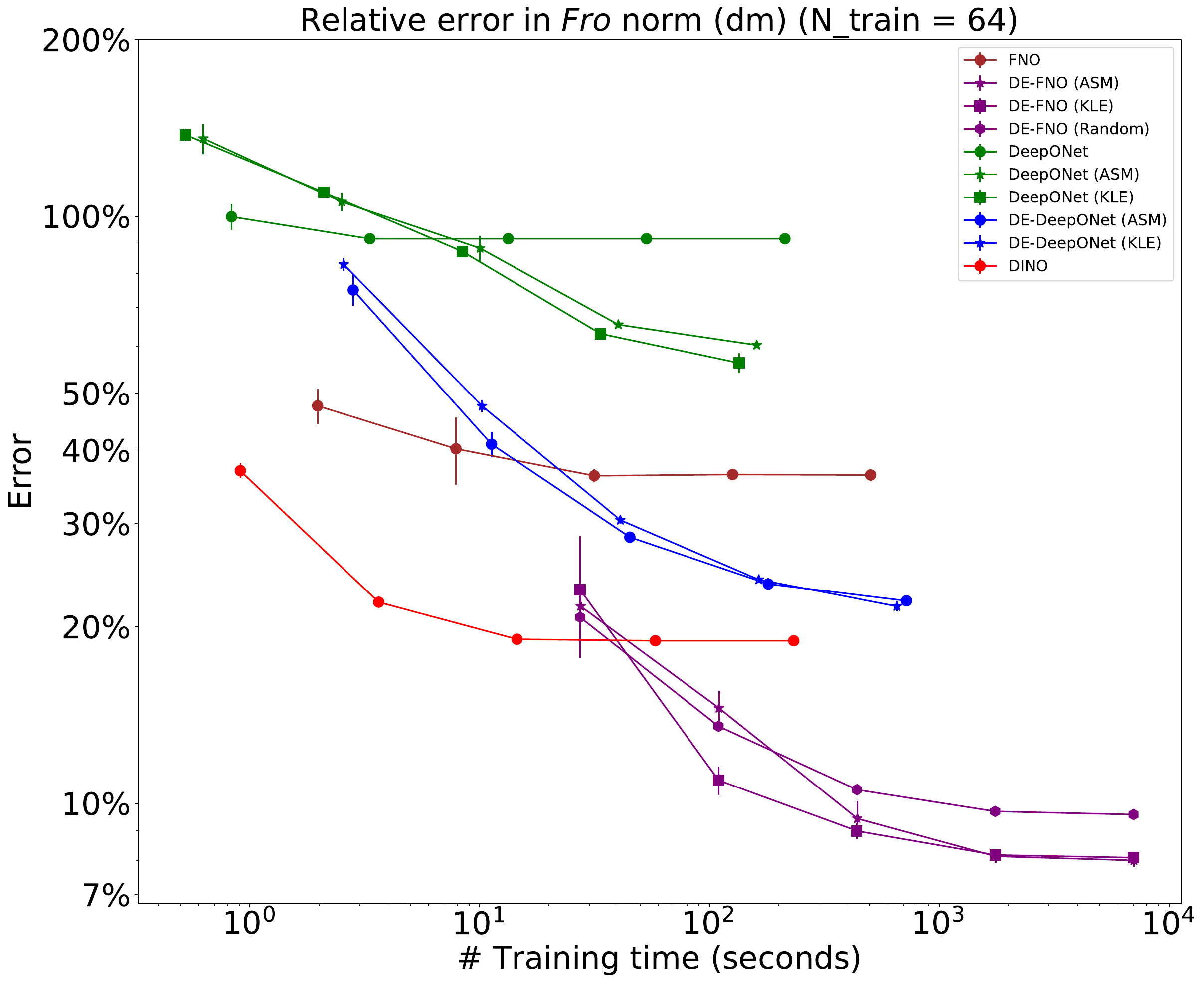}
        }
        \caption{Mean relative errors ($\pm$ standard deviation) over 5 trials versus model training time for the hyperelasticity equations when the number of training samples is 64.
        }
    \label{fig:error_plot_hyperelasticity_convergence_64}
    \end{center}
    \vskip -2em
    \end{figure}

\begin{figure}[H]
        \begin{center}
        \centerline{\includegraphics[width=0.33\columnwidth]{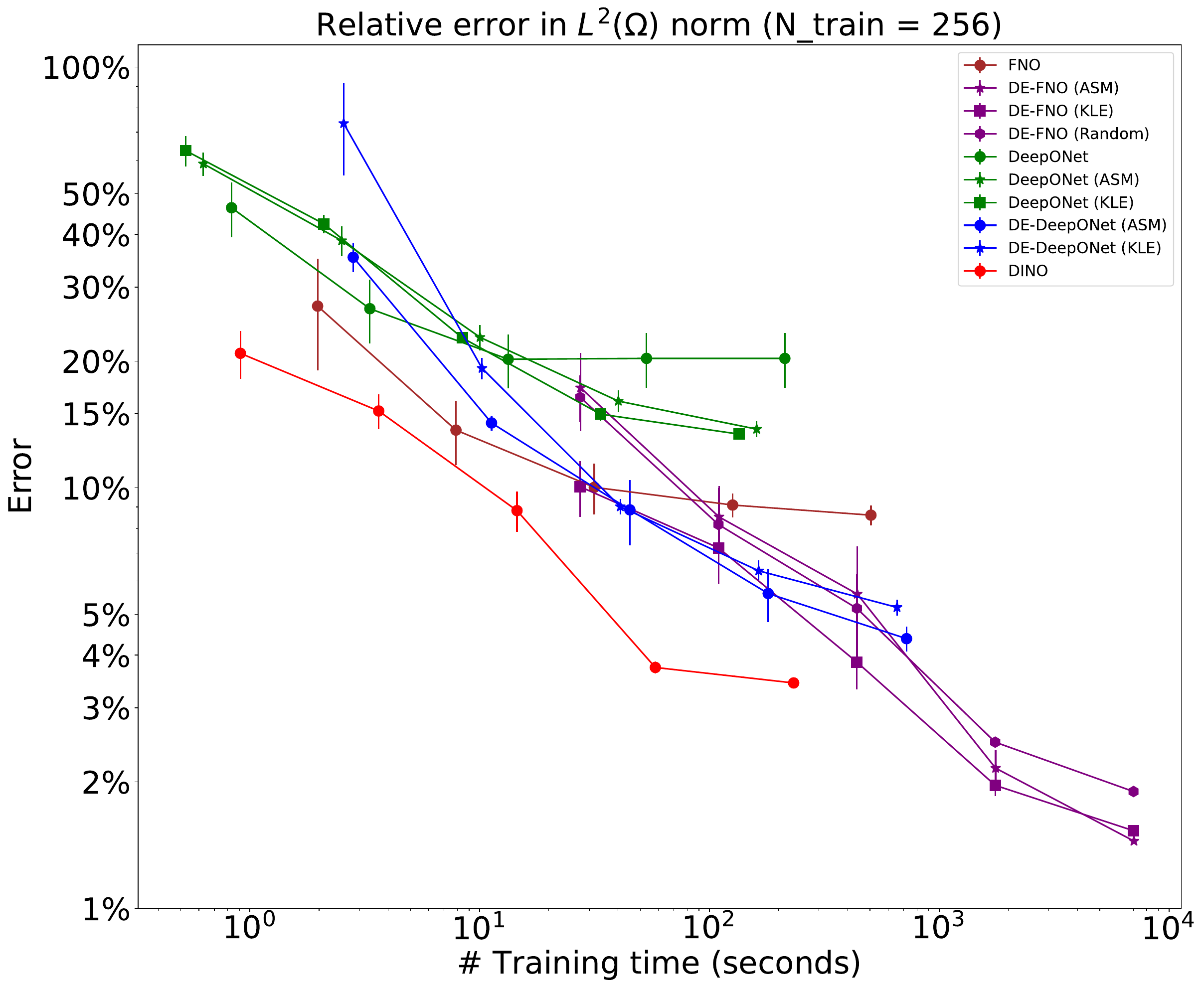}
                    \includegraphics[width=0.33\columnwidth]{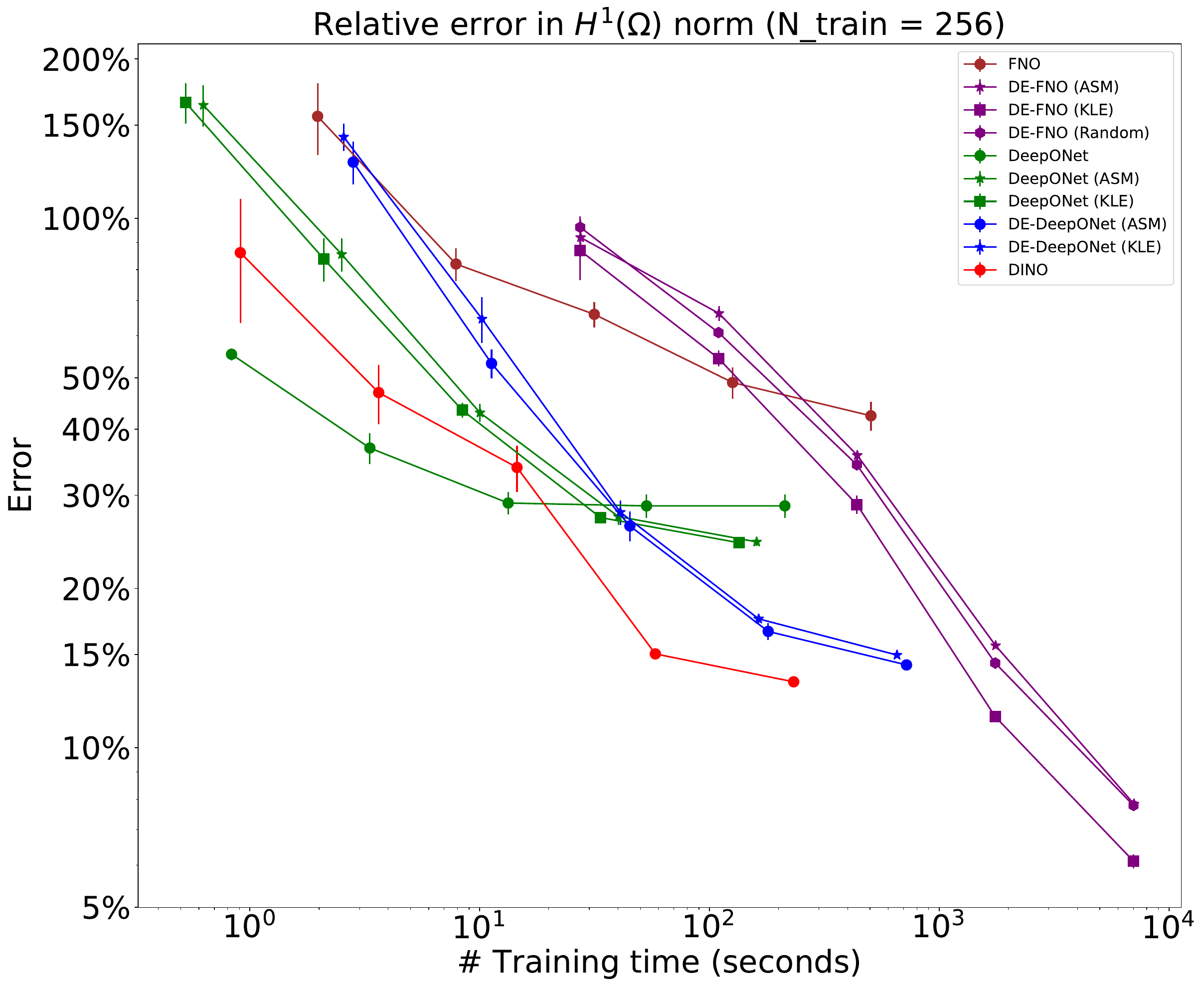}
                    \includegraphics[width=0.33\columnwidth]{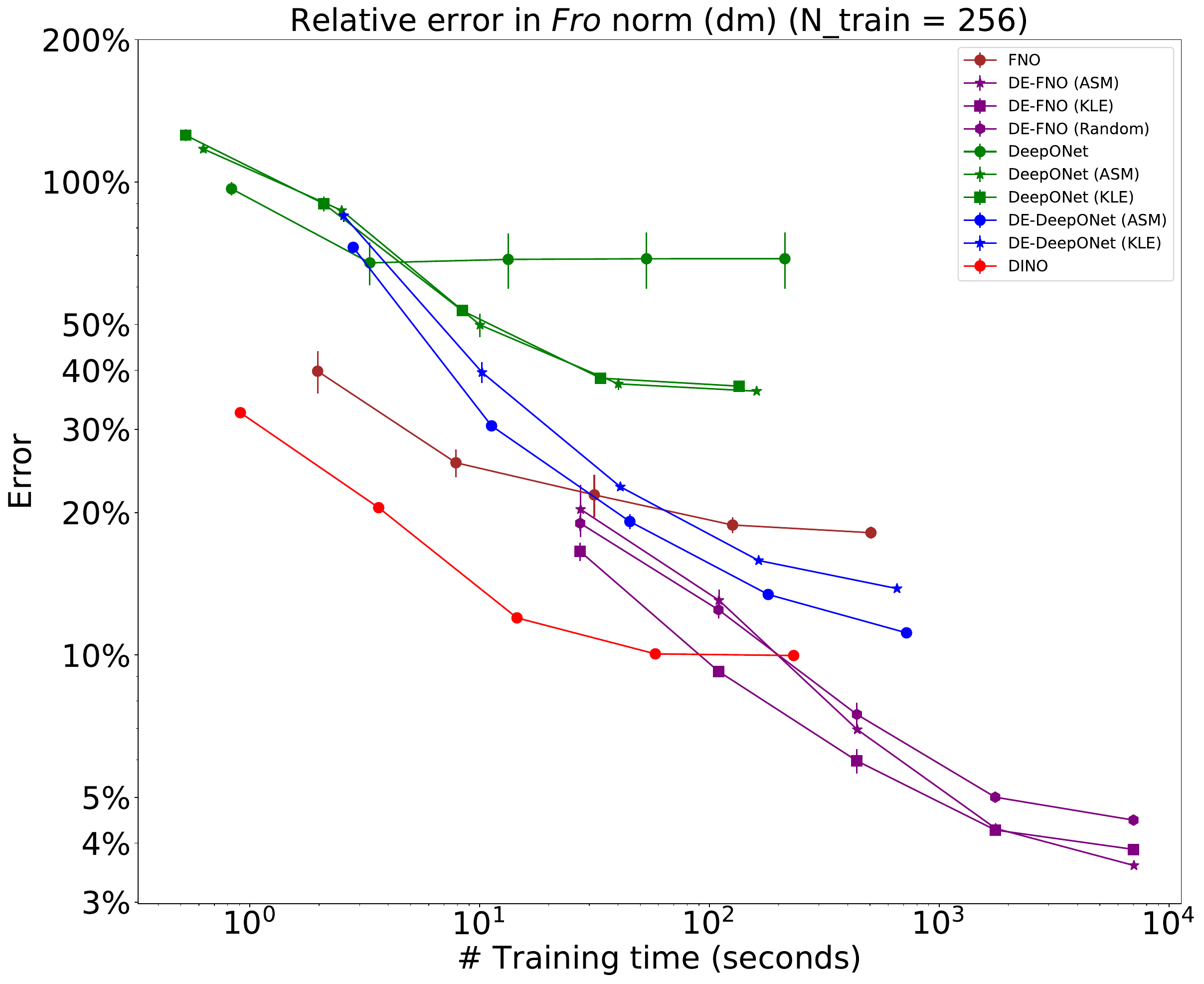}
        }
        \caption{Mean relative errors ($\pm$ standard deviation) over 5 trials versus model training time for the hyperelasticity equations when the number of training samples is 256.
        }
    \label{fig:error_plot_hyperelasticity_convergence_256}
    \end{center}
    \vskip -2em
    \end{figure}

\begin{figure}[H]
        \begin{center}
        \centerline{\includegraphics[width=0.33\columnwidth]{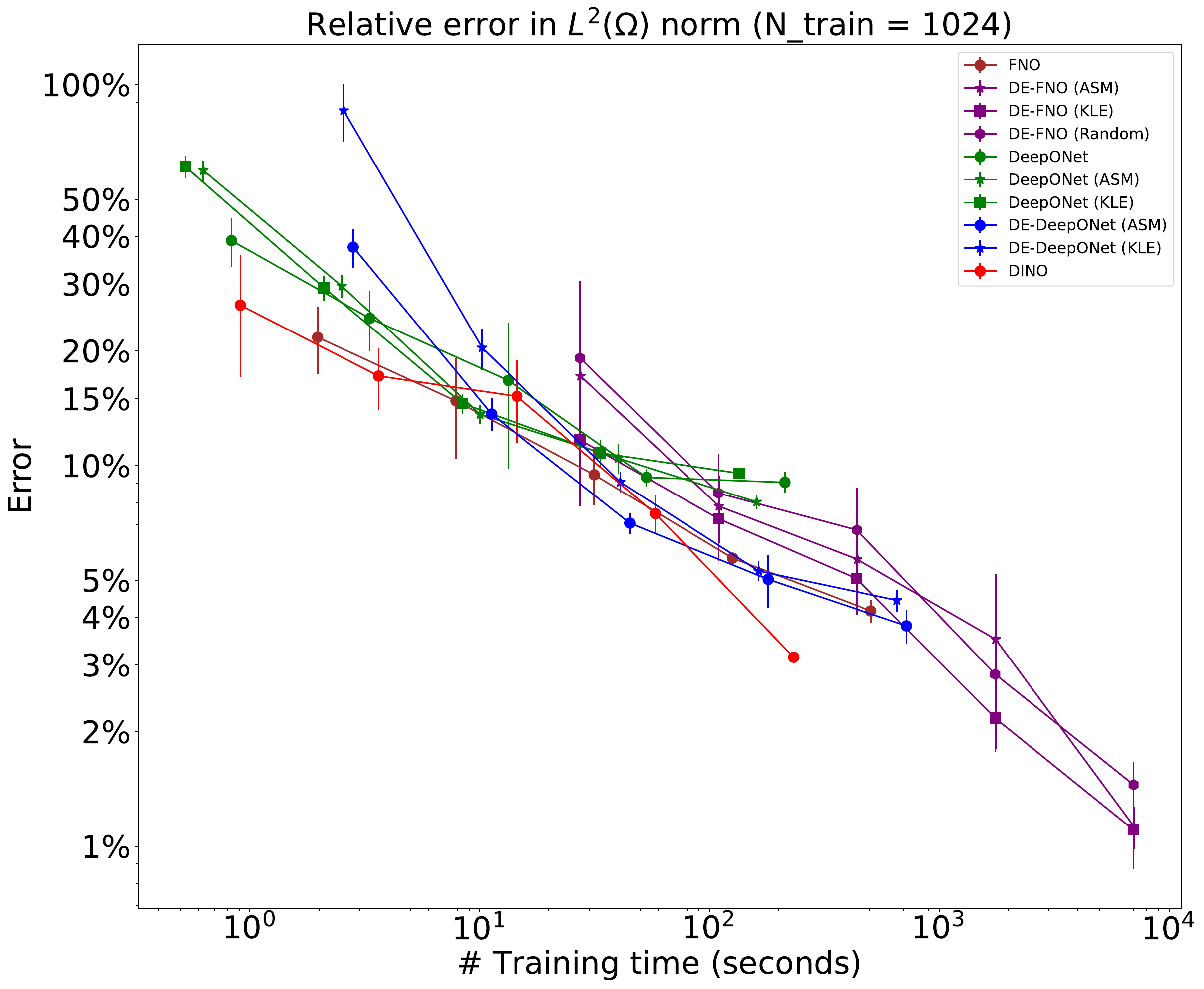}
                    \includegraphics[width=0.33\columnwidth]{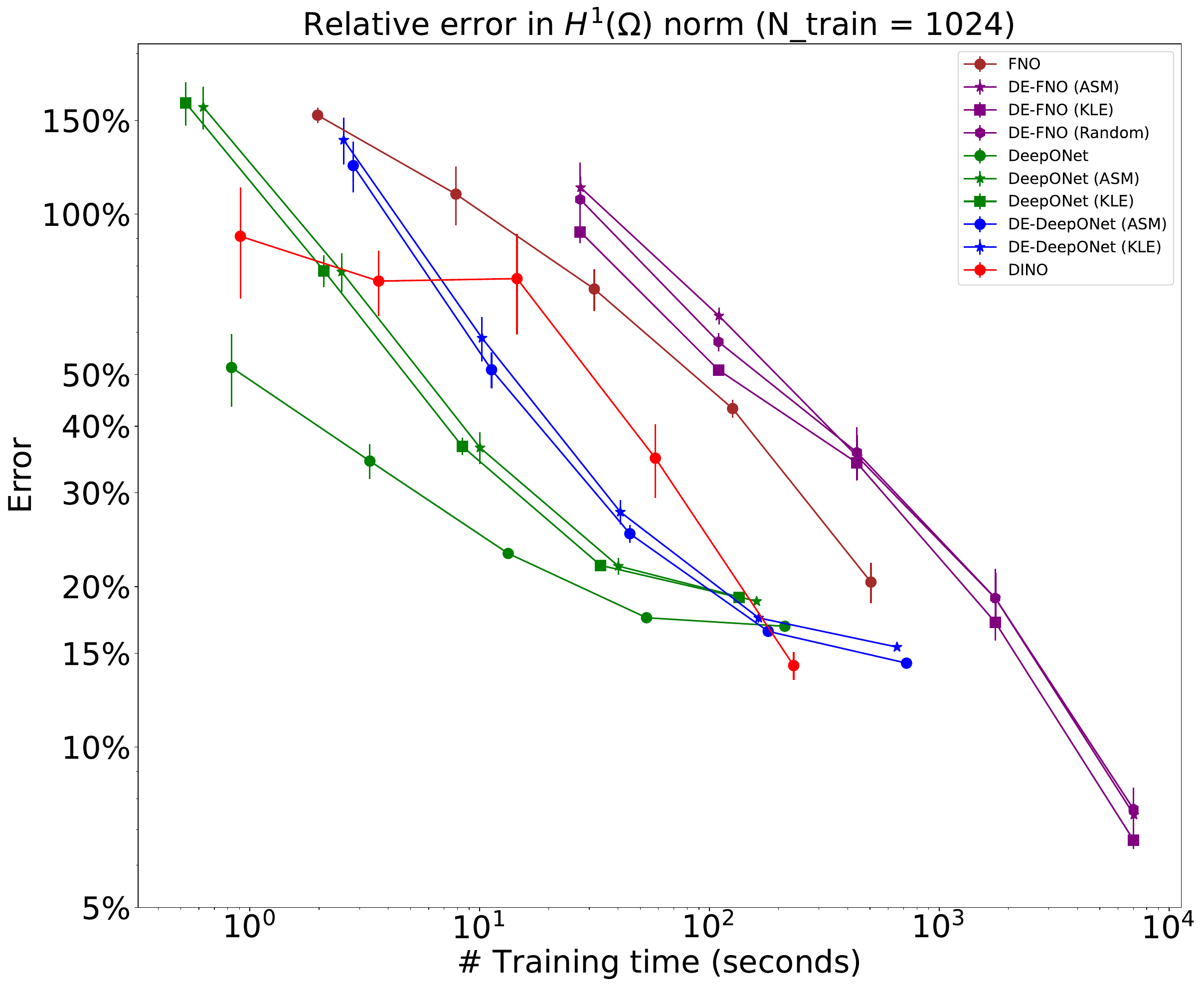}
                    \includegraphics[width=0.33\columnwidth]{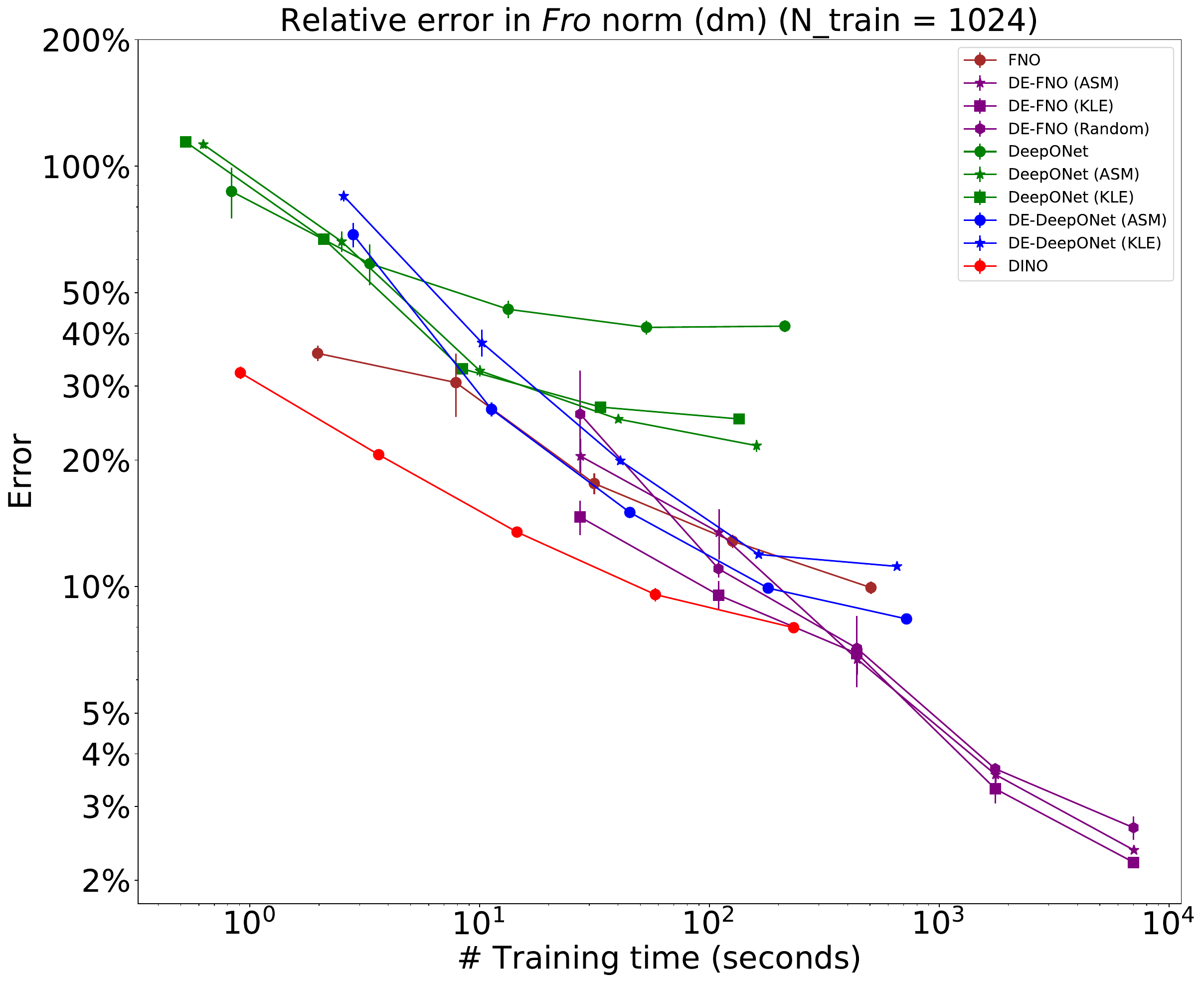}
        }
        \caption{Mean relative errors ($\pm$ standard deviation) over 5 trials versus model training time for the hyperelasticity equations when the number of training samples is 1024.
        }
    \label{fig:error_plot_hyperelasticity_convergence_1024}
    \end{center}
    \vskip -2em
    \end{figure}

\begin{figure}[H]
        \begin{center}
        \centerline{\includegraphics[width=0.33\columnwidth]{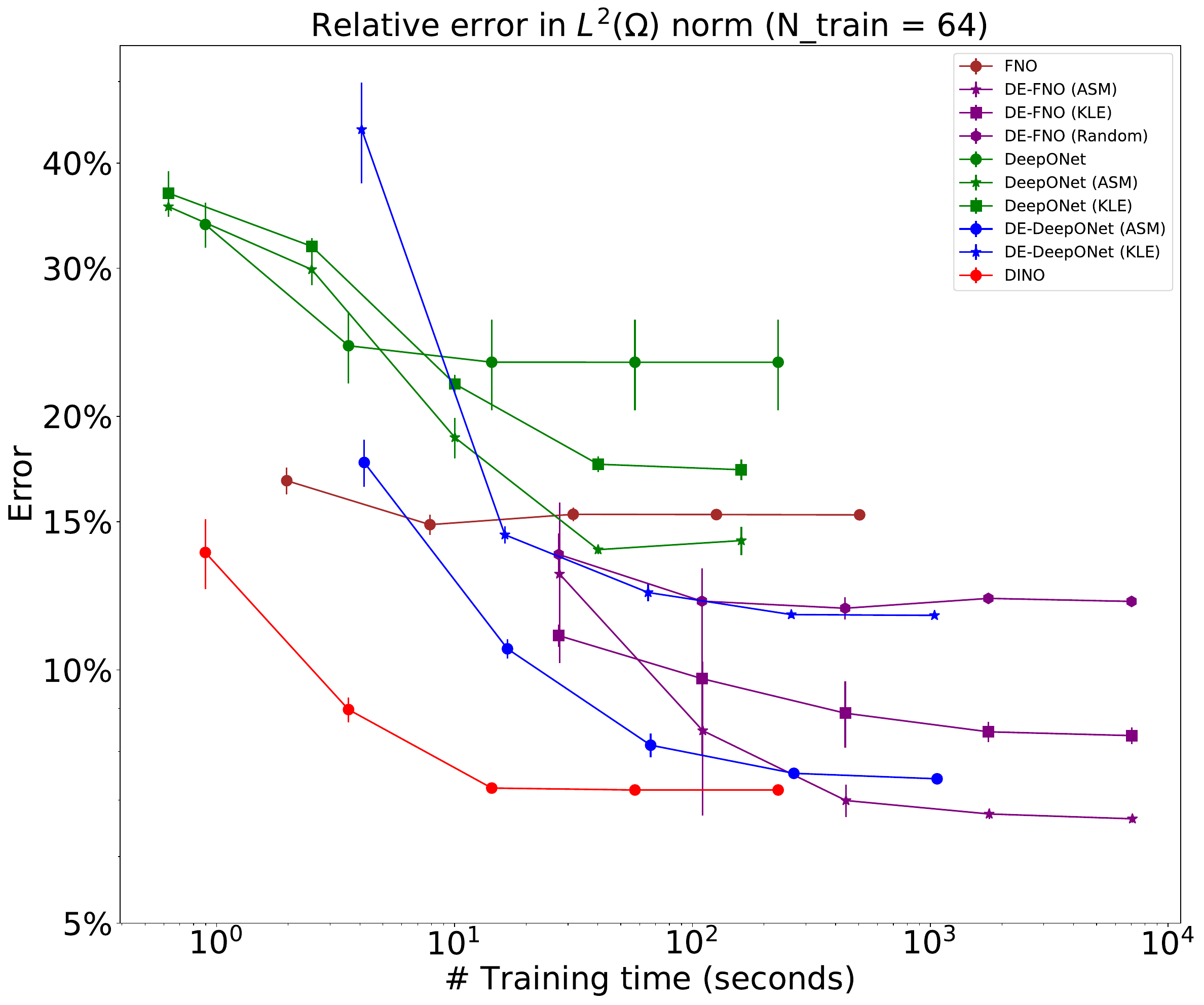}
                    \includegraphics[width=0.33\columnwidth]{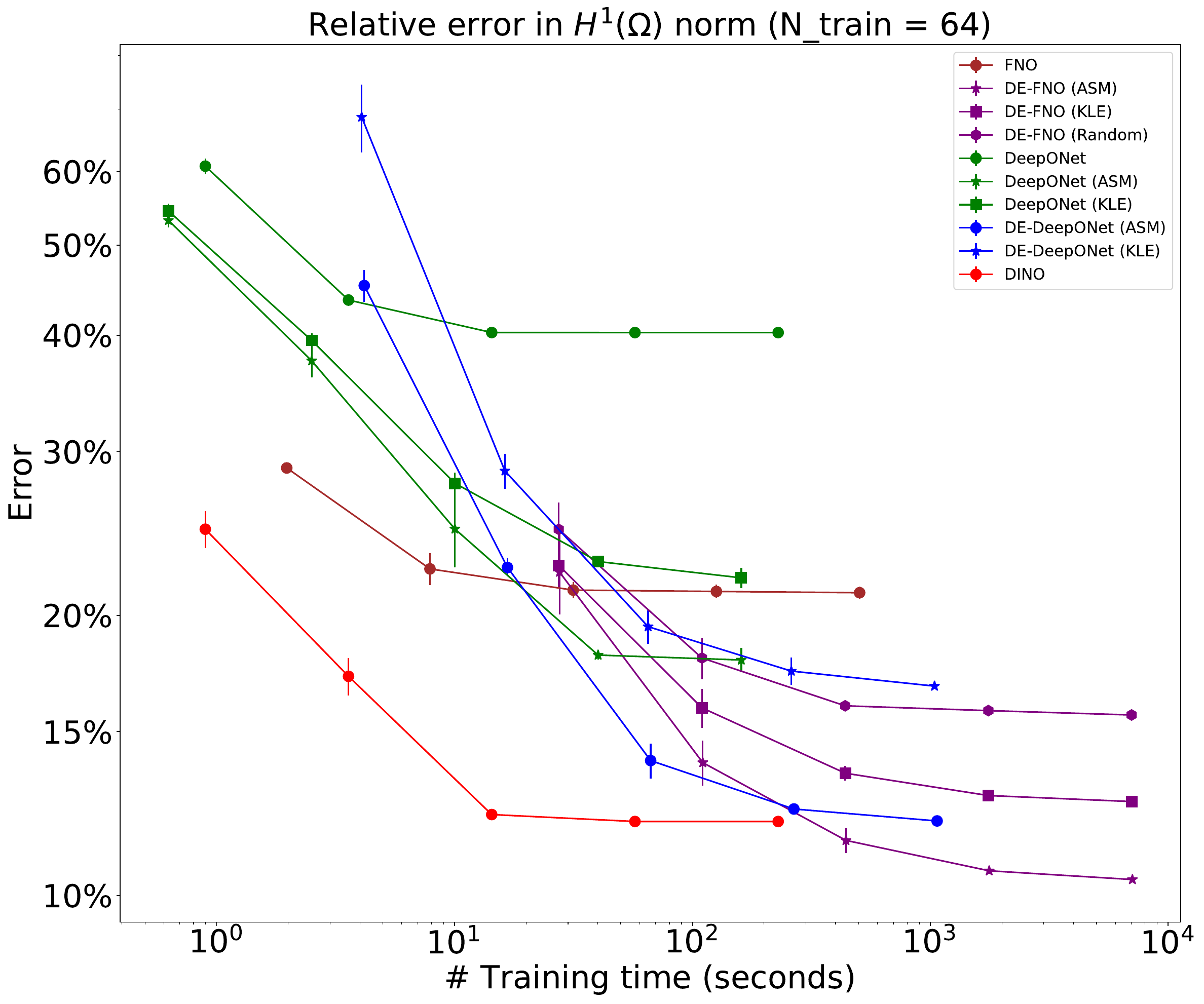}
                    \includegraphics[width=0.33\columnwidth]{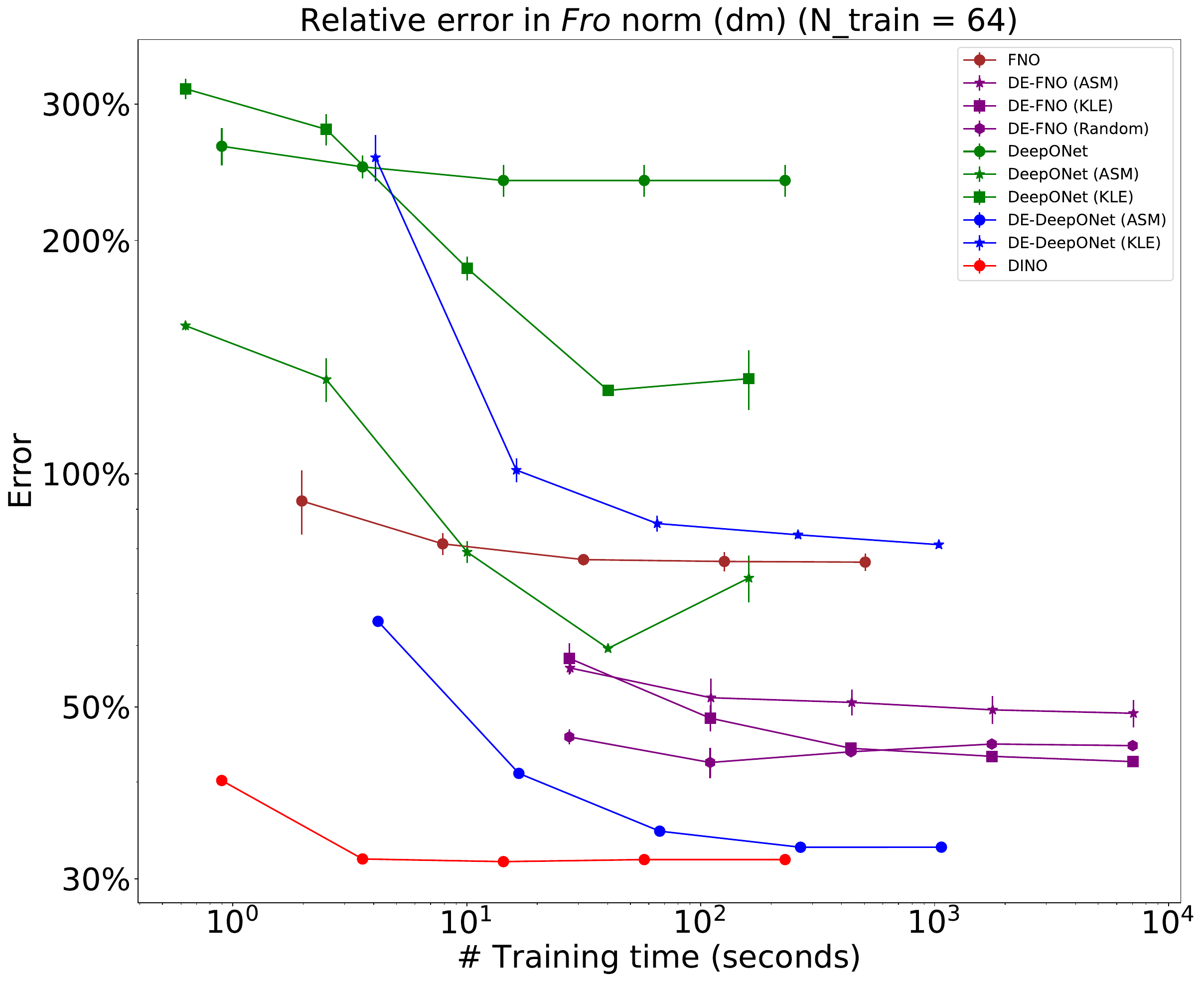}
        }
        \caption{Mean relative errors ($\pm$ standard deviation) over 5 trials versus model training time for the Navier--Stokes equations when the number of training samples is 64.
        }
    \label{fig:error_plot_navier_stokes_convergence_64}
    \end{center}
    \vskip -2em
    \end{figure}

\begin{figure}[H]
        \begin{center}
        \centerline{\includegraphics[width=0.33\columnwidth]{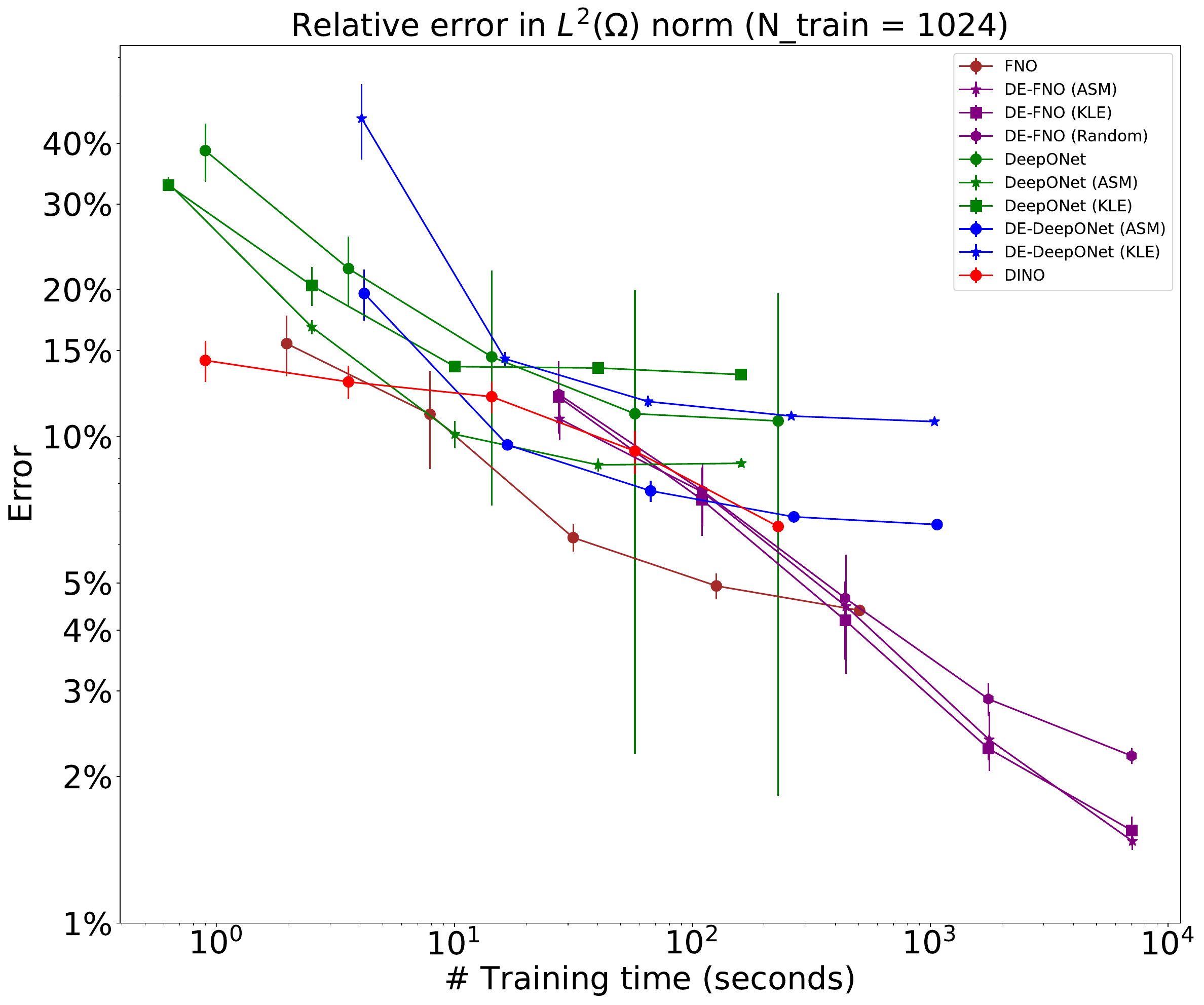}
                    \includegraphics[width=0.33\columnwidth]{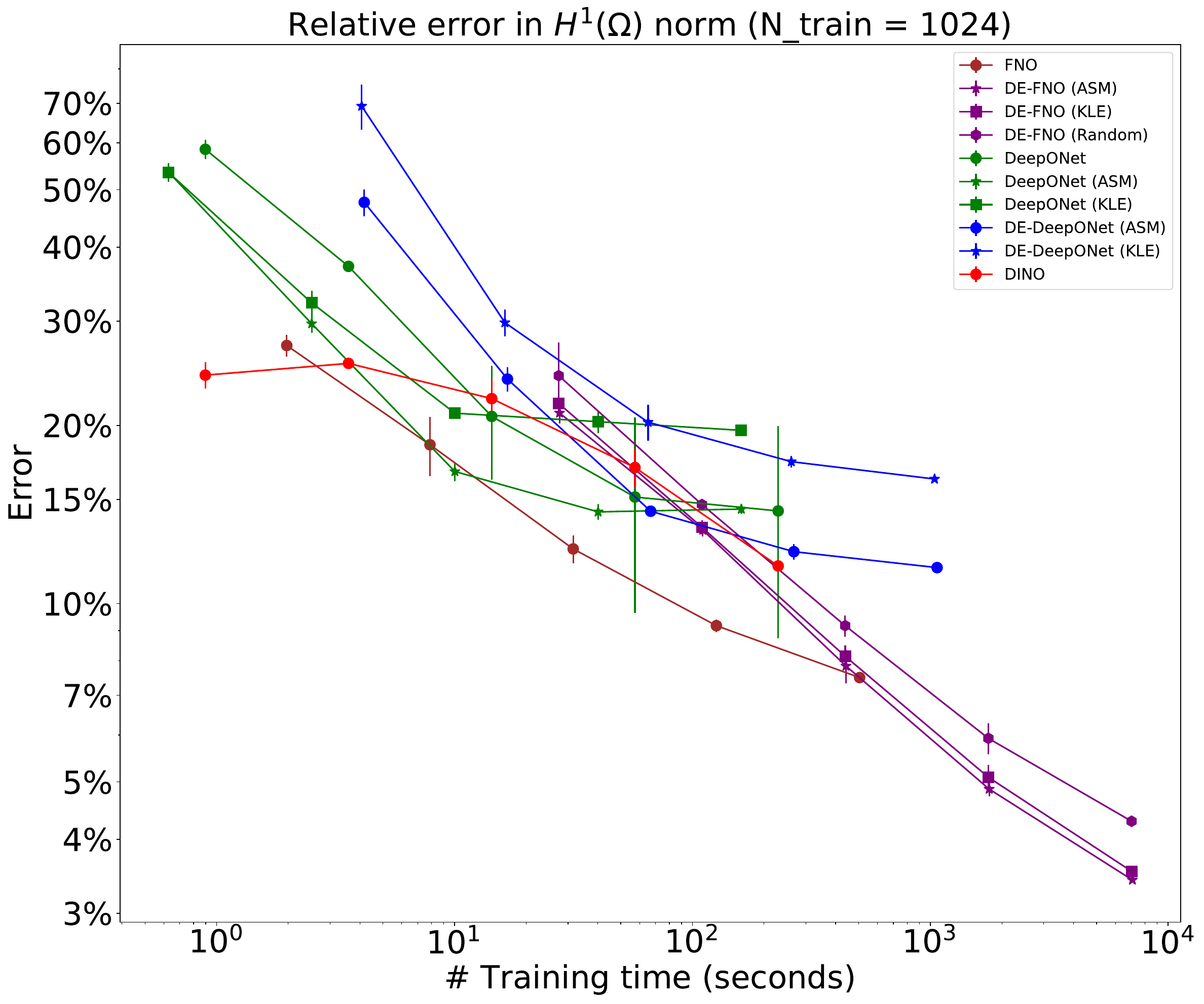}
                    \includegraphics[width=0.33\columnwidth]{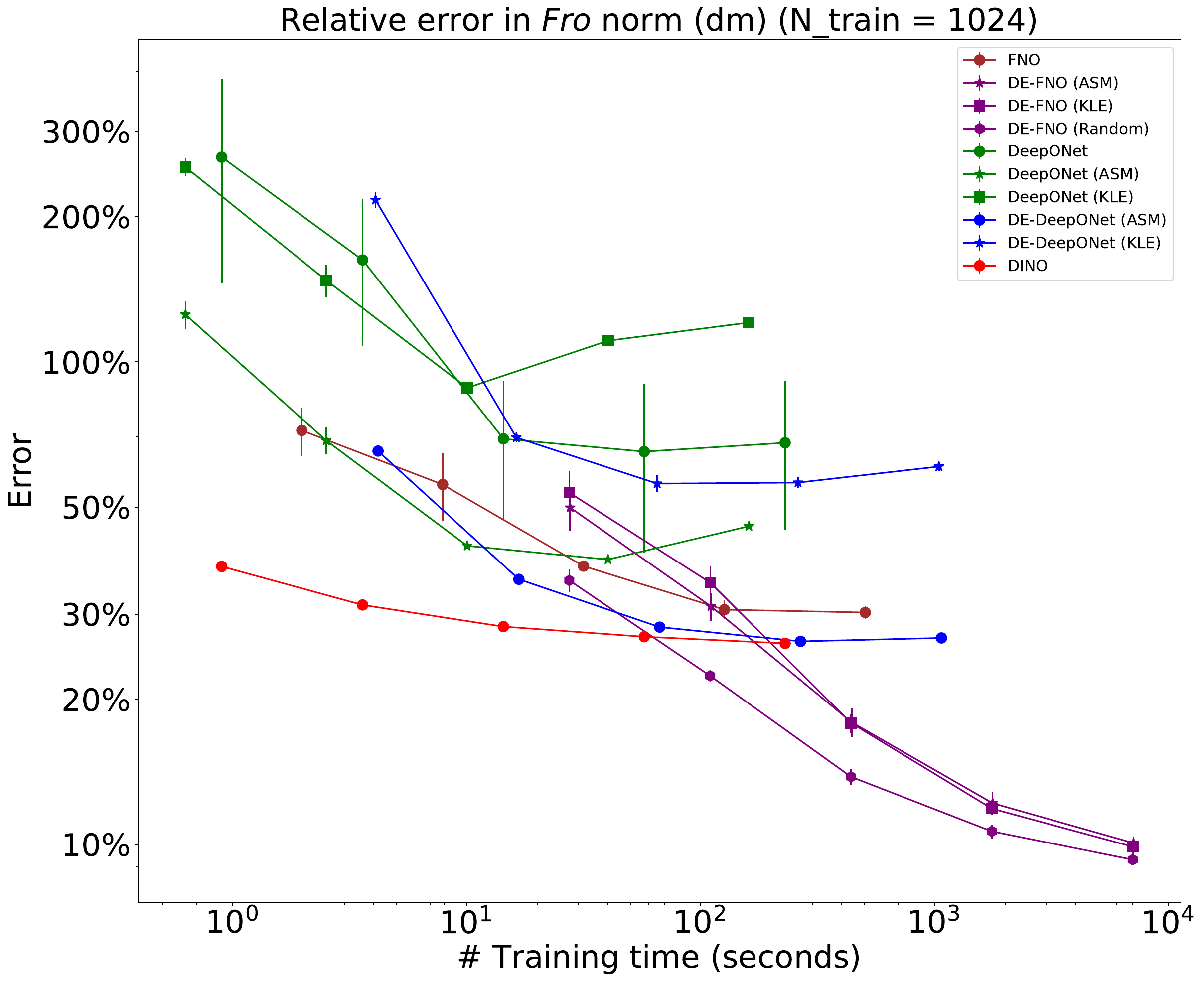}
        }
        \caption{Mean relative errors ($\pm$ standard deviation) over 5 trials versus model training time for the Navier--Stokes equations when the number of training samples is 1024.
        }
    \label{fig:error_plot_navier_stokes_convergence_1024}
    \end{center}
    \vskip -2em
    \end{figure}

   Based on the training time (seconds/iteration) of each model in~\cref{tab:computational_time_neural_network_training}, we obtain the convergence plots when the number of training samples is either limited ($N_{\text{train}}=16, 64$) or sufficient ($N_{\text{train}}=256, 1024$) in~\cref{fig:error_plot_hyperelasticity_convergence_16,fig:error_plot_hyperelasticity_convergence_64,fig:error_plot_hyperelasticity_convergence_256,fig:error_plot_hyperelasticity_convergence_1024} for the hyperelasticity equation, and in~\cref{fig:error_plot_navier_stokes_convergence_16_and_256,fig:error_plot_navier_stokes_convergence_64,fig:error_plot_navier_stokes_convergence_1024} for the Navier--Stokes equations.

%%%%%%%%%%%% Navier--Stokes %%%%%%%%%%%%

    \subsection{Output reconstruction error}
    \label{sec:output_reconstruction_error}
\begin{figure}[H]
        \begin{center}
        \centerline{
                    \includegraphics[width=0.5\columnwidth]{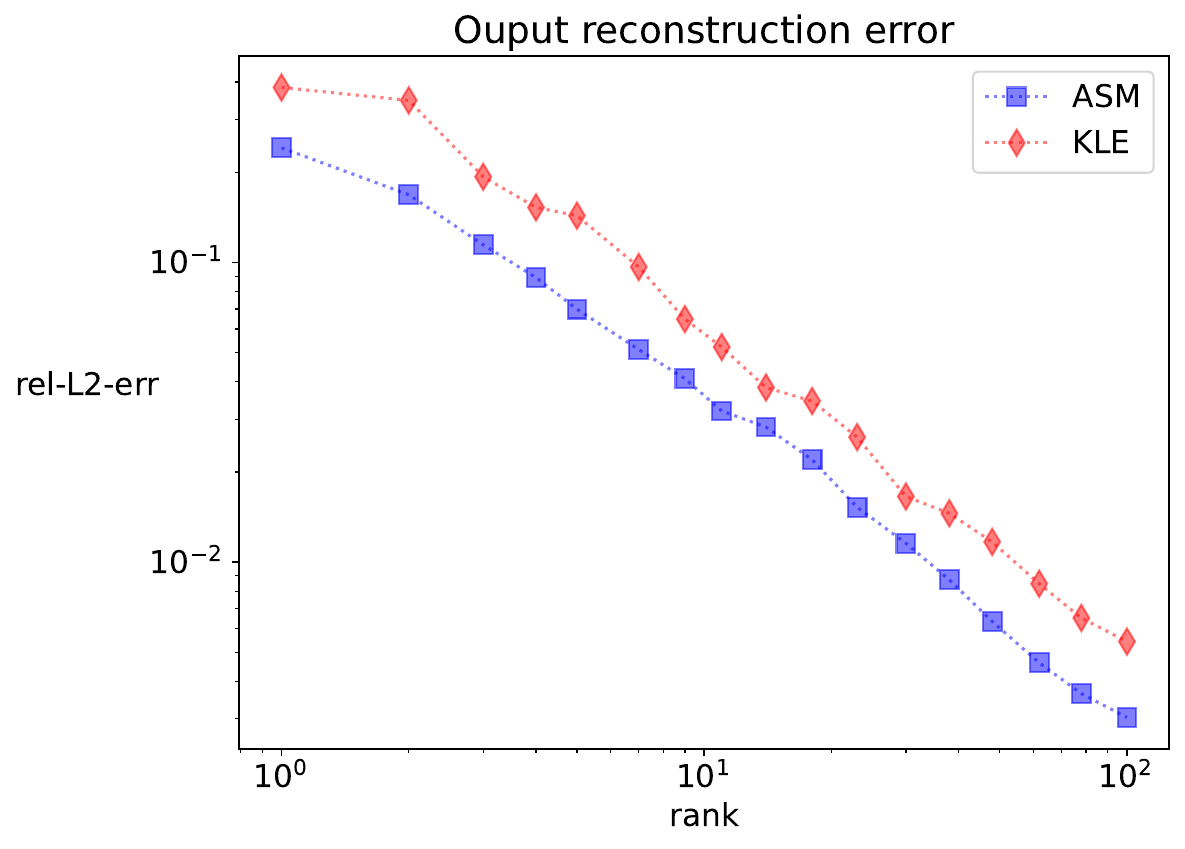}
                    \includegraphics[width=0.5\columnwidth]{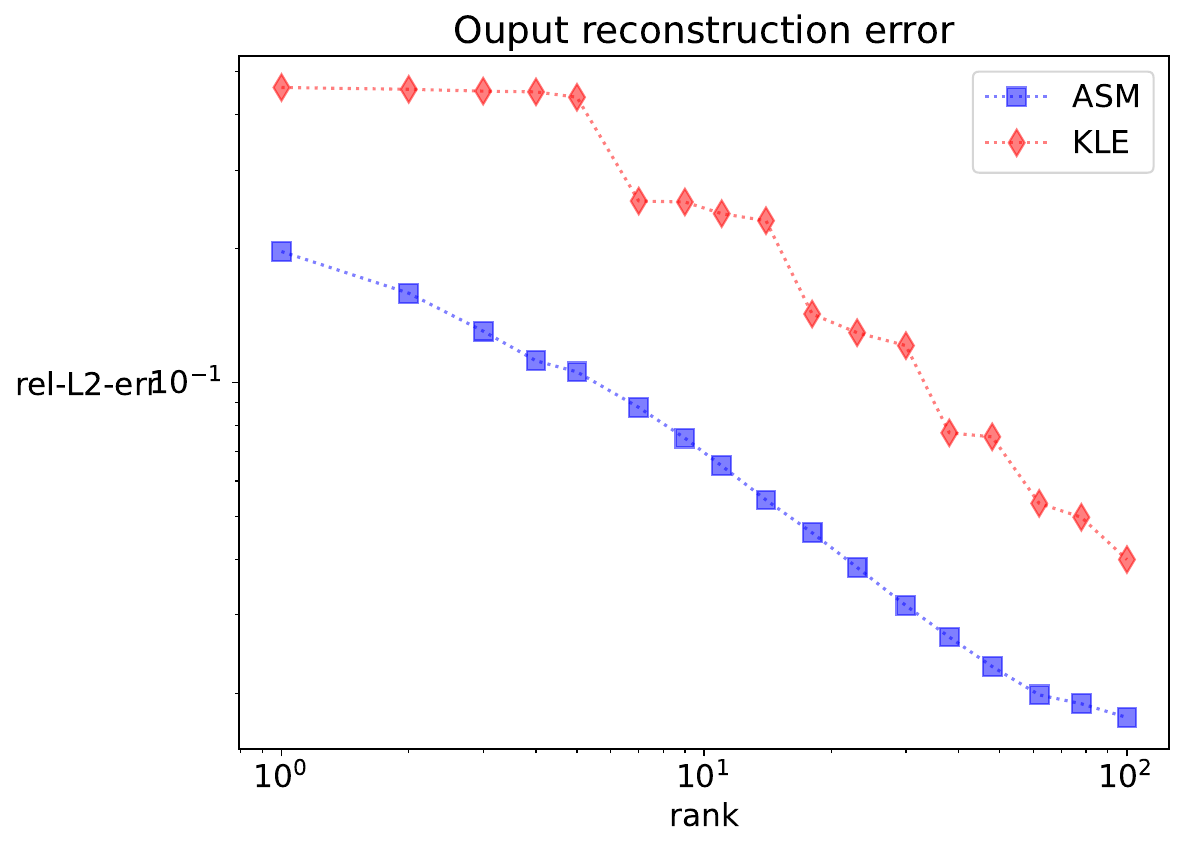}
        }
        \caption{Output reconstruction error using KLE and ASM basis. Left: Hyperelasticity; Right: Navier--Stokes
        }
    \label{fig:output_reconstruction_error}
    \end{center}
    \vskip -2em
    \end{figure}

    To measure the error induced by the projection, we define the output reconstruction error as follows
    \begin{align*}
        \frac{1}{N}\sum_{i=1}^{N}\frac{\|u(P_rm^{(i)})-u(m^{(i)})\|_{L^2}}{\|u(m^{(i)})\|_{L^2}}, 
    \end{align*}
    where $P_r$ is the rank $r$ linear projector. We provide the plots of the output reconstruction error vs number of reduced basis $r$ using KLE basis and ASM basis in~\cref{fig:output_reconstruction_error}. We can see that using the ASM basis results in a lower output reconstruction error than the KLE basis.

    \subsection{Visualization of the ground truth and prediction}
    \label{sec:visualization}

\begin{figure}[H]
        \begin{center}
            \includegraphics[scale=0.45]{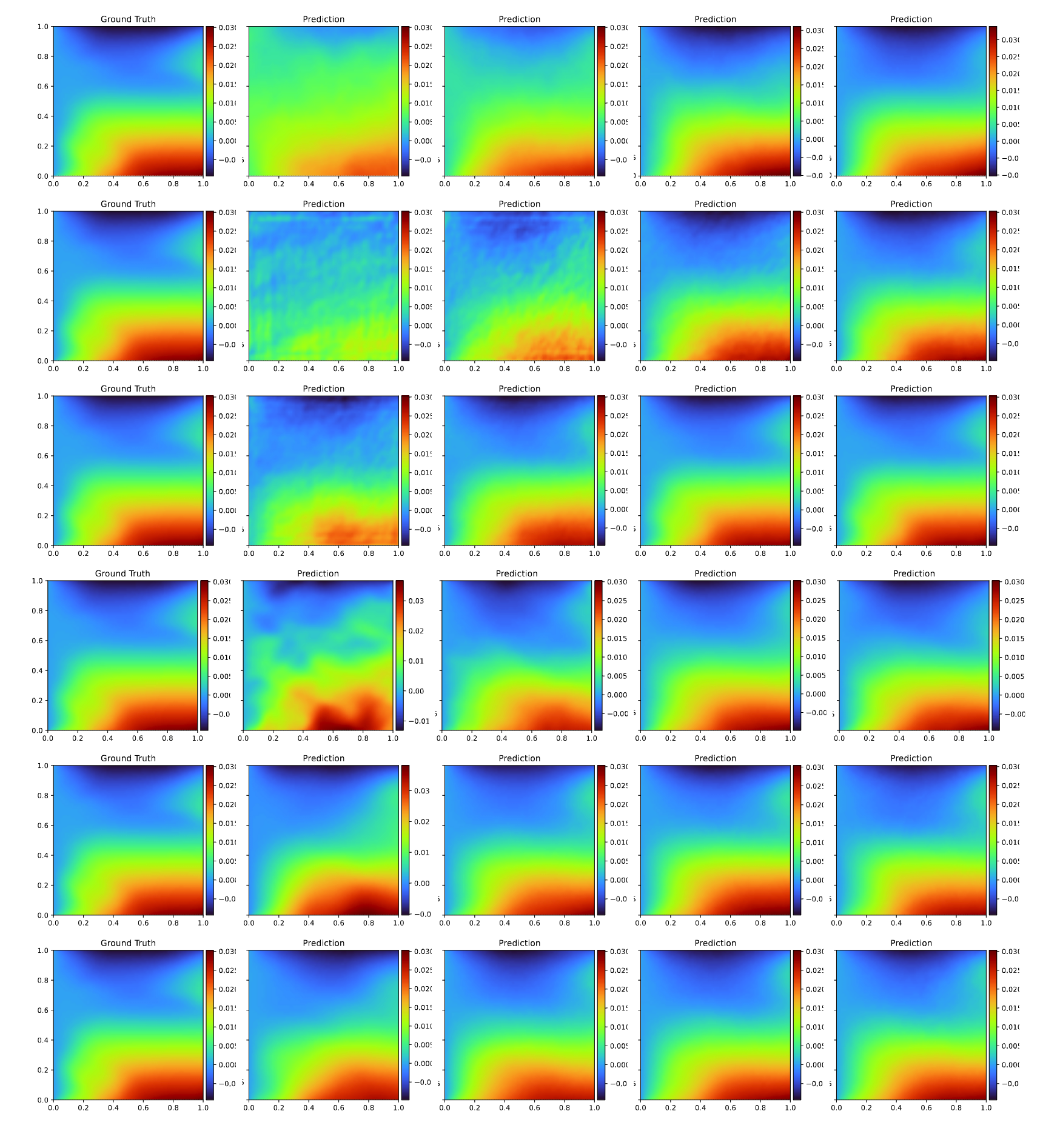}
        \end{center}
        \vskip -0.1in
        \caption{Hyperelasticity. Comparison of the predictions of $u_1$ with (16, 64, 256, 1024) training samples using different methods. Top row --> bottom row: DeepONet, FNO, DE-FNO (Random), DINO, DE-DeepONet (KLE), DE-DeepONet (ASM).}
        \label{fig:truth_prediction_hyperelasticity_u1}
    \end{figure}

\begin{figure}[H]
        \begin{center}
            \includegraphics[scale=0.45]{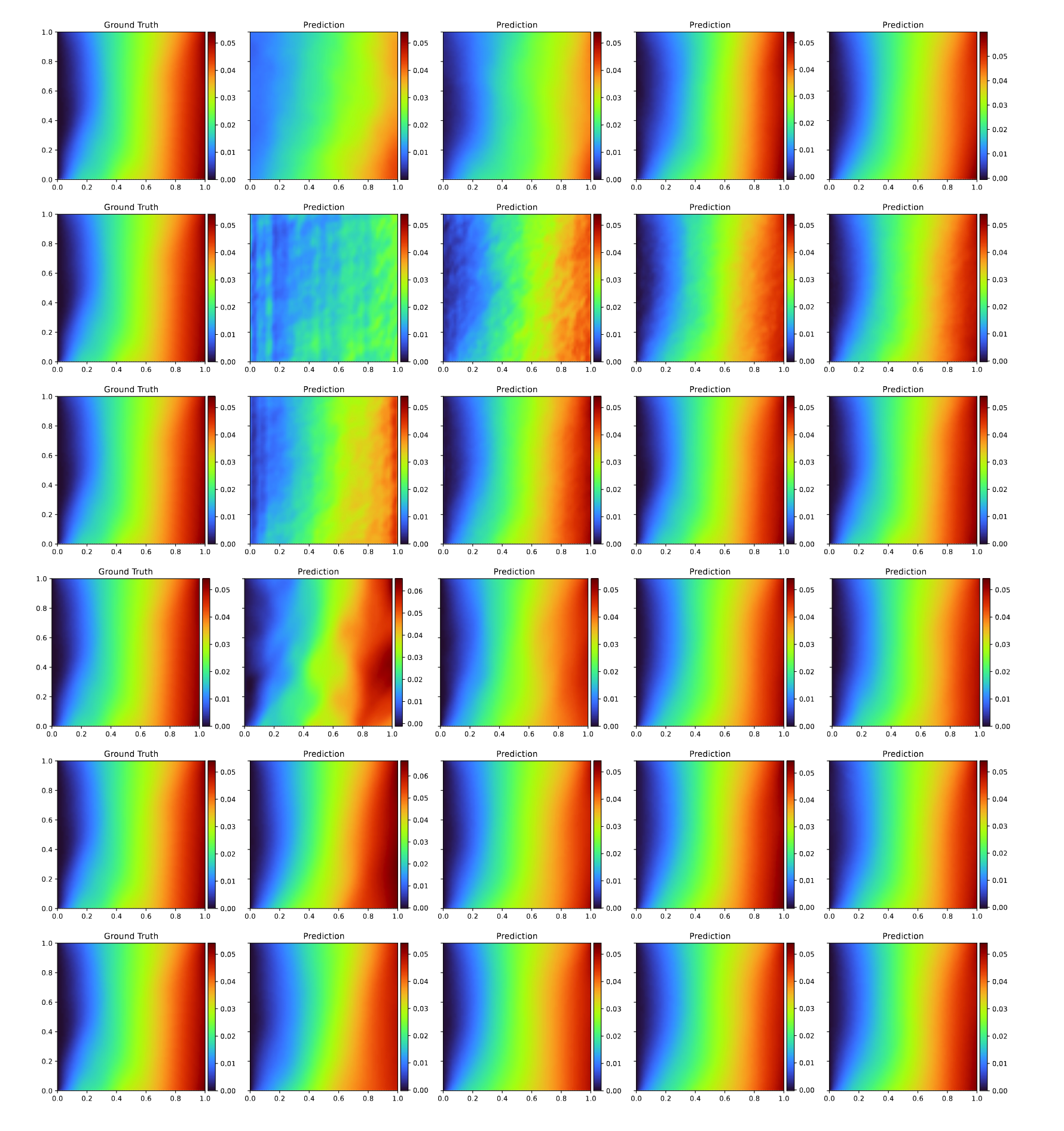}
        \end{center}
        \vskip -0.1in
        \caption{Hyperelasticity. Comparison of the predictions of $u_2$ with (16, 64, 256, 1024) training samples using different methods. Top row --> bottom row: DeepONet, FNO, DE-FNO (Random), DINO, DE-DeepONet (KLE), DE-DeepONet (ASM).}
        \label{fig:truth_prediction_hyperelasticity_u2}
    \end{figure}

\begin{figure}[H]
        \begin{center}
            \includegraphics[scale=0.45]{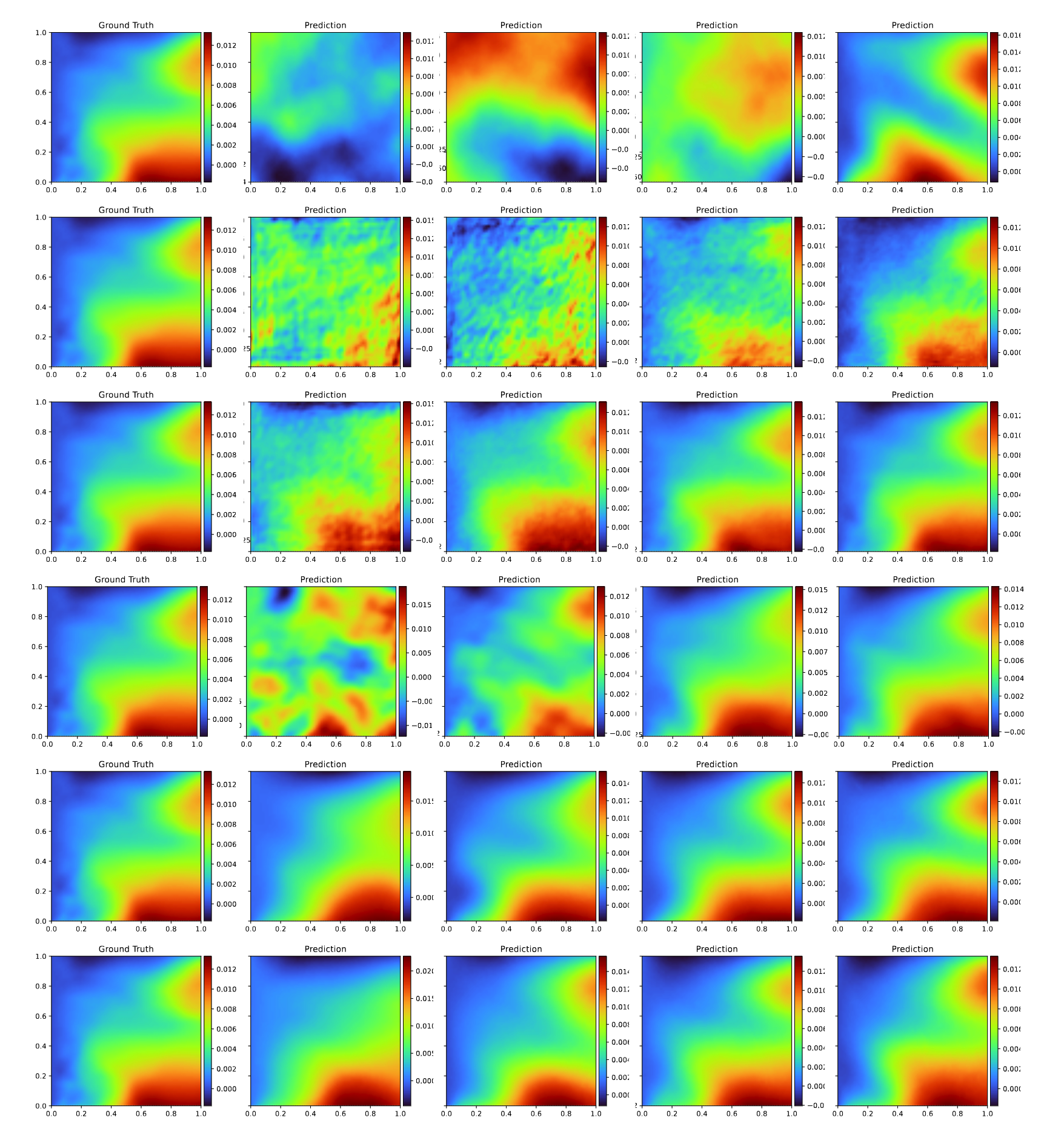}
        \end{center}
        \caption{Hyperelasticity. Comparison of the predictions of directional derivative $du_1(m;\omega_1)$ with (16, 64, 256, 1024) training samples using different methods. Top row --> bottom row: DeepONet, FNO, DE-FNO (Random), DINO, DE-DeepONet (KLE), DE-DeepONet (ASM).}
        \label{fig:truth_prediction_hyperelasticity_dm1}
    \end{figure}

\begin{figure}[H]
        \begin{center}
            \includegraphics[scale=0.45]{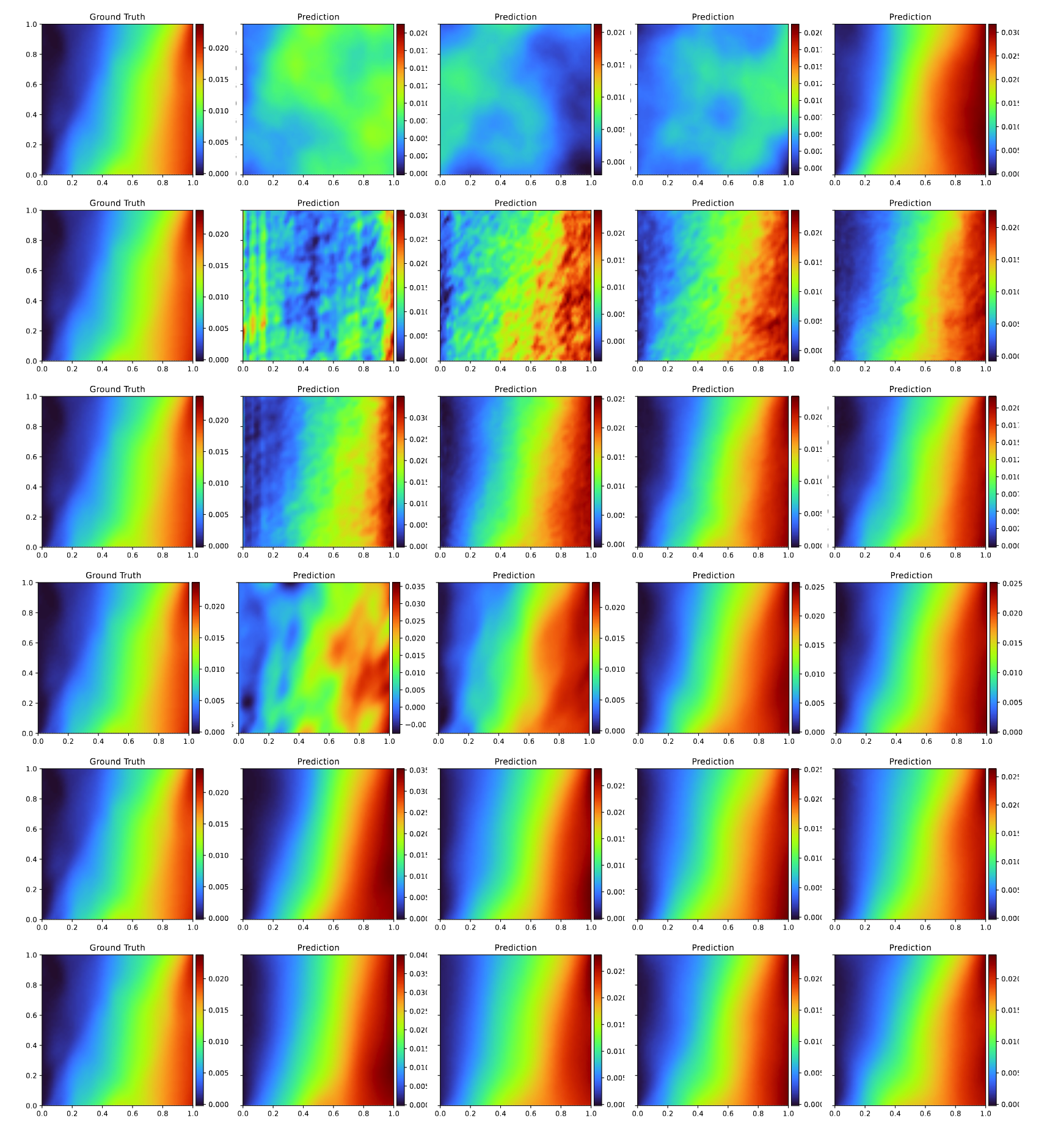}
        \end{center}
        \caption{Hyperelasticity. Comparison of the predictions of directional derivative $du_2(m;\omega_1)$ with (16, 64, 256, 1024) training samples using different methods.Top row --> bottom row: DeepONet, FNO, DE-FNO (Random), DINO, DE-DeepONet (KLE), DE-DeepONet (ASM).}
        \label{fig:truth_prediction_hyperelasticity_dm2}
    \end{figure}

\begin{figure}[H]
        \begin{center}
            \includegraphics[scale=0.45]{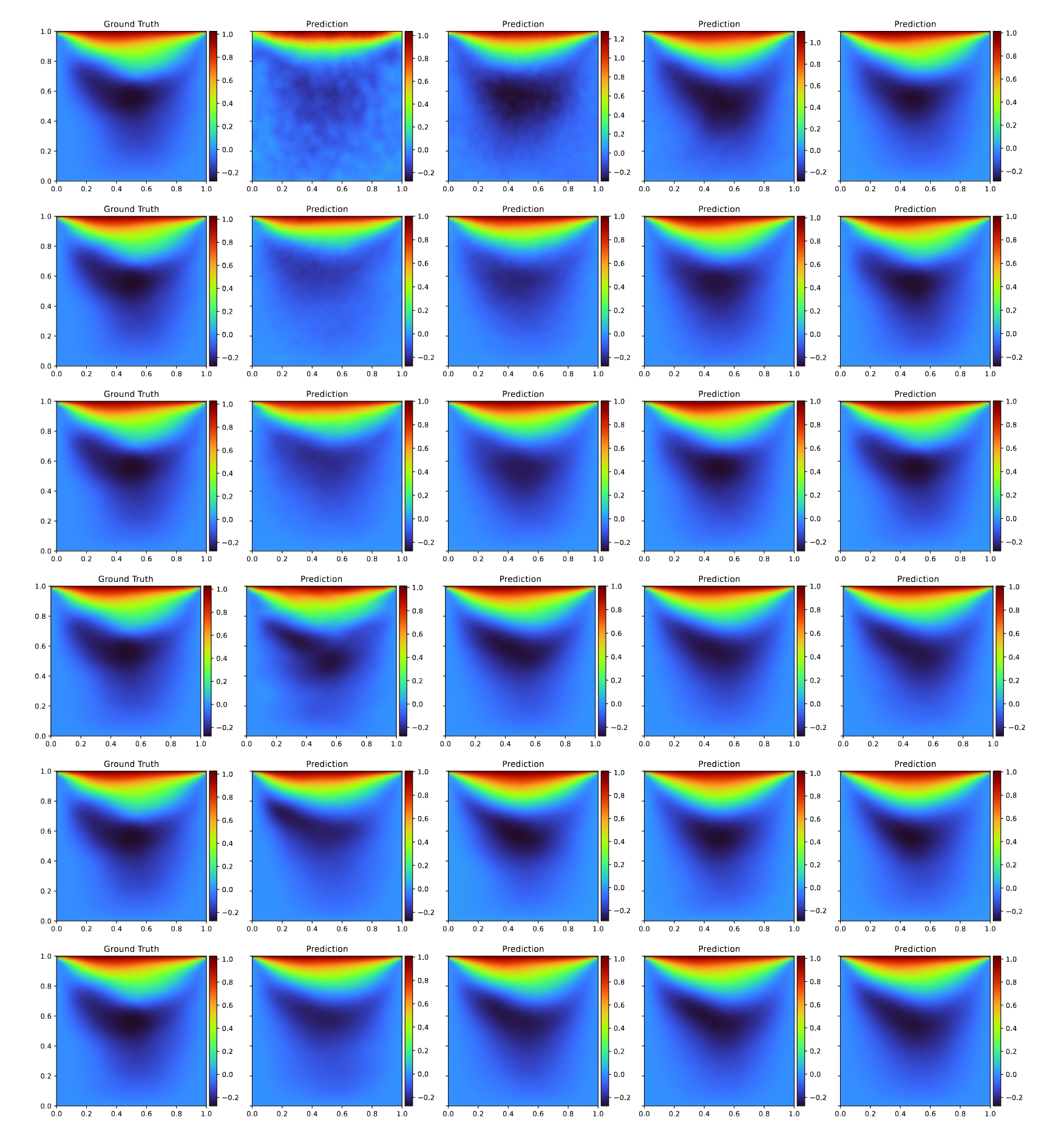}
        \end{center}
        \vskip -0.1in
        \caption{Navier--Stokes. Comparison of the predictions of velocity-x with (16, 64, 256, 1024) training samples using different methods. Top row --> bottom row: DeepONet, FNO, DE-FNO (Random), DINO, DE-DeepONet (KLE), DE-DeepONet (ASM).}
        \label{fig:truth_prediction_navier_stokes_velocity_x}
    \end{figure}

\begin{figure}[H]
        \begin{center}
            \includegraphics[scale=0.45]{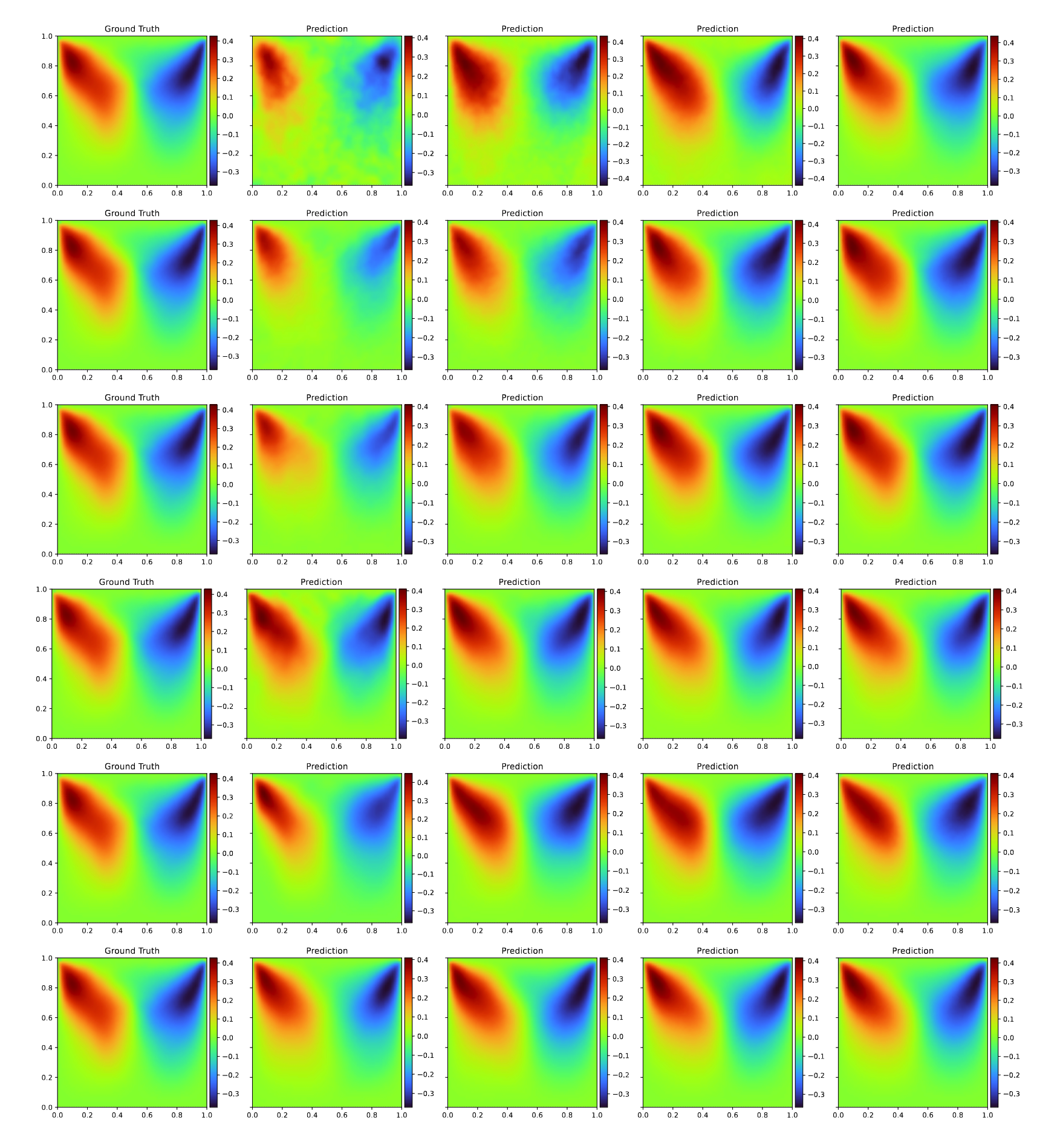}
        \end{center}
        \vskip -0.1in
        \caption{Navier--Stokes. Comparison of the predictions of velocity-y with (16, 64, 256, 1024) training samples using different methods. Top row --> bottom row: DeepONet, FNO, DE-FNO (Random), DINO, DE-DeepONet (KLE), DE-DeepONet (ASM).}
        \label{fig:truth_prediction_navier_stokes_velocity_y}
    \end{figure}

\begin{figure}[H]
        \begin{center}
            \includegraphics[scale=0.45]{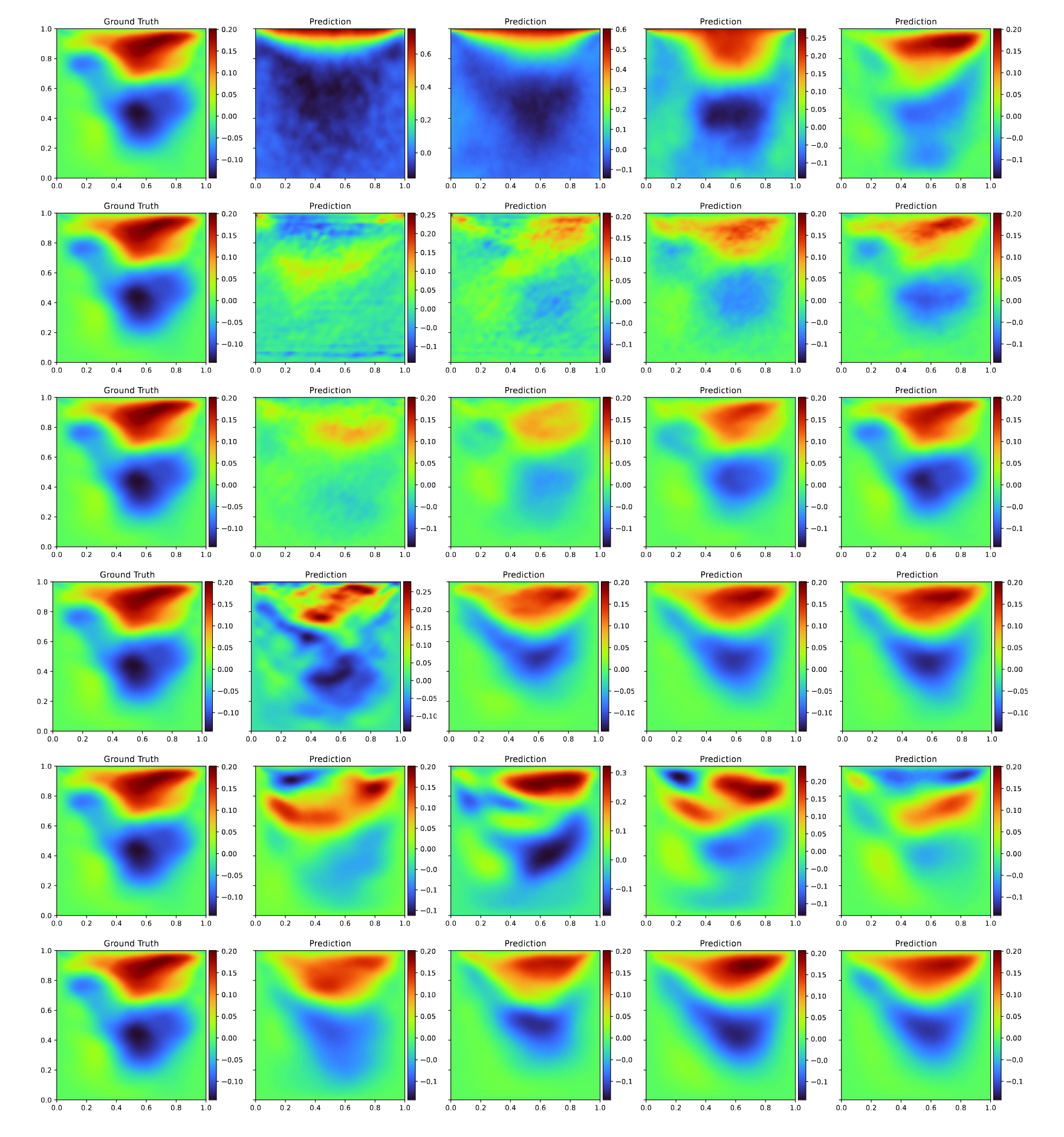}
        \end{center}
        \caption{Navier--Stokes. Comparison of the predictions of directional derivative $du_1(m;\omega_1)$ with (16, 64, 256, 1024) training samples using different methods. Top row --> bottom row: DeepONet, FNO, DE-FNO (Random), DINO, DE-DeepONet (KLE), DE-DeepONet (ASM).}
        \label{fig:truth_prediction_navier_stokes_velocity_x_dm}
    \end{figure}

\begin{figure}[H]
        \begin{center}
            \includegraphics[scale=0.45]{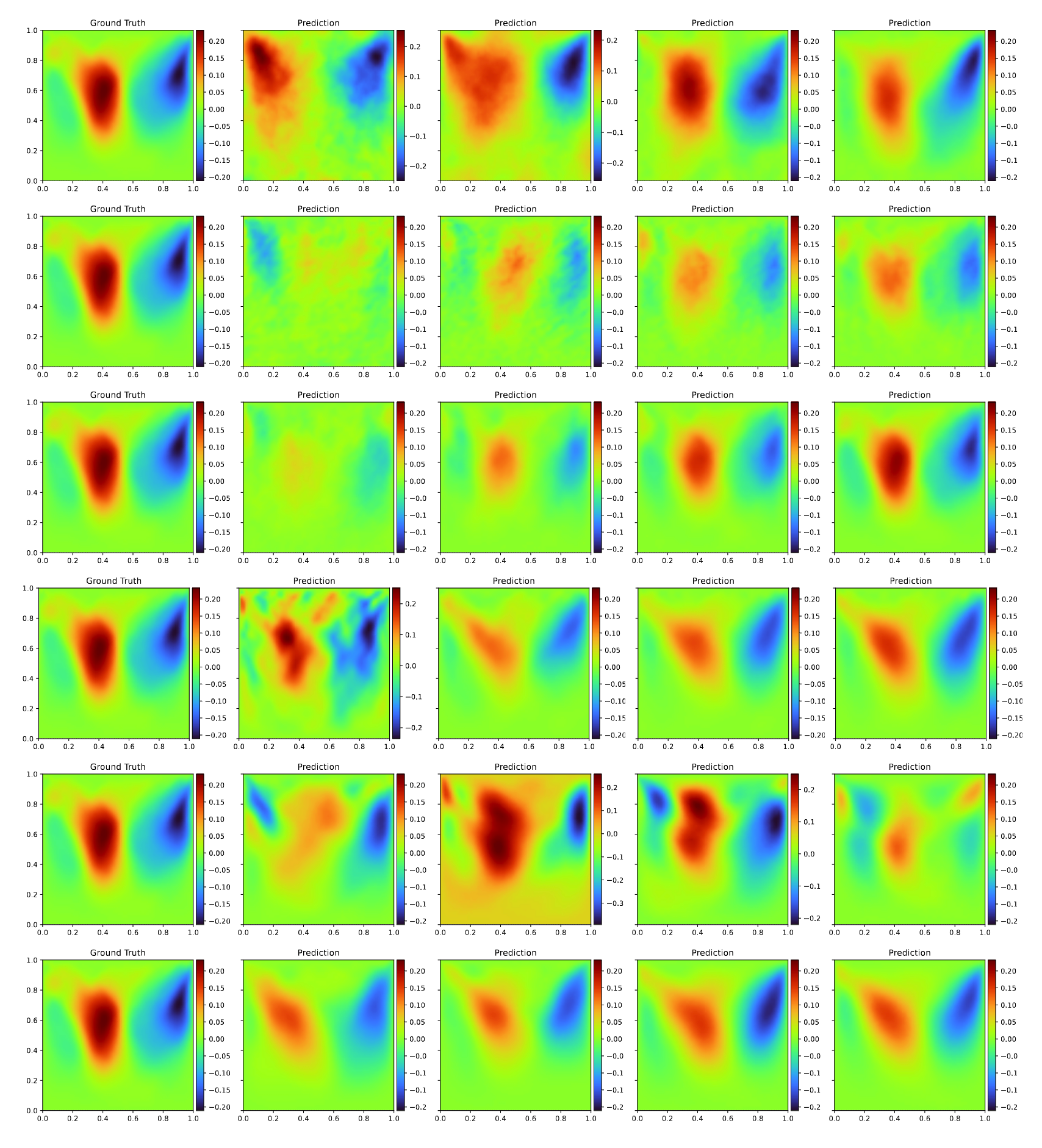}
        \end{center}
        \caption{Navier--Stokes. Comparison of the predictions of directional derivative $du_2(m;\omega_1)$ with (16, 64, 256, 1024) training samples using different methods. Top row --> bottom row: DeepONet, FNO, DE-FNO (Random), DINO, DE-DeepONet (KLE), DE-DeepONet (ASM).}
        \label{fig:truth_prediction_navier_stokes_velocity_y_dm}
    \end{figure}

    We present the comparisons of the ground truth of solution $u$, model prediction $\hat{u}$ using different methods, and absolute value of their difference $|u(x)-\hat{u}(x)|$ in~\cref{fig:truth_prediction_hyperelasticity_u1,,fig:truth_prediction_hyperelasticity_u2} for the hyperelasticity equation, and~\cref{fig:truth_prediction_navier_stokes_velocity_x,,fig:truth_prediction_navier_stokes_velocity_y} for the Navier--Stokes equations. In addition, we present the comparisons of the ground truth of the derivative of $u$ with respect to $m$ in the direction of $\omega_1$, denoted as  $du(m;\omega_1)$, model prediction $d\hat{u}(m;\psi_1^{\text{in}})$ using different methods, and absolute value of their difference $|du(m;\omega_1)-d\hat{u}(m;\omega_1)|$ in~\cref{fig:truth_prediction_hyperelasticity_dm1,,fig:truth_prediction_hyperelasticity_dm2} for the hyperelasticity equation, and~\cref{fig:truth_prediction_navier_stokes_velocity_x_dm,,fig:truth_prediction_navier_stokes_velocity_y_dm} for the Navier--Stokes equations.

\newpage
\section*{NeurIPS Paper Checklist}

\begin{enumerate}

\item {\bf Claims}
    \item[] Question: Do the main claims made in the abstract and introduction accurately reflect the paper's contributions and scope?
    \item[] Answer: \answerYes{} % Replace by \answerYes{}, \answerNo{}, or \answerNA{}.
    \item[] Justification: We include our paper's contributions in the abstract and the second paragraph of introduction and its scope in the first and third paragraphs of introduction. 
    \item[] Guidelines:
    \begin{itemize}
        \item The answer NA means that the abstract and introduction do not include the claims made in the paper.
        \item The abstract and/or introduction should clearly state the claims made, including the contributions made in the paper and important assumptions and limitations. A No or NA answer to this question will not be perceived well by the reviewers. 
        \item The claims made should match theoretical and experimental results, and reflect how much the results can be expected to generalize to other settings. 
        \item It is fine to include aspirational goals as motivation as long as it is clear that these goals are not attained by the paper. 
    \end{itemize}

\item {\bf Limitations}
    \item[] Question: Does the paper discuss the limitations of the work performed by the authors?
    \item[] Answer: \answerYes{} % Replace by \answerYes{}, \answerNo{}, or \answerNA{}.
    \item[] Justification: We discuss the limitations in the last section. 
    \item[] Guidelines:
    \begin{itemize}
        \item The answer NA means that the paper has no limitation while the answer No means that the paper has limitations, but those are not discussed in the paper. 
        \item The authors are encouraged to create a separate "Limitations" section in their paper.
        \item The paper should point out any strong assumptions and how robust the results are to violations of these assumptions (e.g., independence assumptions, noiseless settings, model well-specification, asymptotic approximations only holding locally). The authors should reflect on how these assumptions might be violated in practice and what the implications would be.
        \item The authors should reflect on the scope of the claims made, e.g., if the approach was only tested on a few datasets or with a few runs. In general, empirical results often depend on implicit assumptions, which should be articulated.
        \item The authors should reflect on the factors that influence the performance of the approach. For example, a facial recognition algorithm may perform poorly when image resolution is low or images are taken in low lighting. Or a speech-to-text system might not be used reliably to provide closed captions for online lectures because it fails to handle technical jargon.
        \item The authors should discuss the computational efficiency of the proposed algorithms and how they scale with dataset size.
        \item If applicable, the authors should discuss possible limitations of their approach to address problems of privacy and fairness.
        \item While the authors might fear that complete honesty about limitations might be used by reviewers as grounds for rejection, a worse outcome might be that reviewers discover limitations that aren't acknowledged in the paper. The authors should use their best judgment and recognize that individual actions in favor of transparency play an important role in developing norms that preserve the integrity of the community. Reviewers will be specifically instructed to not penalize honesty concerning limitations.
    \end{itemize}

\item {\bf Theory Assumptions and Proofs}
    \item[] Question: For each theoretical result, does the paper provide the full set of assumptions and a complete (and correct) proof?
    \item[] Answer: \answerYes{} % Replace by \answerYes{}, \answerNo{}, or \answerNA{}.
    \item[] Justification: We provide one theorem with clearly stated assumptions and proof. 
    \item[] Guidelines:
    \begin{itemize}
        \item The answer NA means that the paper does not include theoretical results. 
        \item All the theorems, formulas, and proofs in the paper should be numbered and cross-referenced.
        \item All assumptions should be clearly stated or referenced in the statement of any theorems.
        \item The proofs can either appear in the main paper or the supplemental material, but if they appear in the supplemental material, the authors are encouraged to provide a short proof sketch to provide intuition. 
        \item Inversely, any informal proof provided in the core of the paper should be complemented by formal proofs provided in appendix or supplemental material.
        \item Theorems and Lemmas that the proof relies upon should be properly referenced. 
    \end{itemize}

    \item {\bf Experimental Result Reproducibility}
    \item[] Question: Does the paper fully disclose all the information needed to reproduce the main experimental results of the paper to the extent that it affects the main claims and/or conclusions of the paper (regardless of whether the code and data are provided or not)?
    \item[] Answer: \answerYes{} % Replace by \answerYes{}, \answerNo{}, or \answerNA{}.
    \item[] Justification: We discuss all details (including data generation and model training) that help reproduce the experiments in the appendix.
    \item[] Guidelines:
    \begin{itemize}
        \item The answer NA means that the paper does not include experiments.
        \item If the paper includes experiments, a No answer to this question will not be perceived well by the reviewers: Making the paper reproducible is important, regardless of whether the code and data are provided or not.
        \item If the contribution is a dataset and/or model, the authors should describe the steps taken to make their results reproducible or verifiable. 
        \item Depending on the contribution, reproducibility can be accomplished in various ways. For example, if the contribution is a novel architecture, describing the architecture fully might suffice, or if the contribution is a specific model and empirical evaluation, it may be necessary to either make it possible for others to replicate the model with the same dataset, or provide access to the model. In general. releasing code and data is often one good way to accomplish this, but reproducibility can also be provided via detailed instructions for how to replicate the results, access to a hosted model (e.g., in the case of a large language model), releasing of a model checkpoint, or other means that are appropriate to the research performed.
        \item While NeurIPS does not require releasing code, the conference does require all submissions to provide some reasonable avenue for reproducibility, which may depend on the nature of the contribution. For example
        \begin{enumerate}
            \item If the contribution is primarily a new algorithm, the paper should make it clear how to reproduce that algorithm.
            \item If the contribution is primarily a new model architecture, the paper should describe the architecture clearly and fully.
            \item If the contribution is a new model (e.g., a large language model), then there should either be a way to access this model for reproducing the results or a way to reproduce the model (e.g., with an open-source dataset or instructions for how to construct the dataset).
            \item We recognize that reproducibility may be tricky in some cases, in which case authors are welcome to describe the particular way they provide for reproducibility. In the case of closed-source models, it may be that access to the model is limited in some way (e.g., to registered users), but it should be possible for other researchers to have some path to reproducing or verifying the results.
        \end{enumerate}
    \end{itemize}

\item {\bf Open access to data and code}
    \item[] Question: Does the paper provide open access to the data and code, with sufficient instructions to faithfully reproduce the main experimental results, as described in supplemental material?
    \item[] Answer: \answerYes{} % Replace by \answerYes{}, \answerNo{}, or \answerNA{}.
    \item[] Justification: We provide the code necessary to generate the data and train and test the different models considered in the paper.
    \item[] Guidelines:
    \begin{itemize}
        \item The answer NA means that paper does not include experiments requiring code.
        \item Please see the NeurIPS code and data submission guidelines (\url{https://nips.cc/public/guides/CodeSubmissionPolicy}) for more details.
        \item While we encourage the release of code and data, we understand that this might not be possible, so “No” is an acceptable answer. Papers cannot be rejected simply for not including code, unless this is central to the contribution (e.g., for a new open-source benchmark).
        \item The instructions should contain the exact command and environment needed to run to reproduce the results. See the NeurIPS code and data submission guidelines (\url{https://nips.cc/public/guides/CodeSubmissionPolicy}) for more details.
        \item The authors should provide instructions on data access and preparation, including how to access the raw data, preprocessed data, intermediate data, and generated data, etc.
        \item The authors should provide scripts to reproduce all experimental results for the new proposed method and baselines. If only a subset of experiments are reproducible, they should state which ones are omitted from the script and why.
        \item At submission time, to preserve anonymity, the authors should release anonymized versions (if applicable).
        \item Providing as much information as possible in supplemental material (appended to the paper) is recommended, but including URLs to data and code is permitted.
    \end{itemize}

\item {\bf Experimental Setting/Details}
    \item[] Question: Does the paper specify all the training and test details (e.g., data splits, hyperparameters, how they were chosen, type of optimizer, etc.) necessary to understand the results?
    \item[] Answer: \answerYes{} % Replace by \answerYes{}, \answerNo{}, or \answerNA{}.
    \item[] Justification: We discuss the experimental details in the~\cref{sec:experimental_details}. 
    \item[] Guidelines:
    \begin{itemize}
        \item The answer NA means that the paper does not include experiments.
        \item The experimental setting should be presented in the core of the paper to a level of detail that is necessary to appreciate the results and make sense of them.
        \item The full details can be provided either with the code, in appendix, or as supplemental material.
    \end{itemize}

\item {\bf Experiment Statistical Significance}
    \item[] Question: Does the paper report error bars suitably and correctly defined or other appropriate information about the statistical significance of the experiments?
    \item[] Answer: \answerYes{} % Replace by \answerYes{}, \answerNo{}, or \answerNA{}.
    \item[] Justification: We run five trails with different random seed for each method and report the mean value and standard deviation of the corresponding relative error. 
    \item[] Guidelines:
    \begin{itemize}
        \item The answer NA means that the paper does not include experiments.
        \item The authors should answer "Yes" if the results are accompanied by error bars, confidence intervals, or statistical significance tests, at least for the experiments that support the main claims of the paper.
        \item The factors of variability that the error bars are capturing should be clearly stated (for example, train/test split, initialization, random drawing of some parameter, or overall run with given experimental conditions).
        \item The method for calculating the error bars should be explained (closed form formula, call to a library function, bootstrap, etc.)
        \item The assumptions made should be given (e.g., Normally distributed errors).
        \item It should be clear whether the error bar is the standard deviation or the standard error of the mean.
        \item It is OK to report 1-sigma error bars, but one should state it. The authors should preferably report a 2-sigma error bar than state that they have a 96\% CI, if the hypothesis of Normality of errors is not verified.
        \item For asymmetric distributions, the authors should be careful not to show in tables or figures symmetric error bars that would yield results that are out of range (e.g. negative error rates).
        \item If error bars are reported in tables or plots, The authors should explain in the text how they were calculated and reference the corresponding figures or tables in the text.
    \end{itemize}

\item {\bf Experiments Compute Resources}
    \item[] Question: For each experiment, does the paper provide sufficient information on the computer resources (type of compute workers, memory, time of execution) needed to reproduce the experiments?
    \item[] Answer: \answerYes{} % Replace by \answerYes{}, \answerNo{}, or \answerNA{}.
    \item[] Justification: We provide the machine information and time of execution in the~\cref{tab:data_generation_configuration,,tab:computational_time_neural_network_training}. 
    \item[] Guidelines:
    \begin{itemize}
        \item The answer NA means that the paper does not include experiments.
        \item The paper should indicate the type of compute workers CPU or GPU, internal cluster, or cloud provider, including relevant memory and storage.
        \item The paper should provide the amount of compute required for each of the individual experimental runs as well as estimate the total compute. 
        \item The paper should disclose whether the full research project required more compute than the experiments reported in the paper (e.g., preliminary or failed experiments that didn't make it into the paper). 
    \end{itemize}
    
\item {\bf Code Of Ethics}
    \item[] Question: Does the research conducted in the paper conform, in every respect, with the NeurIPS Code of Ethics \url{https://neurips.cc/public/EthicsGuidelines}?
    \item[] Answer: \answerYes{} % Replace by \answerYes{}, \answerNo{}, or \answerNA{}.
    \item[] Justification: We have reviewed the code of ethics and checked that the research follows them.
    \item[] Guidelines:
    \begin{itemize}
        \item The answer NA means that the authors have not reviewed the NeurIPS Code of Ethics.
        \item If the authors answer No, they should explain the special circumstances that require a deviation from the Code of Ethics.
        \item The authors should make sure to preserve anonymity (e.g., if there is a special consideration due to laws or regulations in their jurisdiction).
    \end{itemize}

\item {\bf Broader Impacts}
    \item[] Question: Does the paper discuss both potential positive societal impacts and negative societal impacts of the work performed?
    \item[] Answer: \answerNA{} % Replace by \answerYes{}, \answerNo{}, or \answerNA{}.
    \item[] Justification: Our work lies in the field of scientific machine learning. We do not notice any societal impact for now. 
    \item[] Guidelines:
    \begin{itemize}
        \item The answer NA means that there is no societal impact of the work performed.
        \item If the authors answer NA or No, they should explain why their work has no societal impact or why the paper does not address societal impact.
        \item Examples of negative societal impacts include potential malicious or unintended uses (e.g., disinformation, generating fake profiles, surveillance), fairness considerations (e.g., deployment of technologies that could make decisions that unfairly impact specific groups), privacy considerations, and security considerations.
        \item The conference expects that many papers will be foundational research and not tied to particular applications, let alone deployments. However, if there is a direct path to any negative applications, the authors should point it out. For example, it is legitimate to point out that an improvement in the quality of generative models could be used to generate deepfakes for disinformation. On the other hand, it is not needed to point out that a generic algorithm for optimizing neural networks could enable people to train models that generate Deepfakes faster.
        \item The authors should consider possible harms that could arise when the technology is being used as intended and functioning correctly, harms that could arise when the technology is being used as intended but gives incorrect results, and harms following from (intentional or unintentional) misuse of the technology.
        \item If there are negative societal impacts, the authors could also discuss possible mitigation strategies (e.g., gated release of models, providing defenses in addition to attacks, mechanisms for monitoring misuse, mechanisms to monitor how a system learns from feedback over time, improving the efficiency and accessibility of ML).
    \end{itemize}
    
\item {\bf Safeguards}
    \item[] Question: Does the paper describe safeguards that have been put in place for responsible release of data or models that have a high risk for misuse (e.g., pretrained language models, image generators, or scraped datasets)?
    \item[] Answer: \answerNA{} % Replace by \answerYes{}, \answerNo{}, or \answerNA{}.
    \item[] Justification: We think our research poses no such risk.  
    \item[] Guidelines:
    \begin{itemize}
        \item The answer NA means that the paper poses no such risks.
        \item Released models that have a high risk for misuse or dual-use should be released with necessary safeguards to allow for controlled use of the model, for example by requiring that users adhere to usage guidelines or restrictions to access the model or implementing safety filters. 
        \item Datasets that have been scraped from the Internet could pose safety risks. The authors should describe how they avoided releasing unsafe images.
        \item We recognize that providing effective safeguards is challenging, and many papers do not require this, but we encourage authors to take this into account and make a best faith effort.
    \end{itemize}

\item {\bf Licenses for existing assets}
    \item[] Question: Are the creators or original owners of assets (e.g., code, data, models), used in the paper, properly credited and are the license and terms of use explicitly mentioned and properly respected?
    \item[] Answer: \answerYes{} % Replace by \answerYes{}, \answerNo{}, or \answerNA{}.
    \item[] Justification: We cite the corresponding papers and respect the licenses when using the packages FEniCS, hIPPYlib, and neuraloperator to conduct experiments. 
    \item[] Guidelines:
    \begin{itemize}
        \item The answer NA means that the paper does not use existing assets.
        \item The authors should cite the original paper that produced the code package or dataset.
        \item The authors should state which version of the asset is used and, if possible, include a URL.
        \item The name of the license (e.g., CC-BY 4.0) should be included for each asset.
        \item For scraped data from a particular source (e.g., website), the copyright and terms of service of that source should be provided.
        \item If assets are released, the license, copyright information, and terms of use in the package should be provided. For popular datasets, \url{paperswithcode.com/datasets} has curated licenses for some datasets. Their licensing guide can help determine the license of a dataset.
        \item For existing datasets that are re-packaged, both the original license and the license of the derived asset (if it has changed) should be provided.
        \item If this information is not available online, the authors are encouraged to reach out to the asset's creators.
    \end{itemize}

\item {\bf New Assets}
    \item[] Question: Are new assets introduced in the paper well documented and is the documentation provided alongside the assets?
    \item[] Answer: \answerNA{} % Replace by \answerYes{}, \answerNo{}, or \answerNA{}.
    \item[] Justification: We do not release new assets at the moment. 
    \item[] Guidelines:
    \begin{itemize}
        \item The answer NA means that the paper does not release new assets.
        \item Researchers should communicate the details of the dataset/code/model as part of their submissions via structured templates. This includes details about training, license, limitations, etc. 
        \item The paper should discuss whether and how consent was obtained from people whose asset is used.
        \item At submission time, remember to anonymize your assets (if applicable). You can either create an anonymized URL or include an anonymized zip file.
    \end{itemize}

\item {\bf Crowdsourcing and Research with Human Subjects}
    \item[] Question: For crowdsourcing experiments and research with human subjects, does the paper include the full text of instructions given to participants and screenshots, if applicable, as well as details about compensation (if any)? 
    \item[] Answer: \answerNA{} % Replace by \answerYes{}, \answerNo{}, or \answerNA{}.
    \item[] Justification: Our research is in the field of the scientific machine learning and does not involve crowdsourcing or human subjects. 
    \item[] Guidelines:
    \begin{itemize}
        \item The answer NA means that the paper does not involve crowdsourcing nor research with human subjects.
        \item Including this information in the supplemental material is fine, but if the main contribution of the paper involves human subjects, then as much detail as possible should be included in the main paper. 
        \item According to the NeurIPS Code of Ethics, workers involved in data collection, curation, or other labor should be paid at least the minimum wage in the country of the data collector. 
    \end{itemize}

\item {\bf Institutional Review Board (IRB) Approvals or Equivalent for Research with Human Subjects}
    \item[] Question: Does the paper describe potential risks incurred by study participants, whether such risks were disclosed to the subjects, and whether Institutional Review Board (IRB) approvals (or an equivalent approval/review based on the requirements of your country or institution) were obtained?
    \item[] Answer: \answerNA{} % Replace by \answerYes{}, \answerNo{}, or \answerNA{}.
    \item[] Justification: Our research is in the field of the scientific machine learning and does not involve crowdsourcing or human subjects.
    \item[] Guidelines:
    \begin{itemize}
        \item The answer NA means that the paper does not involve crowdsourcing nor research with human subjects.
        \item Depending on the country in which research is conducted, IRB approval (or equivalent) may be required for any human subjects research. If you obtained IRB approval, you should clearly state this in the paper. 
        \item We recognize that the procedures for this may vary significantly between institutions and locations, and we expect authors to adhere to the NeurIPS Code of Ethics and the guidelines for their institution. 
        \item For initial submissions, do not include any information that would break anonymity (if applicable), such as the institution conducting the review.
    \end{itemize}

\end{enumerate}

\end{document}